\documentclass{article}
\pdfoutput=1

\usepackage{microtype}
\usepackage{graphicx}
\usepackage{subcaption}
\usepackage{booktabs} %
\usepackage{multirow}
\usepackage{siunitx}

\usepackage{hyperref}

\usepackage{tikz}
\usetikzlibrary{arrows.meta}
\usetikzlibrary{calc}

\PassOptionsToPackage{numbers, compress}{natbib}

\usepackage[main, final]{neurips_2026}

\usepackage[utf8]{inputenc} %
\usepackage[T1]{fontenc}    %

\usepackage{amsmath}
\usepackage{amssymb}
\usepackage{mathtools}
\usepackage{amsthm}

\usepackage[capitalize,noabbrev]{cleveref}
\usepackage{tcolorbox}
\usepackage{wrapfig}
\usepackage{tabularx}
\usepackage{booktabs}
\usepackage{amssymb}   %
\usepackage{pifont}    %
\newcommand{\xmark}{\ding{55}}

\ifPDFTeX
  \usepackage{wasysym} %
  \newcommand{\daysym}{\(\astrosun\)}
  \newcommand{\nightsym}{\(\fullmoon\)}
\else
  \usepackage{fontspec}
  \usepackage{emoji} %
  \newcommand{\daysym}{\(\text{\emoji{sun}}\)}
  \newcommand{\nightsym}{\(\text{\emoji{new-moon}}\)}
\fi

\newcommand{\Ldir}{\(\leftarrow\)}
\newcommand{\Rdir}{\(\rightarrow\)}
\newcommand{\Sdir}{\(\uparrow\)}
\newcommand{\fact}[2]{#1\!\,#2} %

\usepackage{cuted}
\usepackage{capt-of}

\newcounter{finding}
\renewcommand{\thefinding}{\arabic{finding}} %

\newcommand{\findinglabel}[1]{%
  \refstepcounter{finding}%
  \label{#1}%
  \textbf{Finding~\thefinding:}%
}

\crefname{finding}{Finding}{Findings}
\Crefname{finding}{Finding}{Findings}

\definecolor{maskgitval}{gray}{0.55} %
\newcommand{\mgitval}[1]{{\begingroup\color{maskgitval}\bfseries #1\endgroup}}

\definecolor{lightblue}{rgb}{0.4, 0.7, 1}
\definecolor{darkpink}{rgb}{0.8, 0.1, 0.3}
\definecolor{GainGreen}{HTML}{1B7F3B}

\definecolor{darkblue3}{HTML}{1f77b4}   %
\definecolor{lightblue3}{HTML}{aec7e8}  %
\definecolor{paleorange3}{HTML}{ffbb78} %
\definecolor{darkred3}{HTML}{d62728}    %

\newcommand{\gain}[1]{\textcolor{GainGreen}{#1}}
\providecommand{\fact}[2]{\langle #1,\ #2\rangle}    %

\theoremstyle{plain}

\theoremstyle{definition}

\theoremstyle{remark}

\makeatletter
\@ifundefined{@makespecialcolbox}{%
  \@ifundefined{@make@specialcolbox}{}{%
    \let\@makespecialcolbox\@make@specialcolbox
  }%
}{}
\makeatother

\title{What Drives Compositional Generalization in Visual Generative Models? The Importance of Continuous Training Objectives}

\author{%
  Karim Farid$^{*1}$ \And Rajat Sahay$^{*1,2}$ \And Yumna Ali$^{*1}$ \And Simon Schrodi$^{1}$
  \AND
  Volker Fischer$^{2}$ \And Cordelia Schmid$^{3}$ \And Thomas Brox$^{1}$
  \AND
  \normalfont $^{1}$University of Freiburg \quad $^{2}$Bosch Center for Artificial Intelligence \\ $^{3}$Inria, {\'E}cole Normale Sup{\'e}rieure, CNRS, PSL Research University \\
  \AND
  \normalfont \texttt{\{faridk, sahayr, aliy\}@cs.uni-freiburg.de}
}

\begin{document}

\maketitle
{\renewcommand{\thefootnote}{*}\footnotetext{Equal contribution.}}

\begin{figure*}[h]
\vspace{-30pt}
  \centering
  \includegraphics[width=\textwidth]{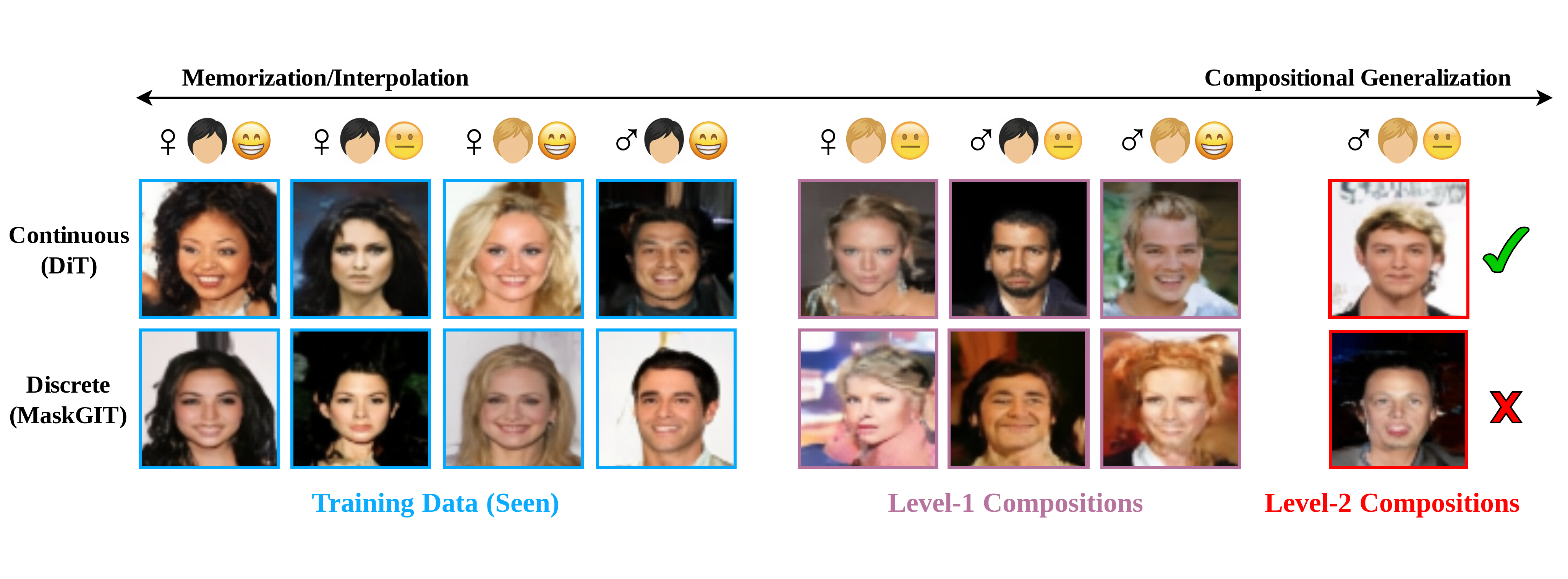}
  \vspace{-20pt}
\caption{
  \textbf{Compositional Generalization Analysis.}
  We compare discrete masking (MaskGIT) and continuous diffusion (DiT) on novel compositions of three binary CelebA factors: gender, hair color, and smile. Models are trained on four observed combinations (\textcolor{lightblue}{blue}) and evaluated on held-out Level-1 (\textcolor{darkpink}{pink}, one-factor change) and Level-2 (\textcolor{red}{red}, two-factor change) compositions. DiT (row 1) generalizes compositionally, while MaskGIT (row 2) struggles.
}

\label{fig:celeba_teaser}
\end{figure*}

\begin{abstract}

Compositional generalization, the ability to generate novel combinations of known concepts, is a key ingredient for visual generative models. Yet, not all mechanisms that enable or inhibit it are fully understood. In this work, we conduct a systematic study of which design choices critically determine compositional generalization in image and video generation. By isolating independent design axes, we identify two key factors strongly associated with compositional success: (i) whether the training objective operates on a discrete or continuous distribution, and (ii) the completeness of conditioning information about constituent factors during training. We also show that relaxing the discrete loss with an auxiliary continuous latent objective can partially recover compositional performance in discrete models like MaskGIT. Our findings, corroborated by diverse compositional tasks and preliminary evidence in world models and LLMs, motivate a shift toward continuous objectives for compositional generalization.

\end{abstract}

\section{Introduction}

Visual generative models, such as diffusion models \citep{kingma2021variational,nichol2021diffusion,yang2023diffusion} and generative transformers \citep{wang2022git,kim2023magvit,hudson2021generative}, can generate high-fidelity images and videos \citep{villegas2022phenaki,ho2022vdm,karras2020stylegan}, especially when trained on large amounts of data.
However, are these models capable of robust \emph{compositional generalization} or are they merely ``interpolating their training data?'' That is, can they systematically decompose the observed data into its underlying factors or ``concepts'' (e.g., objects, attributes) and synthesize novel combinations not seen during training \citep{zhao2022toward,wiedemer2023principles, favero2025compositional}?

Despite significant progress in generative modeling, compositional generalization in current models remains inconsistent. While some studies report successes on targeted compositional tasks \citep{wiedemer2025clip,gaudi2025coind,redhardt2025scaling}, others reveal notable limitations, particularly when trying to generate complex scenes or novel combinations of known elements \citep{an2023context,keysers2019measuring}. The factors that drive (or hamper) compositional generalization are not yet clearly identified. Hence the field operates inefficiently. For example, significant resources are currently partitioned between continuous diffusion \citep{google_veo3_2025, labs2025flux} and discrete masked modeling \citep{liuinfinitystar, wang2024emu3, bruce2024genie} families, a paradigm that has been bolstered by the scaling successes of discrete Large Language Models (LLMs). Yet, which representation generalizes ``better'' out of distribution (OOD) remains open.

\cref{fig:celeba_teaser} illustrates a striking difference between visual generative model families. Although both models are trained on a subset of compositions (i.e., ``non-smiling, women with blonde hair'' and ``smiling men with black hair''), a \emph{continuous} DiT \citep{peebles2023dit} generates a well-looking ``non-smiling man with blonde hair'', whereas a \emph{discrete} MaskGIT \citep{chang2022maskgit} produces a severely distorted image. This brings us to our core research question:

\begin{center}
    \emph{What factors drive compositional generalization in visual generative models?}
\end{center}

To answer this question, we develop a systematic framework for isolating the effects of individual design choices. We categorize modern visual generative models along three primary axes:
(i) the \emph{Tokenizer}, which defines the underlying representation space; (ii) the \emph{Generative model}, which generates samples in this space guided by the conditioning signal (task), and (iii) the \emph{Conditioning signal}, which specifies the factors to be generated. Note that the last is also related to scale/diversity of the training data, as large, diverse data can help identify the factors when they are not specified.
To anchor our study, we select the simplest, most canonical representatives of each model family--MaskGIT (discrete/masked) and DiT (continuous/diffusion)--to establish a controlled baseline.
We then systematically ``interpolate'' between them by varying individual design axes (e.g., tokenizer type, masking strategy, training objective) while holding others fixed. Thereby, we isolate the factors that affect their compositional generalization capability, independent of engineering refinements that could be applied to any model in any family. Adopting these architectural minimal pairs allows us to compare across multiple variants in different families without introducing additional confounders.  %

Concretely, we investigate the following research questions:

\begin{itemize}
    \itemsep -0pt
    \item \textbf{RQ1}: \textit{Does  the type of the tokenizer affect compositional generalization?} (VAE with KL regularization vs. VQ-VAE with quantization/commitment regularization.)
    \item \textbf{RQ2}: \textit{How does the generative model design affect compositionality?} In particular, does it matter whether the modeled distribution is continuous or discrete (e.g., continuous latents vs. discrete tokens)? Is a denoising objective essential, or does a masking loss suffice?
    \item \textbf{RQ3}: \textit{Can generative models, without being conditioned on the full data-generating factors during training, generalize compositionally?} Must conditioning have the exact factors, or can they rely on quantized or missing abstractions (e.g., ``red'' instead of the RGB value, or hiding ``smile'' in \{``blond'', ``smile'', ``girl''\})?
    \item \textbf{RQ4}: Guided by our findings,  \textit{can we intervene on non-compositional models to make them compositional?}
\end{itemize}

To shed light on these questions, we conduct controlled experiments across a range of architectural and training design choices. The results consistently indicate that models trained to learn a \emph{continuous distribution} by their training objective exhibit stronger compositional capabilities than models trained to model a categorical distribution. Possibly less surprising, we also find that \emph{providing full conditioning information of the generating factors during training} is important; quantized or partial conditioning leads to weaker compositional generalization. On the other hand, training the tokenizer with a quantization bottleneck has no significant effect on compositional generalization.

Guided by our findings and to further examine why discrete modeling hinders compositional generalization, we test whether augmenting the categorical training objective with a continuous Joint Embedding Predictive Architecture (JEPA) objective \citep{lecun2022jepa} can improve compositional performance. This modification introduces continuous latent targets and yields clear improvements on held-out factor combinations. At the same time, models trained with this auxiliary objective exhibit more disentangled intermediate representations, suggesting that predictive continuous objectives can shape the internal structure to support compositionality in discrete models.

\begin{tcolorbox}[colback=blue!4, colframe=black, coltitle=black, boxrule=0.3mm]
\textbf{Summary of contributions}:
Through controlled experiments on transformer-based visual generative models, we show that (i) models trained with continuous output objectives exhibit substantially stronger compositional generalization than models trained with discrete categorical objectives across the design choices we study; (ii) complete and precise conditioning on generative factors substantially improves compositional generalization; and (iii) augmenting discrete models with a continuous auxiliary objective partially recovers compositionality.
\end{tcolorbox}

\section{Related Works}

Factorization and compositional generalization have been extensively studied across a range of architectures~\citep{wu2019multimodal,li2024context,yang2024dynamics}, as they are central to building models that can capture and recombine the underlying factors of variation in data. Prior work has approached this challenge from multiple angles. On the theoretical side, several studies seek first-principles characterizations of compositionality based on the data-generating process \citep{wiedemer2023principles} or provide provable guarantees through structural constraints such as object-centric autoencoders \citep{wiedemer2024provable}. Our work instead takes an empirical perspective, systematically probing how architectural and training choices shape compositional generalization in visual generative models.

Complementary to these theoretical accounts, recent empirical investigations highlight how large-scale models such as CLIP \citep{radford2021clip} exhibit degrees of compositionality, often tied to the properties of their training data \citep{kempf2025clip,wiedemer2025clip}. Our focus in this work is not on data, but on architectural, training, and conditioning choices that enhance compositionality.

For generative models, \citet{okawa} studied compositional generalization solely for diffusion models. We extend this work along two dimensions: (i) by considering a broader set of generative frameworks, including both continuous and discrete models, and (ii) by examining additional data modalities such as video. Our work also connects to the rich literature on disentangled representation learning \citep{higgins2018def,casellesdupre2019rep}, which aims to align individual latents with interpretable factors of variation. While disentanglement work focuses on representation quality, our study emphasizes the downstream generative consequences: we analyze how different training objectives and conditioning strategies enable, or hinder, the synthesis of novel compositions.

\section{Experimental Setup}
\label{sec:problem_setup}

Our goal is to identify which architectural and training choices enable or hinder compositional generalization (as seen in \cref{fig:celeba_teaser}). To this end, we design controlled comparisons that systematically vary key factors-representation space, training objective, and conditioning information levels-while holding others fixed. This allows us to ``interpolate'' between the canonical model pairs from different families (such as DiT and MaskGIT), and isolate the contributions of individual design decisions.

\paragraph{``Interpolation'' between the minimal architectural pairs}
The main differences between DiT and MaskGIT are in the following design choices: the tokenizer (VAE/VQ-VAE for DiT vs. strictly VQ-VAE for MaskGIT), the training objective (masking-based for MaskGIT, absent in DiT), the loss function (diffusion in DiT vs. categorical negative log-likelihood (NLL) in MaskGIT), and the nature of the latent output distribution (continuous for DiT vs. categorical discrete for MaskGIT). To isolate which choices affect compositional capability, we systematically vary them one at a time.

 Recent works have effectively implemented several such interpolations; see \cref{tab:model_summary}. For example, MAR \citep{li2024mar} differs from DiT only in adopting a masking-based prediction while retaining the per-token diffusion objective. Similarly, GIVT \citep{tschannen2024givt} replaces MaskGIT’s standard categorical NLL with a Gaussian mixture negative log-likelihood (GMM-NLL), where the token distribution is modeled as a mixture of Gaussians \citep{reynolds2009gmm} (which allows GIVT to operate in a continuous space without employing the denoising diffusion objective). We adopt these controlled variants for our experiments while neutralizing potential confounders. We use \emph{continuous objective} for objectives over continuous outputs, in contrast to categorical prediction over discrete token-ids, which lack a metric structure. Since each objective is only defined over its own prediction (and thus input) space, we treat them as one design axis.

\begin{wraptable}{r}{0.55\columnwidth}
\centering
\begin{small}
\setlength{\tabcolsep}{4pt}
\caption{Summary of generative models.}
\label{tab:model_summary}
\begin{tabularx}{\linewidth}{l c c X}
\toprule
Model & Masking-based & Training loss & Latent Distribution\\
\midrule
MaskGIT &  \checkmark & Categorical NLL & discrete \\
GIVT & \checkmark & GMM-NLL & continuous \\
MAR &  \checkmark & Diffusion & continuous \\
DiT &  \xmark & Diffusion & continuous \\
\bottomrule
\end{tabularx}
\end{small}
\end{wraptable}

\paragraph{Testing compositional generalization}

We train models on pairs of images conditioned on tuples specifying the underlying factors (e.g., object color). We train only on a subset of the possible factor combinations and evaluate compositional generalization on held-out combinations. We distinguish between \textit{level-1} compositions, which differ from the nearest combination seen during training by one factor, and \textit{level-2} compositions, which differ by two factors.
While our primary analysis focuses on simple synthetic images with three binary factors (\textbf{Shapes2D}) following Okawa et al.~\cite{okawa}, we benchmark the robustness of our findings across diverse domains. This includes multi-valued factors (\textbf{Shapes3D}), real-world faces (\textbf{CelebA}), and synthetic videos (\textbf{CLEVRER-Kubric}). Furthermore, we also show that our findings transfer to complex real-world scenes (\textbf{CoVLA}) using the Orbis model, and provide preliminary evidence on a Points24-inspired language reasoning task. Detailed dataset descriptions are in Appendix~\ref{appendix:datasets}.

\paragraph{Evaluation}

To evaluate whether generated images faithfully capture the intended factors, we train probes for each factor to classify its presence in the generated outputs. We train the probes on \textit{all} possible combinations, including held-outs from model training.
Following \citet{okawa,park2024emergence}, a composition is considered correct if all its factors' probe probability is the highest (argmax); we verify that probe decisions are rarely borderline and track visual quality in \cref{app:probe_validity}.
While we additionally report standard metrics (FID, FDD, CRA) on real datasets in \cref{appendix:eval}, probe accuracy directly measures whether the intended factors are present, rather than overall image quality only.

For visualizing training dynamics, we use \textcolor{lightblue}{blue} curves for compositions seen in training, \textcolor{darkpink}{pink} curves for level-1 compositions, and \textcolor{red}{red} curves for level-2 compositions. Performance on level-2 compositions serves as a particularly informative metric for compositional generalization, as they require the model to generate multiple novel factors simultaneously.

\section{Design choices that drive compositionality}
\label{sec:factors}

\begin{figure*}[t]
    \centering
    \resizebox{\textwidth}{!}{
        \begin{minipage}{\textwidth}
        \centering
            \begin{subfigure}[b]{0.3\textwidth}
                \centering
                \includegraphics[width=\textwidth]{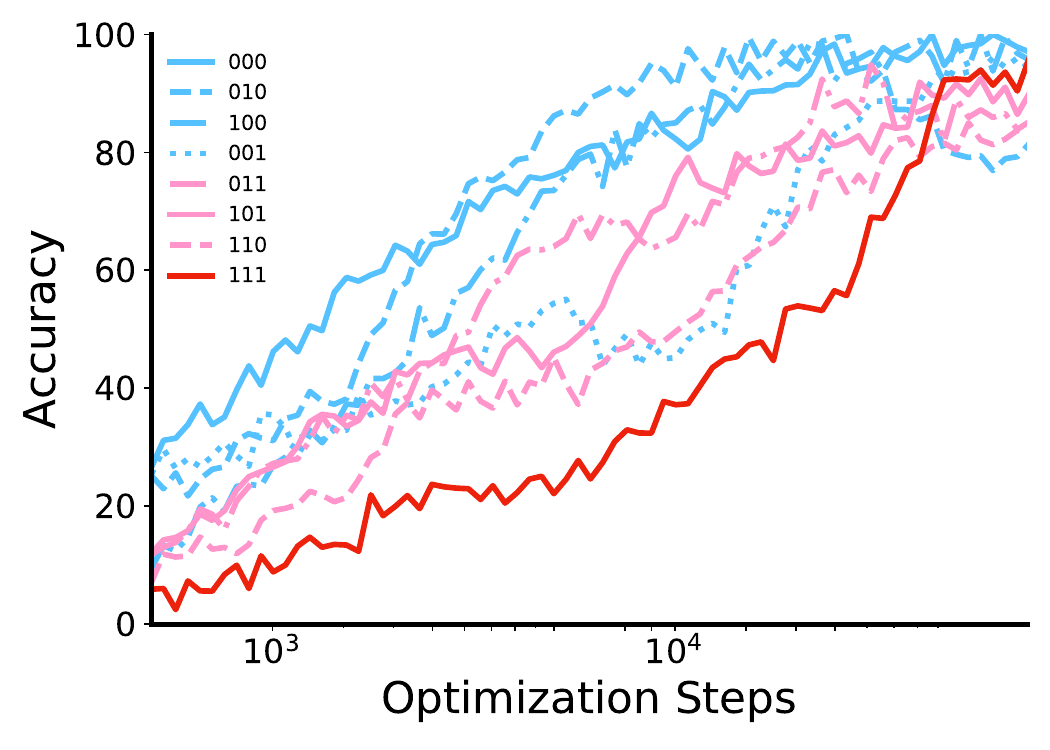}
            \caption{VAE}
            \label{fig:diffusion_token}
            \end{subfigure}
            \begin{subfigure}[b]{0.3\textwidth}
                \centering
                    \includegraphics[width=\textwidth]{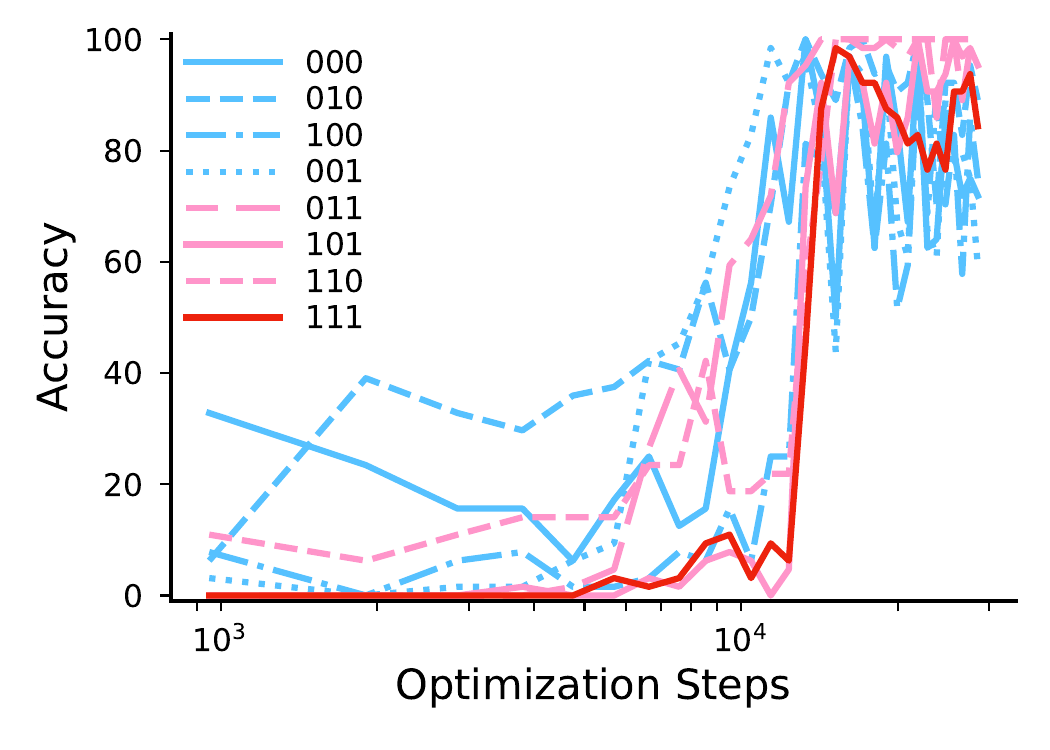}
                \caption{VQ-VAE }
            \label{fig:diffusion_vae}
            \end{subfigure}
        \end{minipage}
    }
        \caption{\textbf{DiT exhibits compositional generalization regardless of the type of tokenizer used.}
        While the training dynamics differ, DiT shows compositional generalization at the end of training.
        The \textcolor{lightblue}{blue}, \textcolor{darkpink}{pink}, and \textcolor{red}{red} curves show linear probe accuracies for the training data, level-1 compositions, or level-2 compositions, respectively. Results are consistent for MAR (\cref{fig:mar_tokenizer_results}) and video (\cref{fig:dit_clevrer_tokenizer_results}).
        }
    \label{fig:tokenizer_results}
\end{figure*}

In this section, we leverage the setup in~\cref{sec:problem_setup} to systematically rule out irrelevant factors in our architectural design space by interpolating from DiT to MaskGIT. This process isolates the design elements that are essential for enabling compositional generalization in DiT but not MaskGIT. We focus on results from the Shapes2D dataset in the following. However, our findings also extend to more complex, real-world, and video datasets (quantitative and qualitative results in Appendix~\ref{appendix:results}). In addition, Appendix~\ref{app:theory} discusses existing theoretical results on compositional generalization that offer perspectives consistent with our findings.

\paragraph{Does the choice of tokenizer matter?}

The tokenizer is one of the main differences between DiT and MaskGIT (\cref{tab:model_summary}). To isolate and assess its effects on compositionality (\textbf{RQ1}), we evaluate the same second-stage architecture (DiT) with two different tokenizers: a discrete tokenizer (VQ-VAE) and a continuous tokenizer (VAE). Additional implementation details are provided in \cref{appendix:tokenizer_results}.

\cref{fig:tokenizer_results} shows that DiT achieves comparable compositional performance with both tokenizer types by the end of training, though training dynamics differ: continuous tokenizers yield gradual progress while discrete tokenizers show more abrupt emergence. Further, we find the discrete tokenizer to be more sensitive to learning rates.\footnote{A detailed analysis of these differences is left for future work.} Hence, tokenizer regularization does not erode compositionality and is unlikely to explain the DiT--MaskGIT gap.

\begin{tcolorbox}[colback=blue!4, colframe=black, coltitle=black, boxrule=0.3mm]
     \findinglabel{find:vae-vs-vqvae}
     The choice of tokenizer does not substantially alter the compositional abilities of a generative model. It primarily affects training efficiency and stability.
\end{tcolorbox}

Since tokenizer choice does not affect compositionality and MaskGIT requires discrete tokens, we fix the tokenizer to VQ-VAE to avoid \textit{VQ} vs. \textit{KL} regularization as a potential confounder. Discrete models use codebook indices; continuous models use pre-quantization latents (details in \cref{appendix:tokenizer_results}).

\paragraph{Does the masking matter?}
Another main difference between DiT and MaskGIT is their objective: MaskGIT uses a masking-based loss\footnote{“Masking-based”: predicting a subset of tokens from the rest, via continuous or discrete masks.}, whereas DiT relies on the standard denoising approach (\cref{tab:model_summary}). MAR can be viewed as a hybrid or ``interpolation'' between these two: it combines a masking-based objective with a diffusion loss per token.

\begin{figure*}[t]
  \centering

  \begin{subfigure}[b]{0.24\textwidth}
    \centering
    \includegraphics[width=\textwidth]{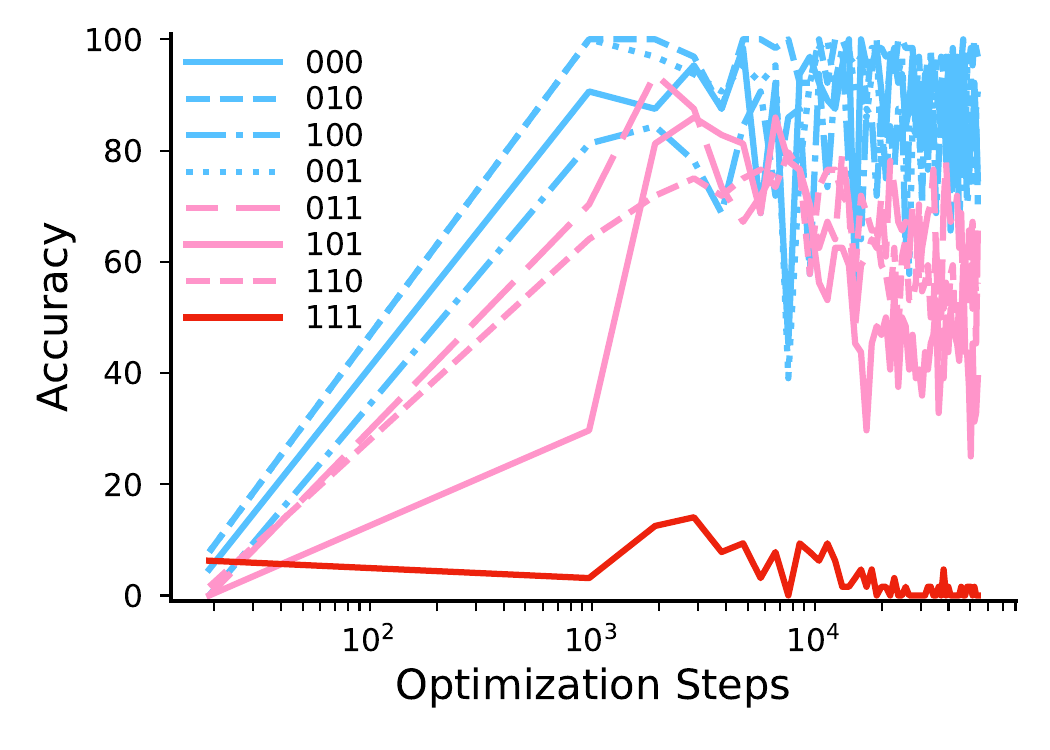}
    \caption{MaskGIT}
    \label{fig:maskgit_results}
    \tikz[remember picture] \coordinate (capA);
  \end{subfigure}\hfill
  \begin{subfigure}[b]{0.24\textwidth}
    \centering
    \includegraphics[width=\textwidth]{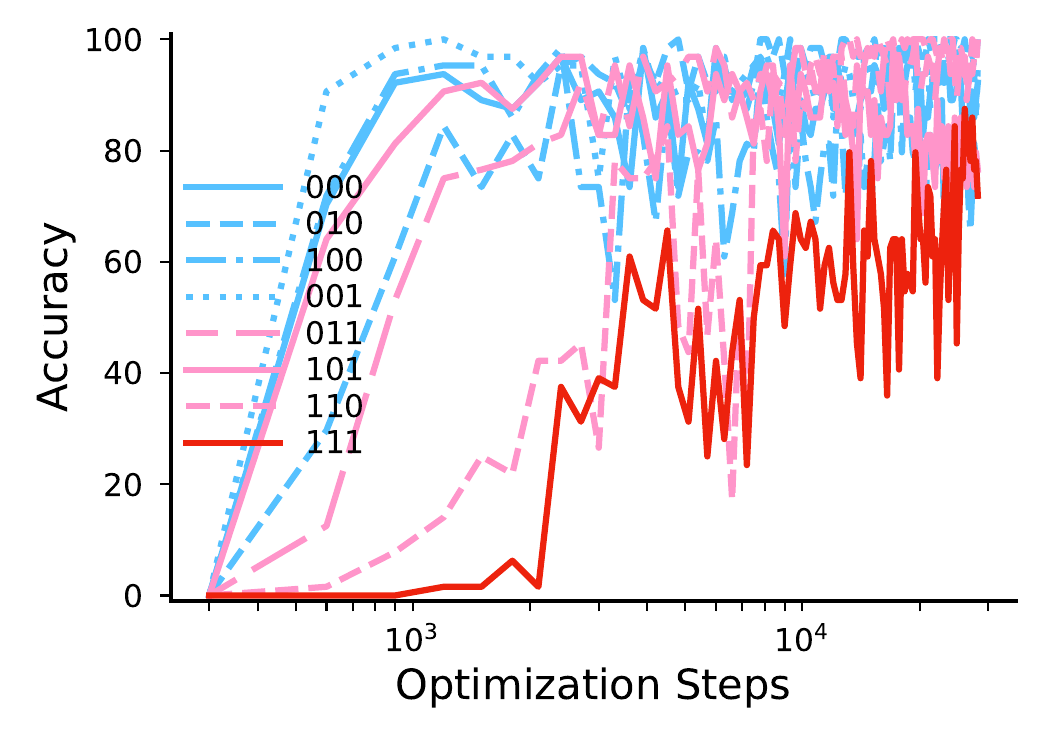}
    \caption{GIVT}
    \label{fig:givt_results}
    \tikz[remember picture] \coordinate (capB);
  \end{subfigure}\hfill
  \begin{subfigure}[b]{0.24\textwidth}
    \centering
    \includegraphics[width=\textwidth]{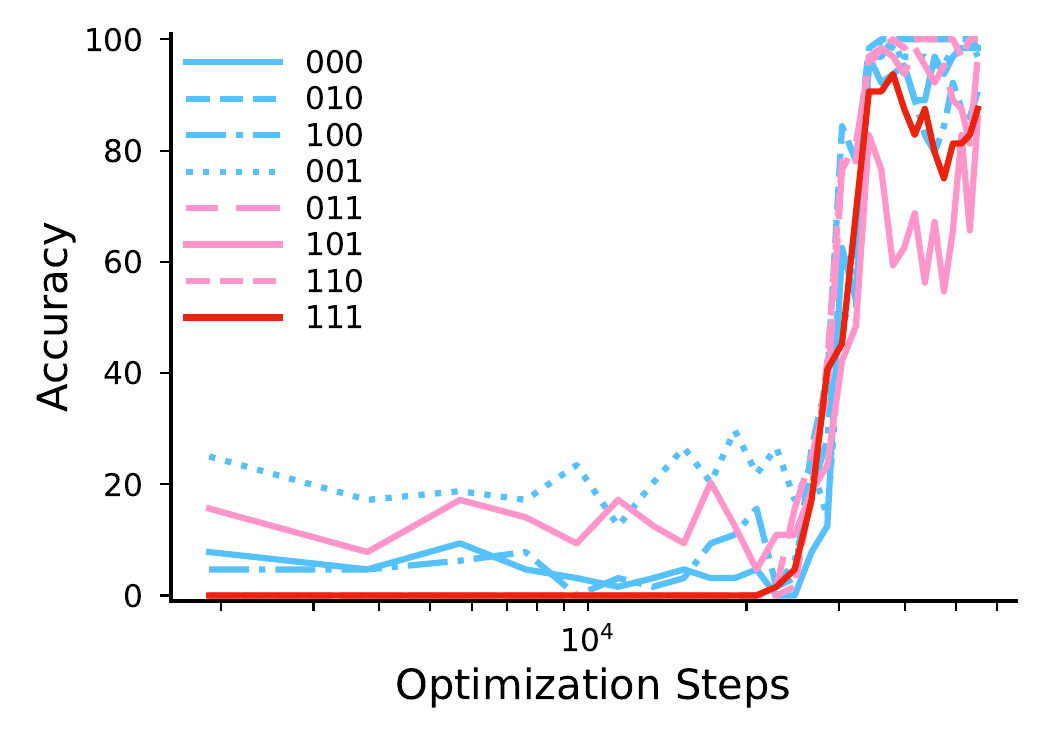}
    \caption{MAR}
    \label{fig:mar_results}
    \tikz[remember picture] \coordinate (capC);
  \end{subfigure}\hfill
  \begin{subfigure}[b]{0.24\textwidth}
    \centering
    \includegraphics[width=\textwidth]{figures/diffusion_token.pdf}
    \caption{DiT}
    \label{fig:diffusion_results}
    \tikz[remember picture] \coordinate (capD);
  \end{subfigure}

  \par\vspace{0\baselineskip}
    \begin{tikzpicture}[remember picture,overlay]
      \tikzset{chglab/.style={
          font=\scriptsize\bfseries,
          fill=white,
          inner sep=1pt,
          outer sep=0pt,
          text opacity=1
      }}

      \draw[-{Latex[length=2mm]},thick,shorten >=6pt,shorten <=6pt]
        ([yshift=0.0ex]capA.south east) -- ([yshift=0.0ex]capB.south west)
        node[pos=0.5,above=0.3ex,chglab]{\textbf{\textcolor{red}{-- categorical discrete}}}
        node[pos=0.5,below=0.3ex,chglab]{\textbf{\textcolor{green!30!black}{+ GMM continuous}}};

      \draw[-{Latex[length=2mm]},thick,shorten >=6pt,shorten <=6pt]
        ([yshift=0.0ex]capB.south east) -- ([yshift=0.0ex]capC.south west)
        node[pos=0.5,above=0.3ex,chglab]{\textbf{\textcolor{red}{-- GMM}}}
        node[pos=0.5,below=0.3ex,chglab]{\textbf{\textcolor{green!30!black}{+ diffusion head per token}}};

      \draw[-{Latex[length=2mm]},thick,shorten >=6pt,shorten <=6pt]
        ([yshift=0.0ex]capC.south east) -- ([yshift=0.0ex]capD.south west)
        node[pos=0.5,above=0.3ex,chglab]{\textbf{\textcolor{red}{-- masking}}}
        node[pos=0.5,below=0.3ex,chglab]{\textbf{\textcolor{green!30!black}{+ diffusion on all tokens}}};
    \end{tikzpicture}

  \caption{\textbf{Compositional generalization performance on Shapes2D across different model architectures.}
  Models that learn continuous distributions (DiT, MAR, and GIVT) consistently show better level-2 compositions than MaskGIT, with the decisive shift in performance occurring at the categorical-to-continuous intervention.
  The \textcolor{lightblue}{blue}, \textcolor{darkpink}{pink}, and \textcolor{red}{red} curves denote training, level-1, and level-2 compositions, respectively. Consistent results are observed for CLEVRER-Kubric (\cref{fig:latent_space_results_clevrer}).}
  \label{fig:latent_space_results}
  \vspace{-10pt}
\end{figure*}

\cref{fig:mar_results} shows that MAR achieves robust compositional generalization by the end of training. This uncouples the masking in the training objective from the compositionality failure modes of MaskGIT and does not critically impact the model's ability to generalize compositionally; neither does MaskGIT's confidence-based unmasking order (\cref{app:unmasking_order}).

\paragraph{Is it the diffusion loss?}
We look at another major difference between DiT and MAR vs MaskGIT: their training losses. GIVT \citep{tschannen2024givt} provides a bridge between MAR and MaskGIT on the objective axis by replacing MaskGIT's categorical head with a continuous Gaussian Mixture Model (GMM). This intervention effectively substitutes the diffusion loss of MAR/DiT with maximum likelihood training. This is done by using the continuous parameters of a GMM \citep{reynolds2009gmm} while preserving the continuous output space.
\cref{fig:maskgit_results} $\rightarrow$ \ref{fig:givt_results} shows that the MaskGIT $\rightarrow$ GIVT intervention (categorical $\rightarrow$ continuous) shows a substantial jump in compositional performance, whereas GIVT $\rightarrow$ MAR (GMM $\rightarrow$ diffusion) brings comparatively smaller gains as seen in \cref{fig:givt_results} $\rightarrow$\ref{fig:mar_results}. This identifies a primary driver: the continuous training objective on a continuous output space, rather than the exact form of the objective (denoising vs. parameter prediction), is a key factor for enabling compositional generalization along the axes we varied.

\paragraph{Output-space continuity emerges as the primary factor}
By systematically isolating the architectural axes separating the two main classes of visual generative models, including the tokenizer, masking strategies, and loss functions, we have ruled out these factors as primary drivers of compositional generalization. The remaining distinguishing factor is the nature of the predicted outputs: DiT, MAR, and GIVT predict continuous-valued quantities, while MaskGIT predicts discrete tokens. We conclude that this difference in output representation (and thereby the objective) is a primary factor behind the compositional generalization of continuous models, which discrete models such as MaskGIT lack (also under causal autoregressive decoding; \cref{app:ar_baseline}).

\begin{tcolorbox}[colback=blue!4, colframe=black, coltitle=black, boxrule=0.3mm]
    \findinglabel{find:cont-vs-cat} Generative models trained to model continuous distributions consistently exhibit stronger compositionality, compared to models trained on discrete, categorical distributions.
\end{tcolorbox}
\vspace{-10pt}
\section{Importance of the conditioning information levels}
\label{sec:conditioning}

\begin{figure*}[t]
    \centering
            \begin{subfigure}[b]{0.23\textwidth}
                \centering
                \includegraphics[width=\textwidth]{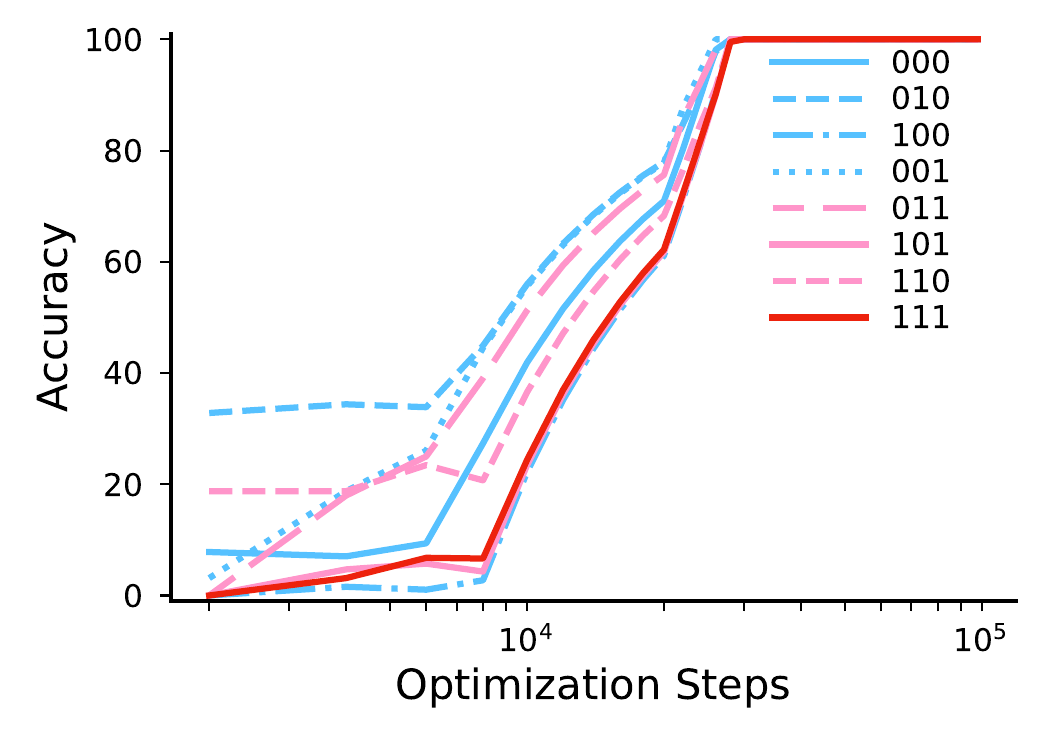}
                \caption{Full-information conditioning}
                \label{fig:conditiong_continuous}
            \end{subfigure}
            \hfill
            \begin{subfigure}[b]{0.23\textwidth}
                \centering
                \includegraphics[width=\textwidth]{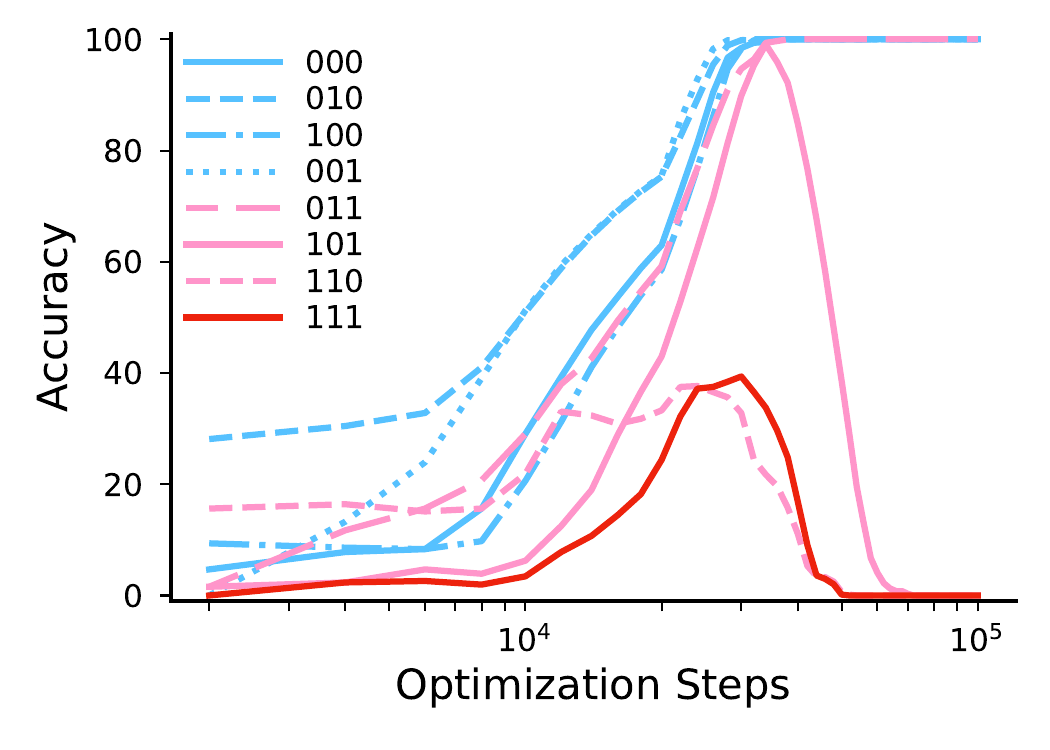}
                \caption{Full-information conditioning + dropout}
                \label{fig:conditiong_cont_dropout}
            \end{subfigure}
            \hfill
            \begin{subfigure}[b]{0.23\textwidth}
                \centering
                \includegraphics[width=\textwidth]{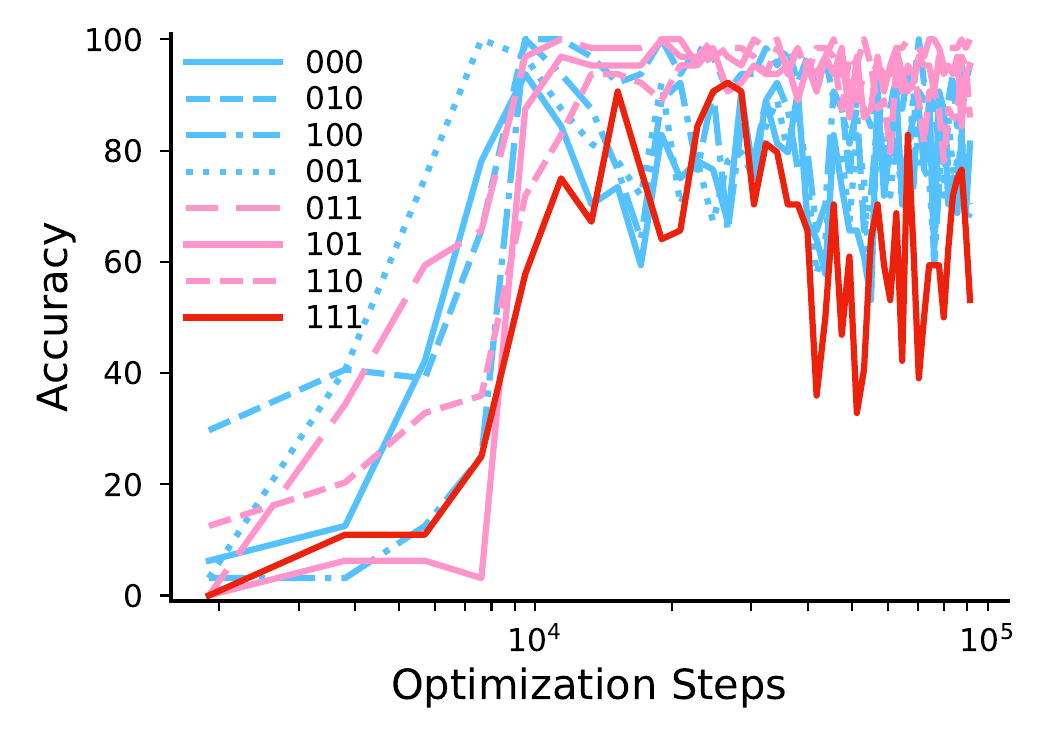}
                \caption{Quantized conditioning }
                \label{fig:conditiong_discrete}
            \end{subfigure}
            \hfill
            \begin{subfigure}[b]{0.23\textwidth}
                \centering
                \includegraphics[width=\textwidth]{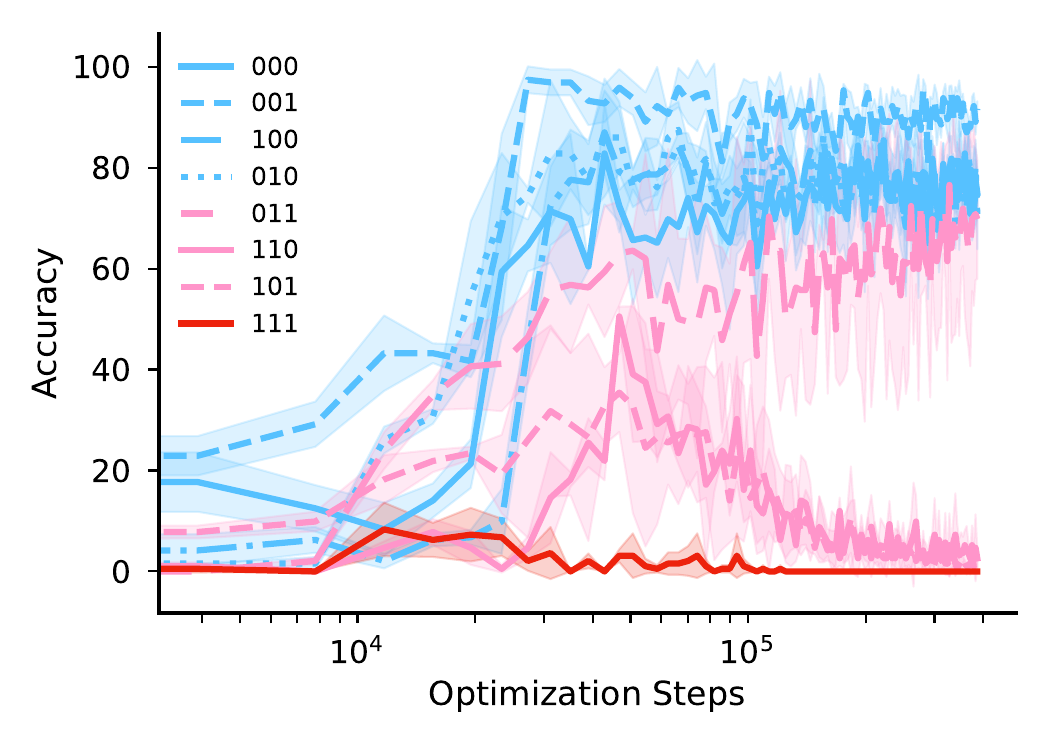}
                \caption{Quantized conditioning + dropout}
                \label{fig:conditiong_mechanism_dropout}
            \end{subfigure}
    \caption{\textbf{Comparison of conditioning information levels and their impact on compositional generalization in DiT on Shapes2D.}
    \textbf{(a)} Continuous (full-information) conditioning leads to uniform convergence across all compositions.
 \textbf{(b)} Label dropout conditioning leads to inconsistent generalization; several unseen compositions fail. \textbf{(c)} Discrete (quantized) conditioning leads to partial generalization, with some failing samples. \textbf{(d)} Discrete (quantized) conditioning with dropout, the most severe loss of information, leads to the most failure. Shaded areas indicate standard deviation across three different seeds. We provide additional results in \cref{appendix:conditioning_results}.
    The \textcolor{lightblue}{blue} curves show performance on training data, \textcolor{darkpink}{pink} curves depict level-1 compositions, and \textcolor{red}{red} curve denotes level-2.
    }
    \label{fig:conditioning_results}
\end{figure*}

We also investigate how different forms of conditioning affect compositional generalization (\textbf{RQ3}). This question is related to previous findings in literature that scale --- or rather diverse data --- foster compositional generalization. In our previous experiments, we conditioned models on complete, continuous factor representations (e.g., raw hue or size). In more realistic scenarios, however, factors are often quantized (e.g., the word ``red'' instead of the exact hue) or provided as lossy description (i.e., some factors are missing in the conditioning signal during training). The latter is particularly problematic, since the factors must then be inferred from the images (unsupervised identification).

For quantized signals, we convert continuous signals into discrete binary signals. For lossy signals, we randomly drop each factor with a 10\% probability, so that, on average, one in ten factors is missing.
\cref{fig:conditiong_discrete} shows that compositional generalization becomes less stable under quantized conditioning. When some factors present in the image are occasionally missing from the conditioning signal, compositional generalization typically fails (\cref{fig:conditiong_cont_dropout}). Combining both conditions--a setting common in practice--produces the strongest negative effect (\cref{fig:conditiong_mechanism_dropout}).

Overall, these results show that limited information--whether through quantization or incomplete conditioning-- can impair compositional generalization, even when all factors are provided during inference. We observe the same behavior when factors are specified through text prompts (\cref{app:text_conditioning}).
This is in line with findings on the importance of scale/diversity for compositional generalization~\citep{uselis2025does,kempf2025clip,redhardt2025scaling}, as only diverse data with overlapping conditional distributions allow unsupervised identification of the relevant factors.

\begin{tcolorbox}[colback=blue!4, colframe=black, coltitle=black, boxrule=0.3mm]
    \textbf{Finding 3:} Full, precise factor labeling during training is important for robust compositional generalization. Models trained with quantized or incomplete signals (concept dropout)  lead to weaker and inconsistent recombination of factors.
\end{tcolorbox}

\section{Enhancing Compositional Learning with Joint Embedding Predictive Architectures}

In Section~\ref{sec:factors}, we identified a key factor that can substantially weaken compositional generalization in modern generative models: training objectives that operate over discrete, categorical distributions.
Despite this, discrete training objectives--such as the one used in MaskGIT--offer advantages in, e.g., sampling speed compared to alternatives like DiT. This raises the question: can we retain these advantages while also improving compositional generalization?

\begin{figure*}[t]
    \centering
    \includegraphics[width=0.7\textwidth,trim={0 0 20 20},clip]{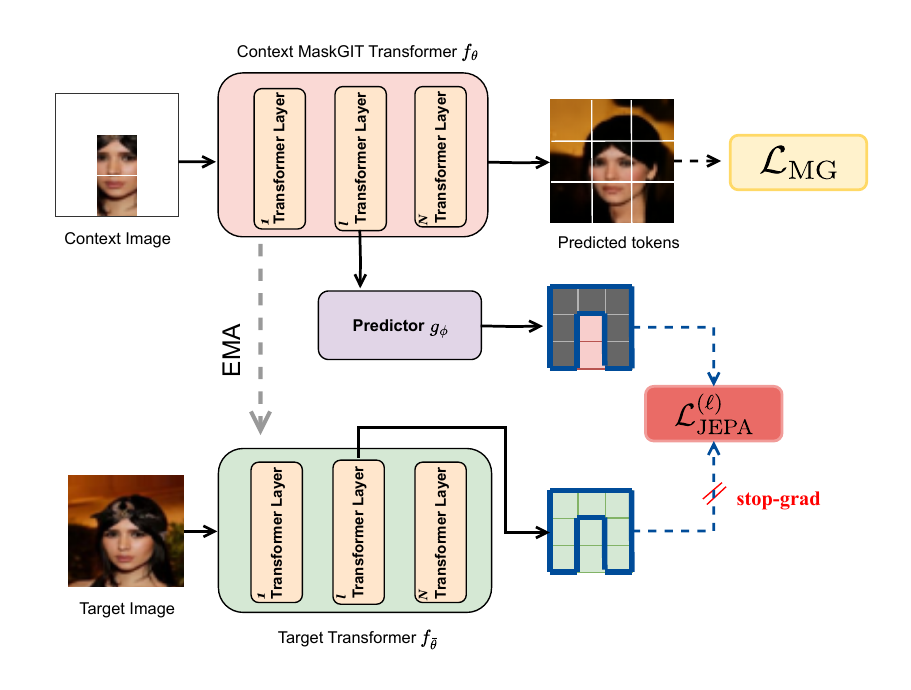}
    \vspace{-5pt}
    \caption{\textbf{An overview of MaskGIT combined with the JEPA-based training objective.} We apply the JEPA loss at specific layers ($l$) on an intermediate masked token representation in the transformer $(H_{C}^{(l)})$ and train a lightweight predictor to reconstruct target states $(H_{T}^{(l)})$ using MSE as an error metric and a stop-gradient signal to avoid representation collapse.
    }
    \vspace{-1.5pt}
    \label{fig:jepa_arch}
\end{figure*}

\subsection{An Auxiliary Continuous Relaxation Objective for MaskGIT}

Building on our previous findings, we extend MaskGIT's training objective with an auxiliary continuous representational loss, similar to Joint Embedding Predictive Architecture (JEPA) \citep{lecun2022jepa,bardes2024vjepa}. JEPA learns to reconstruct target patch representations by using (other) context patches from the same image. This auxiliary objective resembles MaskGIT’s own masking-based formulation but operates directly on continuous latent representations rather than discrete tokens.

Formally, following MaskGIT, we encode the input image or video $x$ into the discrete tokens $z=\{1 \dots K\}^N$ using a VQ-VAE. MaskGIT is then trained to reconstruct a masked subset of tokens $z_{\mathcal{M}}$ from the unmasked tokens $z_{\bar{\mathcal{M}}}$ ($\mathcal{L}_{\text{MG}}$). For images, masking is applied spatially, whereas for videos, it is applied causally in time.
Now, we add the auxiliary representation alignment objective based on JEPA ($\mathcal{L}_{\text{JEPA}}$). Given the context latent representations $H_C^{(l)}$ from selected intermediate layers $l$, the model is trained to reconstruct the target latent representations $H_T^{(l)}$. We use Mean Squared Error (MSE) with a stop-gradient on the target representations from an EMA version of the model to stabilize training. In the final objective, we jointly optimize $\mathcal{L}_{\text{MG}}$ and $\mathcal{L}_{\text{JEPA}}$. More information on the objective and results on video tasks can be found in Appendix~\cref{appendix:jepa}.

While this auxiliary loss resembles the recent REPresentation Alignment (REPA) framework \citep{yu2025repa}, REPA aligns generative latents with an external pretrained encoder. In contrast, the JEPA-based objective is completely self-supervised, structuring the model’s own intermediate representations with a continuous loss to support our findings and, in turn, compositional generalization.

\subsection{Results and Analysis}

\begin{figure*}[t]
    \centering
            \begin{subfigure}[b]{0.24\textwidth}
                \centering
                \includegraphics[width=\textwidth]{figures/maskgit.pdf}
                \caption{Standard Objective}
                \label{fig:maskgit_img}
            \end{subfigure}
            \hfill
            \begin{subfigure}[b]{0.24\textwidth}
                \centering
                \includegraphics[width=\textwidth]
                {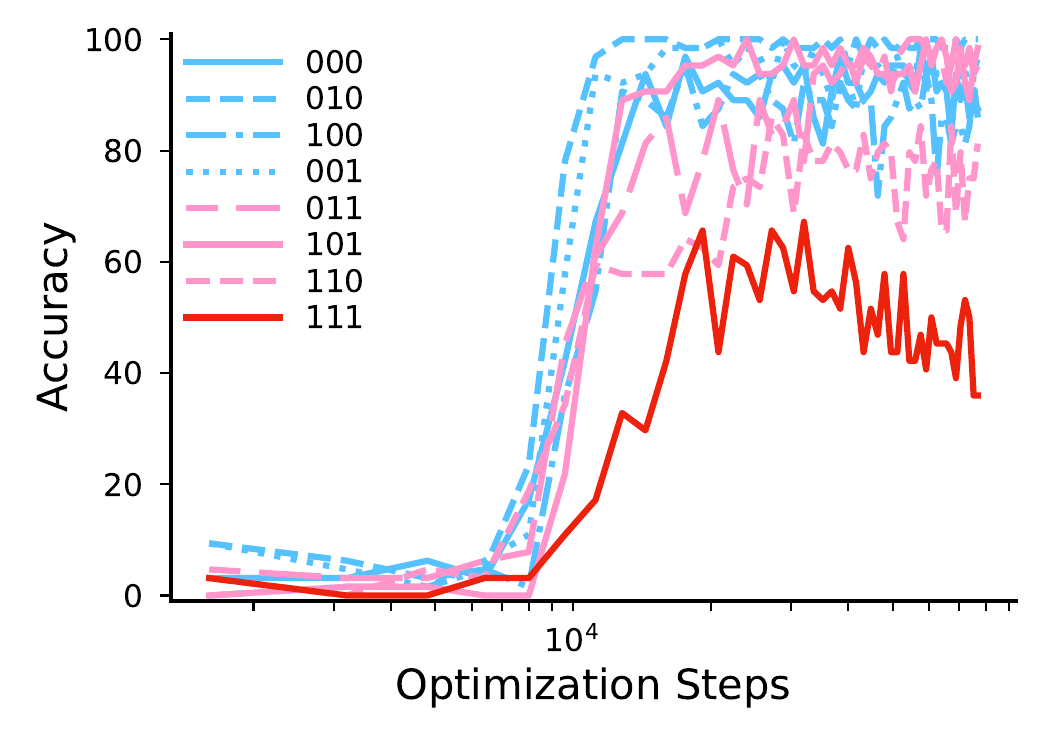}
                \caption{JEPA-based Objective}
                \label{fig:maskgit_img_jepa}
            \end{subfigure}
            \hfill
            \begin{subfigure}[b]{0.24\textwidth}
                \centering
                \includegraphics[width=\textwidth]{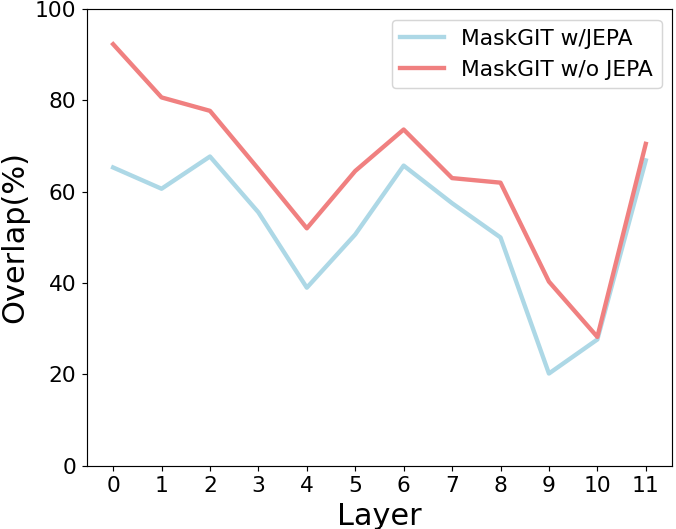}
                \caption{Polysemanticity Trend}
                \label{fig:polysemanticity_images}
            \end{subfigure}
            \hfill
            \begin{subfigure}[b]{0.24\textwidth}
                \centering
                \includegraphics[width=\textwidth]{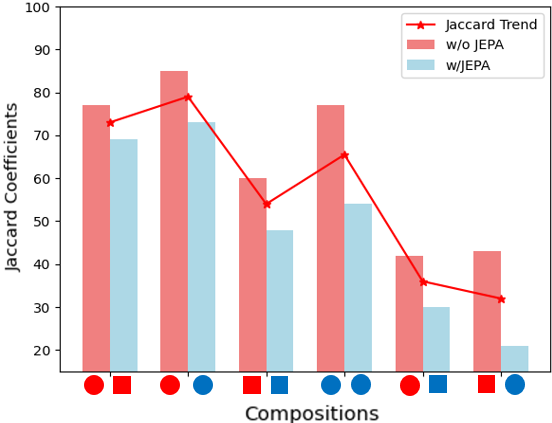}
                \caption{Jaccard Trend}
                \label{fig:jaccard}
            \end{subfigure}
    \caption{Comparison of linear probe accuracy for MaskGIT with (\subref{fig:maskgit_img}) Standard and (\subref{fig:maskgit_img_jepa}) JEPA-based training objectives for Shapes2D. The JEPA-based objective clearly enhances compositional abilities even though it cannot fully compensate the problems introduced by the discrete space. We also see lower polysemanticity between attention heads (\subref{fig:polysemanticity_images}) and a decreasing Jaccard trend over shared circuits (\subref{fig:jaccard}).
    The \textcolor{lightblue}{blue} curves show performance on training data, \textcolor{darkpink}{pink} curves depict level-1 compositions, and \textcolor{red}{red} curve denotes level-2 compositions.
    Consistent videos results shown in \cref{fig:jepa_video_results_clevrer}.
    }
    \label{fig:jepa_results}
\end{figure*}

Extending MaskGIT with the auxiliary objective improves compositional generalization
(\cref{fig:celeba_teaser,fig:maskgit_img_jepa}),
particularly for level-2 compositions. In contrast, without the auxiliary loss, MaskGIT exhibits substantially weaker compositional generalization (\cref{fig:maskgit_img}).
These results reinforce our earlier finding that having an objective operating on continuous outputs (or, by extension, intermediate latent representations) strongly supports compositional generalization. To better understand the effect of our auxiliary loss, we analyze its effect on the learned representations below.
We apply mechanistic interpretability techniques \citep{bereska2024mechanistic}, which allow us to probe how individual components encode specific factors during generation.

\paragraph{Polysemanticity in attention heads}

An attention head can easily become \textit{polysemantic}, meaning that it ``attends to'' multiple distinct and unrelated features simultaneously. This can occur due to phenomena like superposition \citep{elhage2022toy}. However, superposition, resulting in polysemanticity, causes undesired entanglement: if an attention head responsible for processing \texttt{color} also strongly activates for \texttt{shapes}, it becomes difficult for the model to disentangle these factors.

To quantify polysemanticity, we assess the effect of each attention head on the predefined visual factors/features. We pass pairs of images that differ only in a single factor (e.g., a \emph{red} large square vs. \emph{blue} large square) and compute the feature similarity for the global token. We then systematically deactivate attention heads and re-compute the similarity. A head is considered polysemantic if the feature similarity difference exceeds a threshold, suggesting that it attends to multiple, entangled factors that may hinder compositional control. \Cref{fig:polysemanticity_images} shows that fewer attention heads are polysemantic in MaskGIT with the auxiliary continuous loss suggesting less entangled factors. %

\paragraph{Mechanistic similarity using circuits}

The above analysis shows that attention heads mix up some features, but it only provides a coarse view of representational entanglement. To probe deeper into how concepts are internally represented and transformed across layers, we analyze the model's \textit{internal circuitry} \citep{olah2020zoom}-groups of neurons that jointly implement specific mechanisms.

We aim to identify the most influential components that generate each composition. To do so, we focus on MLP neurons within each transformer block, since prior work has shown that these often encode semantic factors \citep{geva2020transformer,dai2021knowledge}.
For each composition, we identify the top-$k$ most influential neurons per layer using activation patching \citep{vig2020investigating,meng2023locating}. Specifically, we compute each neuron's \textit{indirect effect} \citep{pearl2022direct}, which captures its downstream influence on the model's output rather than just its direct contribution. This is achieved by comparing the output when running the model on a factual input (the correct composition condition, e.g., ``red circle'') with the output when running on a counterfactual input (the same condition but with a key intermediate activation corrupted, e.g., replaced with noise), and then patching in the neuron's clean activation. Intuitively, this measures how strongly a neuron shapes the generated image or video.

We then compare the overlap of the top-$k$ neurons across different compositions using the Jaccard index, following \citet{kempf2025clip}. A high overlap implies that the same neuron is used for multiple factors, suggesting that circuits are more entangled. In contrast, a low overlap suggests separate circuits. (More in \cref{appendix:circuits}). \Cref{fig:jaccard} shows that adding the auxiliary JEPA objective consistently reduces neuron overlap across layers, particularly for composition pairs where both factors differ (e.g., \texttt{red circle-blue square}), suggesting more factor-specific circuits.

\begin{tcolorbox}[colback=blue!4, colframe=black, coltitle=black, boxrule=0.3mm]
    \textbf{Finding 4:} A JEPA-based training objective induces more disentangled and semantically structured representations, and improves compositional generalization.
\end{tcolorbox}

\section{Discussion and Limitations}

\paragraph{Do our results extend to world models?}
Our previous experiments focused on academic datasets to enable controlled experimentation. To test whether the observed trend extends to a more realistic setting, we curate compositional splits on CoVLA \citep{arai2025covla}, a driving scenes video dataset. The factors are \emph{time of day} (day \daysym /night \nightsym) and \emph{turn direction} (left \Ldir/straight \Sdir/right \Rdir). We hold out \fact{\daysym}{\Rdir} and \fact{\nightsym}{\Ldir}. As our generative video model, we adopt state-of-the-art Orbis \citep{mousakhan2025orbis} conditioned only via the previous frames. Since Orbis does not know the explicit factors, it must thus infer them via unsupervised identification. We train the two Orbis variants, Orbis-DiT (continuous) and Orbis-MaskGIT (discrete), with identical training and inference budgets. We measure compositionality as the fraction of nearest neighbors sharing the conditioning factors (\cref{appendix:covla_eval}).

\cref{tab:compgen_retrieval_pretty} shows that DiT outperforms MaskGIT on \emph{novel} compositions, with relative gains of \textbf{\gain{+161\%}} on \fact{\daysym}{\Rdir} and \textbf{\gain{+207\%}} on \fact{\nightsym}{\Ldir}. Per-split nearest-neighbor ratios further suggest that MaskGIT collapses to seen combinations, while DiT assigns more mass to the intended novel target (see~\cref{fig:orbis-wide}). Although absolute accuracies remain modest---reflecting the difficulty of real-world video with implicit contextual factors---the trend matches our controlled interventions in \cref{find:cont-vs-cat}: \emph{continuous objectives improve compositionality under complete conditioning relative to categorical objectives}.

\paragraph{Do our results extend to language?}

Although our study focuses on \textit{visual} generative models, compositional generalization has been central in language modeling~\citep{furuta2023language,sakai2025revisiting}. Does the benefit of continuous objectives transfer to language?

We study compositional generalization for Llama-3.2 \citep{llama} on a controlled card-arithmetic task inspired by the Points24 dataset \citep{points24}. In this task, the model is given four cards and a target value, and must construct an arithmetic expression that uses each card once to reach the specified target number. Training involves single rules (restricting the target value or doubling the value of red-suit cards), while the (compositional) test split requires composing both rules simultaneously (details in \cref{appendix:language}). We compare two reasoning mechanisms: standard Chain-of-Thought (CoT) \citep{cot} and its continuous variant, COCONUT \citep{coconut}. Here, reasoning mechanisms play a role analogous to training objectives in our experiments: whether intermediate tokens are treated as discrete symbolic traces (CoT) or continuous thoughts (COCONUT).
COCONUT outperforms CoT on the compositional splits (12.33\% vs.\ 4.87\%), suggesting that the advantage of continuous objectives may carry over to language.

\paragraph{Limitations}
In comparison to previous works on diffusion models generalization, our work aimed to move beyond toy data while retaining controlled compositional factors. This remains a fundamental limitation for studying web-scale models, where the relevant factors are unknown or entangled.

While our evidence shows that continuous objectives matter for compositional generalization, we can only speculate about the mechanism. We hypothesize that more continuous loss gradients make it easier to establish compositional factors as separate components in the representation, whereas gradients that strongly favor correct categorical decisions can lose them between the output, input and intermediate representations, even when they are given explicitly. Current mechanistic analysis tools are not yet crisp enough to verify this hypothesis.

Our findings may also have implications for LLMs, which are typically trained with categorical objectives. We provide preliminary evidence of similar effects in language models, but controlled evaluation of compositional generalization in language is non-trivial and requires dedicated study.

\begin{ack}

This work was funded by the German Federal Ministry for Economic Affairs and Energy within the project ``NXT GEN AI METHODS'' (19A23014R). The compute used in this project was also funded by the German Research Foundation (DFG) under grants 417962828 and 539134284. Karim Farid and Rajat Sahay are part of the European Lab for Learning and Intelligent Systems (ELLIS) PhD program. This work was funded in part by the French government under management of Agence Nationale de la Recherche as part of the “France 2030” program, reference ANR-23-IACL-0008(PR[AI]RIE-PSAI project) and the ANR project VideoPredict ANR-21- FAI1-0002- 01. Cordelia Schmid would like to acknowledge the support by the K\"orber European Science Prize. The authors acknowledge support from the state of Baden\-W\"urttemberg through bwHPC. We thank Arian Mousakhan, Silvio Galesso, Sudhanshu Mittal, Johannes Dienert, and Jelena Bratulic for valuable discussions and feedback during the development of this work.
\end{ack}

\enlargethispage{\baselineskip}

\bibliographystyle{plainnat}
\bibliography{ref}

\newpage
\appendix

\section{Models}
\label{appendix:models}

We selected a representative set of generative models to span key design axes relevant to compositional generalization. \textbf{Diffusion Transformers (DiTs)} \citep{peebles2023dit} operate in a continuous latent space derived from a VAE and are trained with a standard denoising objective; conditioning is applied via adaptive layer normalization. In contrast, \textbf{MaskGIT} \citep{chang2022maskgit} works in a discrete latent space obtained from a VQ-VAE and employs a bidirectional transformer to iteratively predict masked tokens, corresponding to a categorical, masking-based objective. \textbf{MAR} \citep{li2024mar} also uses a continuous latent space but masks subsets of latent tokens and leverages an encoder-decoder to generate intermediate conditions for a diffusion head that denoises each masked token independently, combining aspects of masking and denoising. Together, these models cover continuous versus discrete representations, denoising versus masking objectives, and a variety of conditioning strategies.

\subsection{Diffusion Transformer}

To evaluate models in continuous representation spaces, we train \textbf{Diffusion Transformers} (DiTs) \cite{peebles2023dit} which adapt the Transformer architecture \cite{vaswani2017attention} within the Latent Diffusion Model framework \cite{nichol2021diffusion}. Initially, a pretrained VAE encodes an input image $x$ into a compressed latent representation $z=E(x)$ which is processed into patch embeddings $s$. A standard forward diffusion process then progressively adds Gaussian noise to this latent representation $s$ over $t$ timesteps. The core of the model is a Transformer network trained to denoise the noisy latent $s_{t}$ by predicting the added noise $\epsilon_{t}$ at each  timestep $t$, typically minimizing $L=\mathbb{E}_{s_{0},\epsilon,t,c}\left[\|\epsilon-\epsilon_{\theta}(s_{t},t,c)\|_{2}^{2}\right]$, where $\mathbb{E}$ denotes the expectation over the initial latent representation, $\epsilon_{\theta}$ is the noise predicted by the Transformer network with parameters $\theta$, and $\|\circ\|_{2}^{2}$ denotes the squared L2 norm (Euclidean norm). We implement conditional generation by integrating guidance information $c$ into the Transformer blocks using adaptive layer normalization (adaLN), which modulates network activations based on $c$ and $t$.

\subsection{MAR}

\textbf{MAR} \cite{li2024mar} explores generation within a continuous latent space defined by a VAE while preserving autoregressive principles- iteratively generating a set of tokens conditioned on previously generated tokens. It consists of an encoder, a decoder, and a diffusion head, with both encoder and decoder consisting of a stack of self-attention blocks.
Specifically, MAR employs a pretrained VAE to encode an image into a latent representation $\mathcal{Z} \in \mathbb{R}^{h \times w \times d}$ while randomly masking a subset of image tokens within $\mathcal{Z}$. Let $\mathcal{U}=\{{z_{i_{\backslash M}}}\}_{i=1}^{{N_{\backslash M}}}$ and $\mathcal{M}=\{{z_{i}}\}_{i=1}^{{N}}$ represent the set of unmasked and masked tokens, respectively, where each ${z_{\backslash M}}$ and $z$ corresponds to a spatial element in $Z$. Here, ${N_{\backslash M}}$ denotes the number of unmasked tokens, while $N$ represents the number of masked tokens. The encoder processes $\mathcal{U}$ to extract latent features. The decoder then takes as input both these latent features and a set of learnable tokens of size $N$, each corresponding to a masked token in $\mathcal{M}$. It subsequently generates the features of these learnable tokens, referred to as conditions. Finally, the diffusion head independently performs a denoising process on each masked token $z_{i}$ using its corresponding condition.

\subsection{MaskGIT}

\textbf{MaskGIT} or Masked Generative Image Transformer \cite{chang2022maskgit} is an autoregressive model operating in a discrete latent space.
An input data sample $x$ is first encoded into a sequence of discrete visual tokens $E_{VQ}(x) \in \{1,...,K\}^{N}$ where $E_{VQ}$ is the encoder trained with a VQ-VAE objective and $N$ is the sequence length. The standard MaskGIT objective $(\mathcal{L}_{MG})$ focuses on reconstructing masked input tokens.
We employ block-masking \cite{bardes2024vjepa} for images and causal block masking for videos \cite{bardes2024vjepa} to sample a mask $\mathcal{M} \subset \{1,...,N\}$. The MaskGIT Transformer $T$ predicts the probability distribution $p(z_{i}|z_{\backslash M})$ for each masked token $z_{i}~(i \in M)$ based on unmasked token $z_{\backslash M}$. The loss is the negative log-likelihood of the true tokens at the masked positions
$\mathcal{L}_{MG} = -\mathbb{E}_{x,M} \left[\sum_{i \in M}\log p(z_{i}|z_{\backslash M}) \right]$.

MaskGIT then learns a bidirectional transformer model $p_{\theta}(z|c)$ trained via the masking procedure, predicting masked tokens based on unmasked one and conditioning $c$. Generation is performed iteratively, starting with all tokens masked and progressively predicting and committing to high-confidence tokens. Conditioning $c$ is embedded to the token sequence using a two-layer MLP and integrated using adaptive layer normalization (adaLN).

\section{Datasets}
\label{appendix:datasets}

Below, we provide an overview of the general characteristics and qualitative examples of the three datasets used in our paper. Results for generation from the generative models can be found in \cref{appendix:results}.

\subsection{Shapes2D}

Shapes2D is a synthetic dataset introduced by Okawa et al.~\cite{okawa} to systematically study compositional generalization in diffusion models. Each image is composed of a single geometric object characterized by three binary-valued concepts: \textit{shape}, \textit{color}, and \textit{size}, with the composition space defined as \texttt{shape=\{circle, triangle\}, color=\{red, blue\}, size=\{large, small\}}. Unless stated otherwise, these concepts are encoded as binary tuples—e.g., the tuple \texttt{000} represents a large, red circle. This structured format allows the use of concept tuples as conditioning signals during diffusion model training, mimicking abstract textual prompts. The dataset is constructed by uniformly sampling over the concept combinations, yielding 1500 images per training concept class and a total of 6000 images. Although the concept variables are nominally discrete, the \textit{size} and \textit{color}  attributes are realized through a spectrum of sizes within the big and small labels and hues within the red and blue labels, resulting in tightly clustered yet visually diverse instantiations. These clusters are sparsely distributed and exhibit limited continuity in the pixel space, effectively forming “islands” within the data manifold. Such fragmentation introduces discontinuities in the data-generating factors space that further complicate the model’s ability to learn smooth conceptual compositional interpolations.

\subsection{Shapes3D}

Shapes3D \citep{shapes3d} is a synthetic dataset of rendered 3D scenes designed to study disentanglement and compositional generalization. Each image depicts a single object in a colored room, generated by varying six independent factors: floor color (10 values), wall color (10), object color (10), object shape (4), object size (8), and camera azimuth (15). The full dataset contains 480,000 images, corresponding to all possible combinations of these factors. This makes it a significantly more complex data space than Shapes2D and its fully factorial structure makes Shapes3D well-suited for systematic generalization experiments.

For our experiments, we use all 10 object colors, three of the four shapes (cube, cylinder, sphere), and all eight size values (0–7), yielding 240 compositions.

The model is conditioned on the specific colour, shape, and size values. For compositional evaluation, we group the 240 compositions into 8 supergroups to structure the train–test split and to order results by degree of novelty. As in Shapes2D, we define three levels of novelty: first level includes compositions seen during training. second and third levels are used for testing and differ from the training set in one and two factors, respectively, with level three representing the highest degree of novelty.

We partition the factors of color, shape, and size into two supergroups based on predefined thresholds. Specifically, for the color factor, which consists of ten categories, Supergroup 0 includes the first seven colors (red, orange, yellow, lime green, green, cyan, blue), and Supergroup 1 includes the remaining three (indigo, purple, pink). For shape, which contains three categories, Supergroup 0 includes cube and cylinder, and Supergroup 1 includes sphere. Finally, for size, which is defined over ordinal values from 0 to 7, values 0 through 5 belong to Supergroup 0, while values 6 and 7 belong to Supergroup 1.

\subsection{CelebA}

To assess the robustness of our findings under real-world settings, we extend our evaluation to the CelebA dataset. Following prior work, we select three visually distinguishable attributes as concept variables: \textit{Gender}, \textit{Smiling}, and \textit{Hair Color}, with the composition space defined as \texttt{Gender=\{Male, Female\}, Smiling=\{Smiling, Not Smiling\}, Hair Color=\{Black Hair, Blonde Hair\}}. Conditioning concepts in CelebA are quantized with less information about the degree of smiling or hair color, which results in harder convergence for novel concept generation. These attributes are treated as binary concept tuples similar to the Shapes2D setting, enabling consistent evaluation across synthetic and real-world domains. We curate 10,000 images per concept class, yielding a total of 40,000 training examples. This setup allows us to examine how well models trained under structured compositional constraints in CelebA can generalize to novel combinations of facial attributes.

\subsection{CLEVRER-Kubric}

The CLEVRER-Kubric dataset is a reimplementation of CLEVRER \cite{yi2020clevrer} from scratch using Kubric \cite{greff2022kubric}. This dataset is generated by simulating physics-based scenes using Blender. The scenes are configured with parameters similar to CLEVRER, including a resolution of $480 \times 320$ pixels, a frame rate of 12 fps, and a simulation step rate of 240Hz. Each scene includes a static floor with a gray material and a directional light source. A perspective camera is positioned to view the scene. Stationary objects, numbering between 5 and 7, are randomly chosen from cubes or spheres, and are placed within a pre-defined spawn region. The material properties for individual objects are controlled based on the composition being tested (that is, all instances of a specific \texttt{color} or \texttt{shape} can be held out to ensure that the model does not see a specific combination during training).

The dynamic aspect involves 1 to 3 projectiles, also from the pre-defined shapes, with similar controlled material and size properties. The projectiles are launched from random points around the perimeter with a velocity between 10 and 15 units, slight randomness to the trajectory, and zero angular velocity. There is increased angular damping to maintain consistency with the original CLEVRER videos. Each composition has 250 videos, each with 28 frames, yielding 7000 frames per composition.

As with Shapes2D, we control for \texttt{shape=\{cube, sphere\}, size=\{large, small\}, color=\{red, green\}}. For every composition, we put the desired composition as one (or more) of the static objects to force the model to generate that (and not learn a shortcut by simply generating other parts of the video). We increase the probability of having the dynamic moving object also follow the same properties as the composition to ensure that our methods also learn to model temporal consistency for compositions that are out of the training distribution.

\section{Experiments}
\label{appendix:experiments}

\subsection{Tokenizer}
\label{appendix:tokenizer_results}

\begin{figure}[t]
    \centering
    \resizebox{\textwidth}{!}{
        \begin{minipage}{\textwidth}
            \begin{subfigure}[b]{0.3\textwidth}
                \centering
                \includegraphics[width=\textwidth]{figures/mar_token_final.pdf}
            \caption{VQ-VAE}
            \label{fig:mar_token}
            \end{subfigure}
            \hfill
            \begin{subfigure}[b]{0.3\textwidth}
                \centering
        \includegraphics[width=\textwidth]{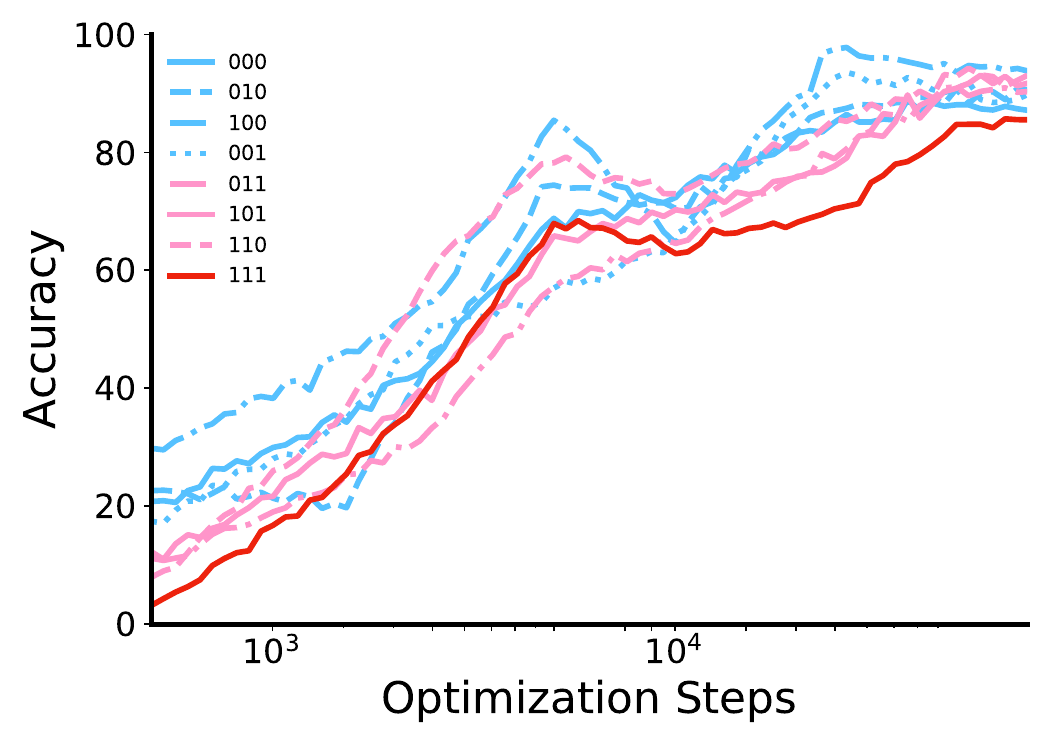}
                \caption{VAE}
            \label{fig:mar_vae}
            \end{subfigure}
            \hfill
            \begin{subfigure}[b]{0.3\textwidth}
                \centering
           \includegraphics[width=\textwidth]{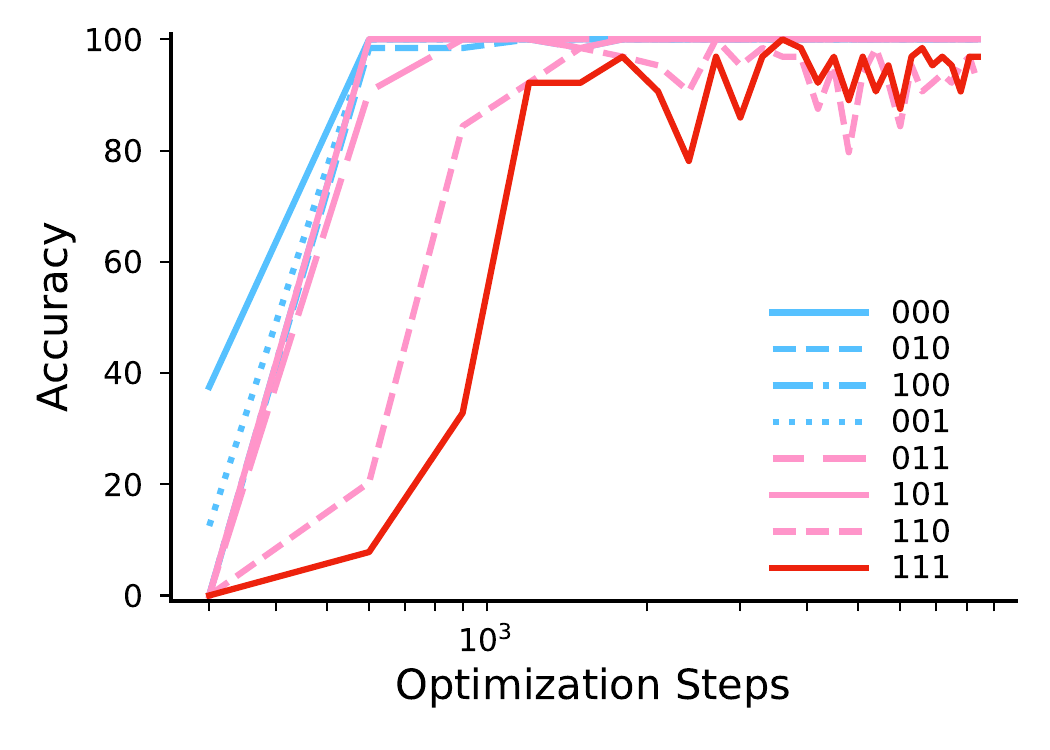}
                \caption{Pixel (No tokenizer)}
     \label{fig:mar_pixel}
            \end{subfigure}
        \end{minipage}
    }
        \caption{MAR exhibits generalization on both level-1 and level-2 compositions in Shapes2D regardless of whether (a) discrete, (b) continuous, or (c) no tokenizer is used.
        The \textcolor{lightblue}{blue} curves show performance on training data, \textcolor{darkpink}{pink} curves depict level-1 compositions, and \textcolor{red}{red} curve denotes level-2 compositions.
        }
    \label{fig:mar_tokenizer_results}
\end{figure}

\begin{figure}[t]
    \centering
    \resizebox{\textwidth}{!}{
        \begin{minipage}{\textwidth}
            \begin{subfigure}[b]{0.3\textwidth}
                \centering
                \includegraphics[width=\textwidth]{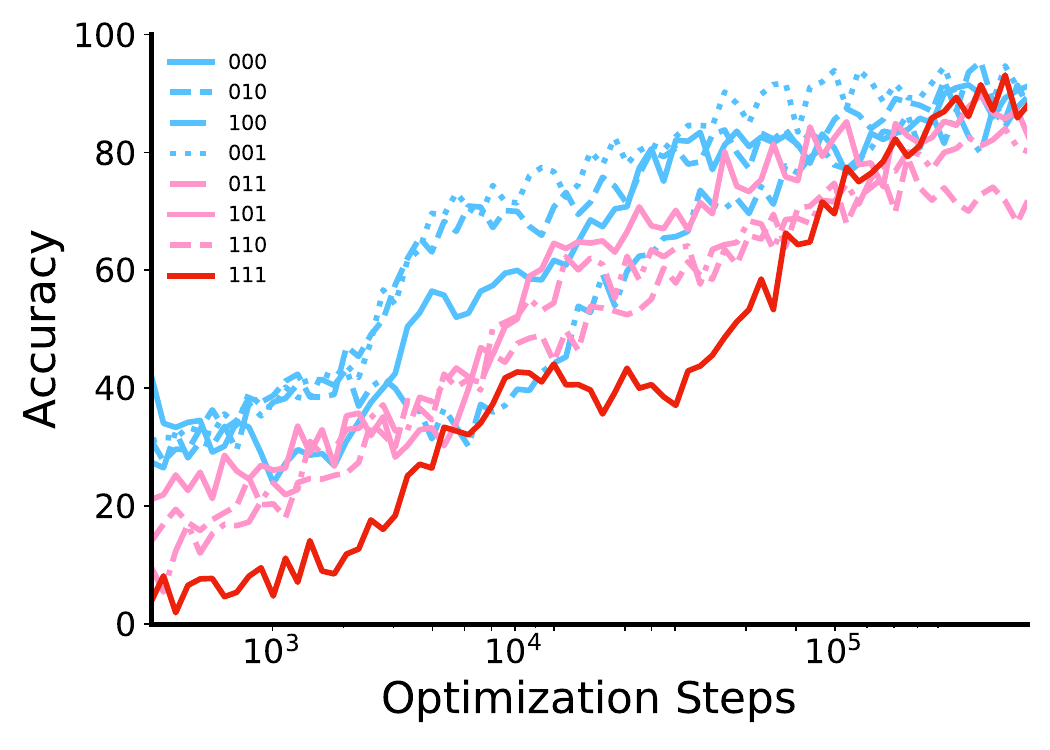}
            \caption{VQ-VAE}
            \label{fig:dit_token_clevrer}
            \end{subfigure}
            \hfill
            \begin{subfigure}[b]{0.3\textwidth}
                \centering
        \includegraphics[width=\textwidth]{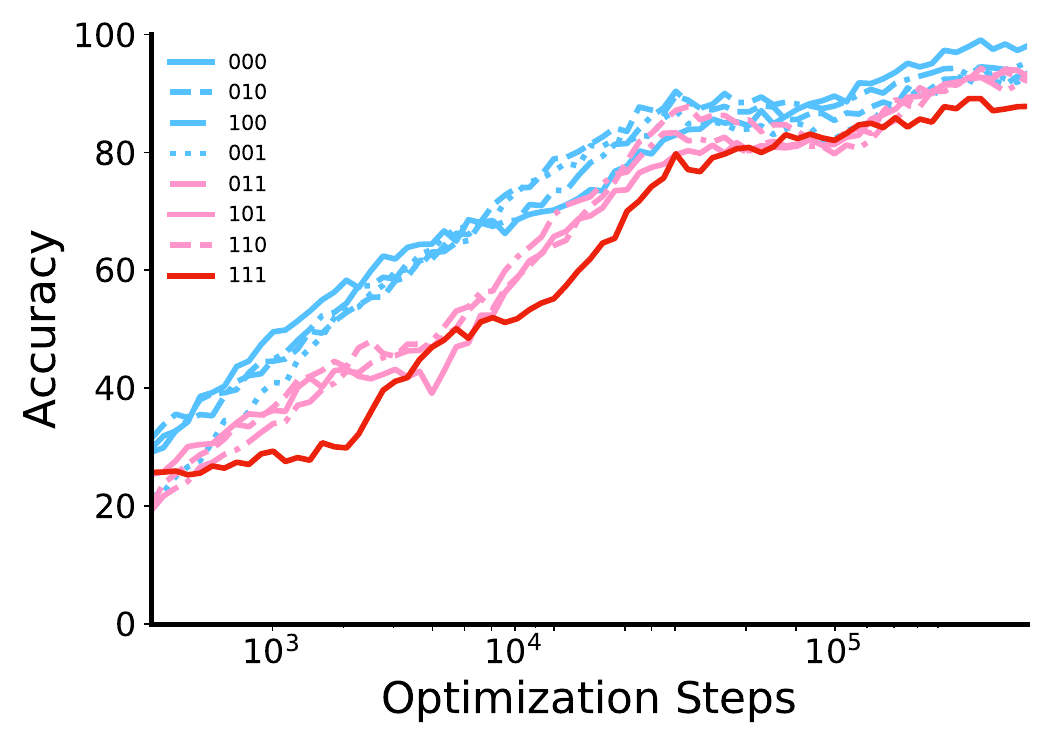}
                \caption{VAE}
            \label{fig:dit_vae_clevrer}
            \end{subfigure}
            \hfill
            \begin{subfigure}[b]{0.3\textwidth}
                \centering
           \includegraphics[width=\textwidth]{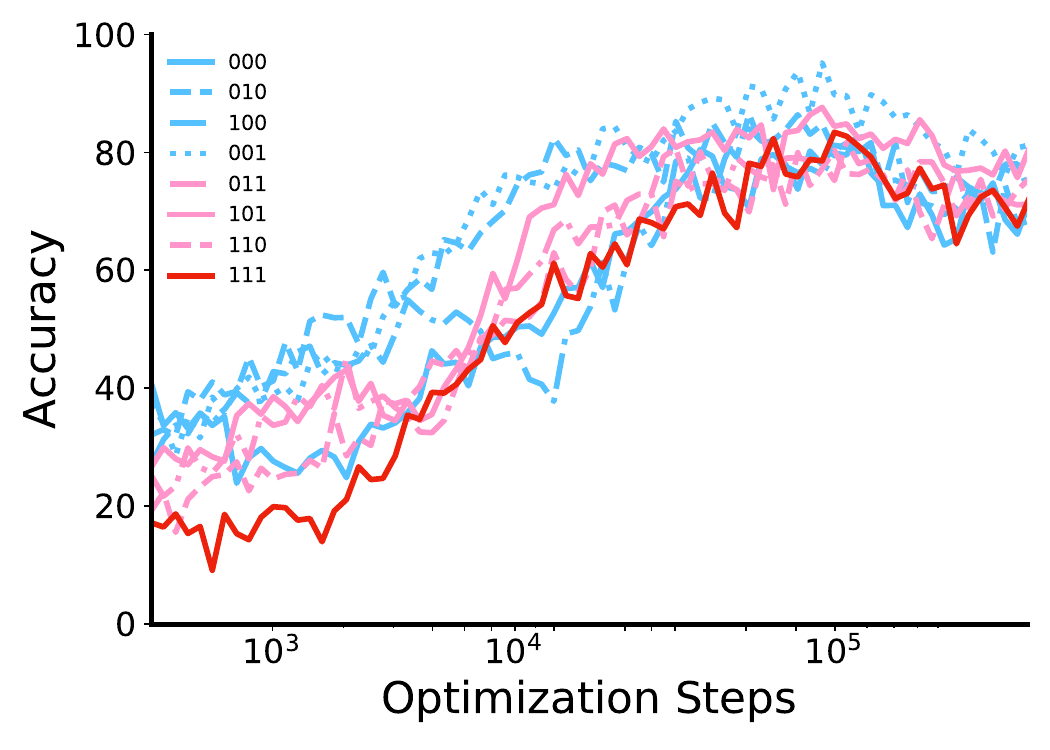}
                \caption{Pixel (No tokenizer)}
     \label{fig:dit_pixel_clevrer}
            \end{subfigure}
        \end{minipage}
    }
        \caption{DiT exhibits generalization on both level-1 and level-2 compositions in CLEVRER-Kubric regardless of whether (a) discrete, (b) continuous, or (c) no tokenizer is used.
        The \textcolor{lightblue}{blue} curves show performance on training data, \textcolor{darkpink}{pink} curves depict level-1 compositions, and \textcolor{red}{red} curve denotes level-2 compositions.
        }
    \label{fig:dit_clevrer_tokenizer_results}
\end{figure}

Similar to Section~4.1 of the main text, we also trained MAR with three different types of tokenization schemes to prove that the choice of tokenizer does not alter the compositional capabilities of a model. We show linear probe results in \cref{fig:mar_tokenizer_results}. By showing a similar trend, we complement the experiments on DiT shown in the main text, and also show that the choice of downstream generative architecture is not a confounding factor when comparing different tokenizers. To further reinforce the generality of our experiments and demonstrate that the same trend extends to videos, we also present results on CLEVRER-Kubric using a DiT model trained under three different tokenization schemes (\cref{fig:dit_clevrer_tokenizer_results}).

\paragraph{Mapping discrete tokens to continuous vectors for downstream models}
When employing VQ-VAE tokenizer with models like DiT and MAR, the process involves an intermediate step to bridge the discrete nature of VQ-VAE tokens with the continuous vector processing expected by these architectures. \emph{For our experiments in the main text, we use the continuous embedding vectors before the quantization step in the tokenizer}. However, we add a learnable embedding layer in our pipeline for our experiments on CLEVRER-Kubric. The VQ-VAE first encodes an input data sample into a spatial grid of discrete token IDs, where each ID corresponds to a learned codebook vector. For a downstream model like a transformer, these discrete token IDs are then passed through a learnable embedding layer. This layer maps each token ID to a unique, dense, continuous vector representation. It is this grid of continuous vectors (token embeddings) that then serves as the actual input to the downstream generative model. This embedding step effectively translates the discrete symbolic representation from the VQ-VAE into a continuous vector space where the model can perform its computations to achieve generation and composition.

\subsection{Training Objective}
\label{appendix:latent_space_results}

\begin{figure}[t]
    \centering
    \resizebox{\textwidth}{!}{
        \begin{minipage}{\textwidth}
            \begin{subfigure}[b]{0.3\textwidth}
                \centering
                \includegraphics[width=\textwidth]{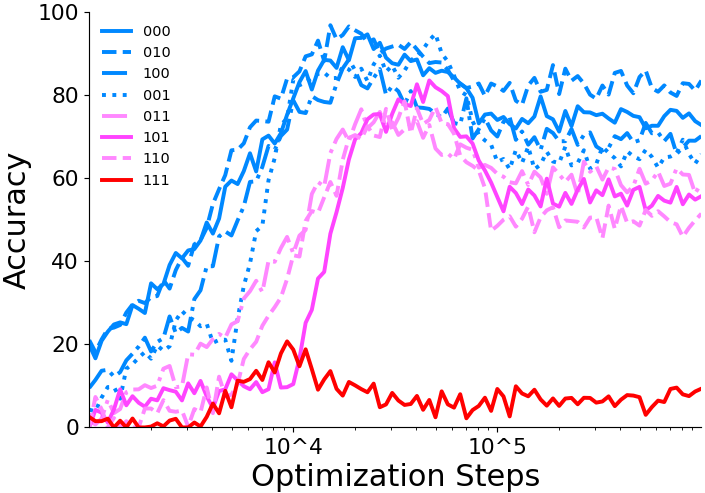}
            \caption{MaskGIT}
            \label{fig:maskgit_clevrer}
            \end{subfigure}
            \hfill
            \begin{subfigure}[b]{0.3\textwidth}
                \centering
        \includegraphics[width=\textwidth]{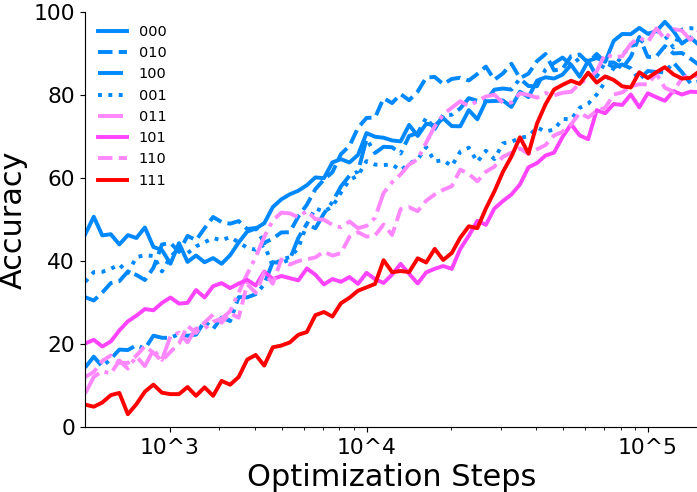}
                \caption{DiT}
            \label{fig:diffusion_clevrer}
            \end{subfigure}
            \hfill
            \begin{subfigure}[b]{0.3\textwidth}
                \centering
           \includegraphics[width=\textwidth]{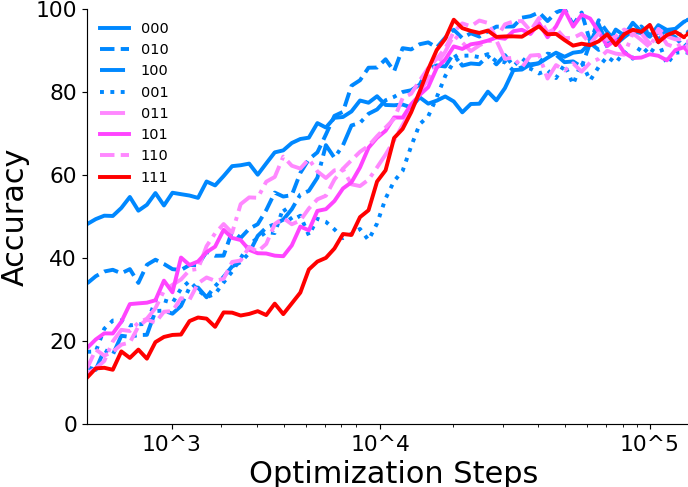}
                \caption{MAR}
     \label{fig:MAR_clevrer}
            \end{subfigure}
        \end{minipage}
    }
        \caption{Comparison of linear probe accuracy for CLEVRER-Kubric across (\subref{fig:maskgit_clevrer}) MaskGIT, (\subref{fig:diffusion_clevrer}) DiT, and (\subref{fig:MAR_clevrer}) MAR. Models leveraging a continuous latent space (DiT, and MAR) show better level-2 compositions than MaskGIT.
        The \textcolor{lightblue}{blue} curves show performance on training data, \textcolor{darkpink}{pink} curves depict level-1 compositions, and \textcolor{red}{red} curve denotes level-2 compositions.
        }
    \label{fig:latent_space_results_clevrer}
\end{figure}

Complementing results from Shapes2D \cite{okawa} and CelebA \cite{liu2018celeba} in the main text, we show that our findings also hold true for more complex modalities like video. \cref{fig:latent_space_results_clevrer} contrasts the performance of a discrete objective (MaskGIT) against a continuous objective (DiT) on the CLEVRER-Kubric dataset. We see the same trend of MaskGIT showing significantly worse performance, especially on level-2 compositions. Following the structure in the main text, we also show results on MAR (\cref{fig:MAR_clevrer}) to disentangle the effects of the architectural design from the properties of the latent space and further reinforce our results that objectives modeling underlying continuous factors achieve better compositionality.

\begin{figure}[ht]
    \centering

    \begin{minipage}{0.3\textwidth}
        \centering
        \includegraphics[width=\linewidth]{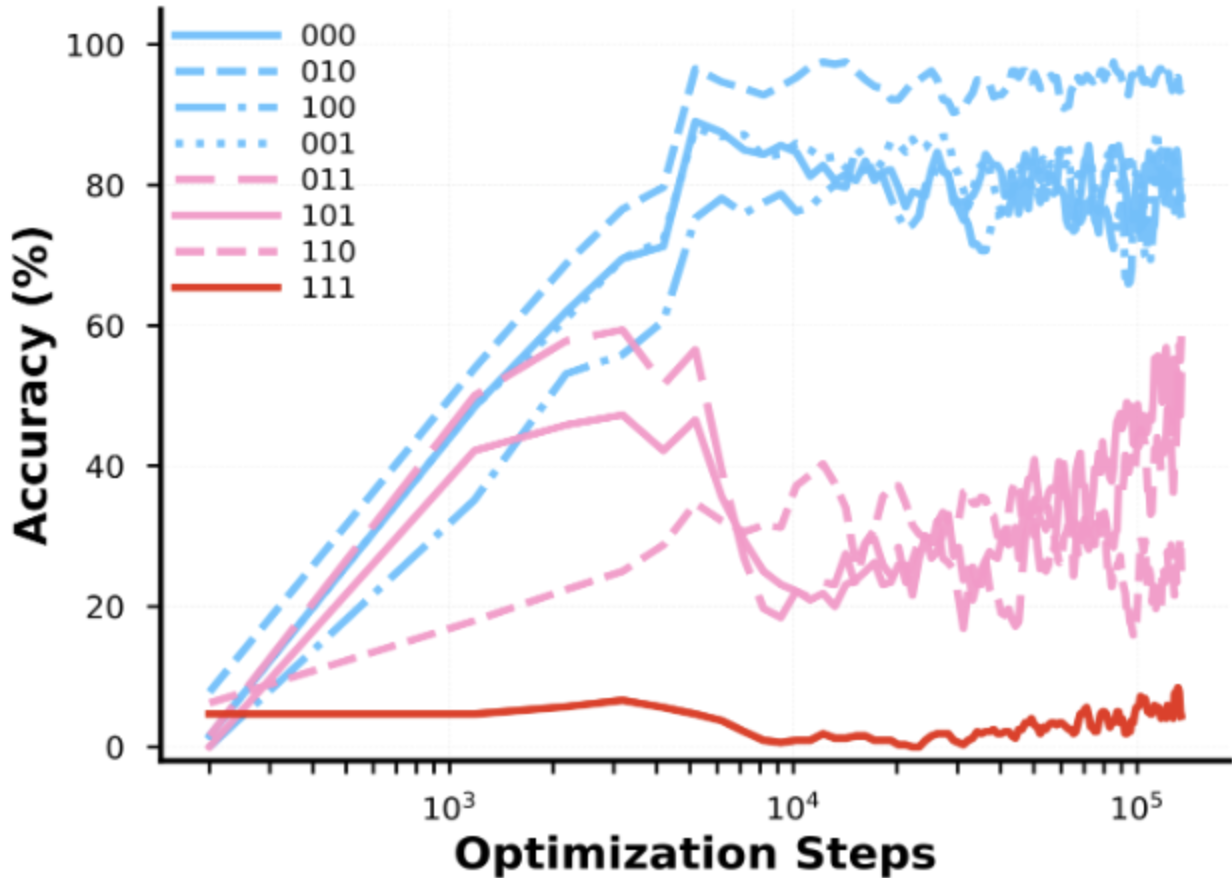}
        \caption*{(a) High Learning Rate (5e-5 $\rightarrow$ 3e-4)}
    \end{minipage}
    \hfill
    \begin{minipage}{0.3\textwidth}
        \centering
        \includegraphics[width=\linewidth]{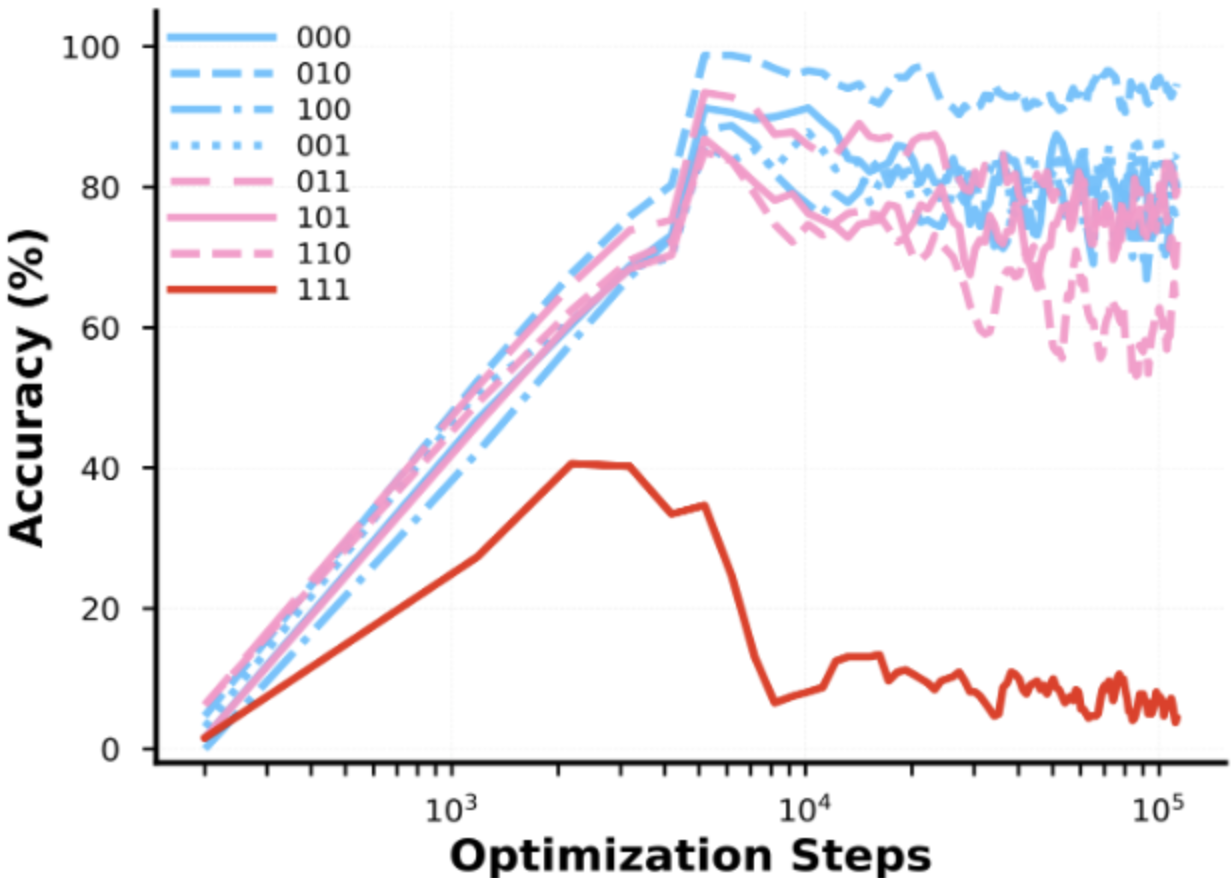}
        \caption*{(b) Large Model (12 $\rightarrow$ 18 Layers) -- Batch Size 128}
    \end{minipage}
    \hfill
    \begin{minipage}{0.3\textwidth}
        \centering
        \includegraphics[width=\linewidth]{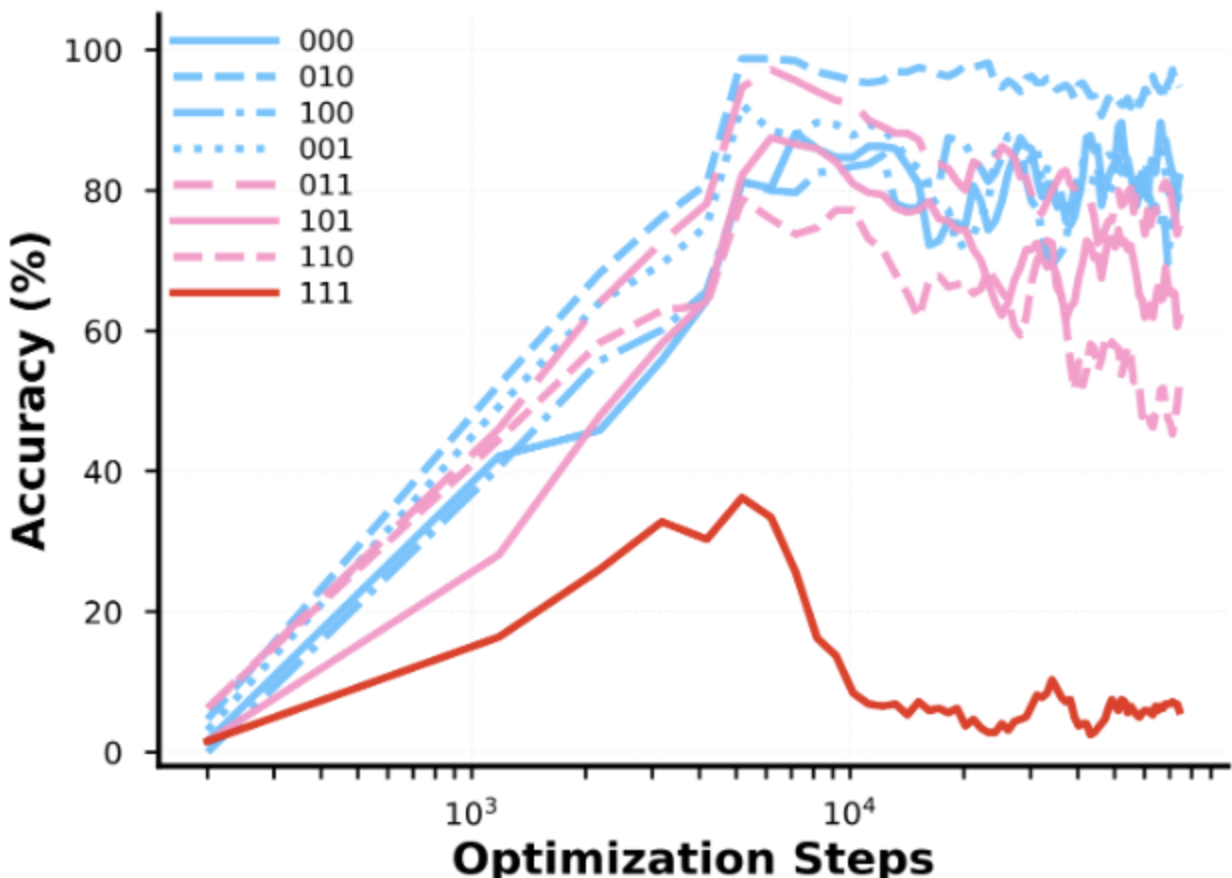}
        \caption*{(c) Large Model (12 $\rightarrow$ 18 Layers) -- Batch Size 128}
    \end{minipage}

    \vspace{0.5cm}

    \begin{minipage}{0.3\textwidth}
        \centering
        \includegraphics[width=\linewidth]{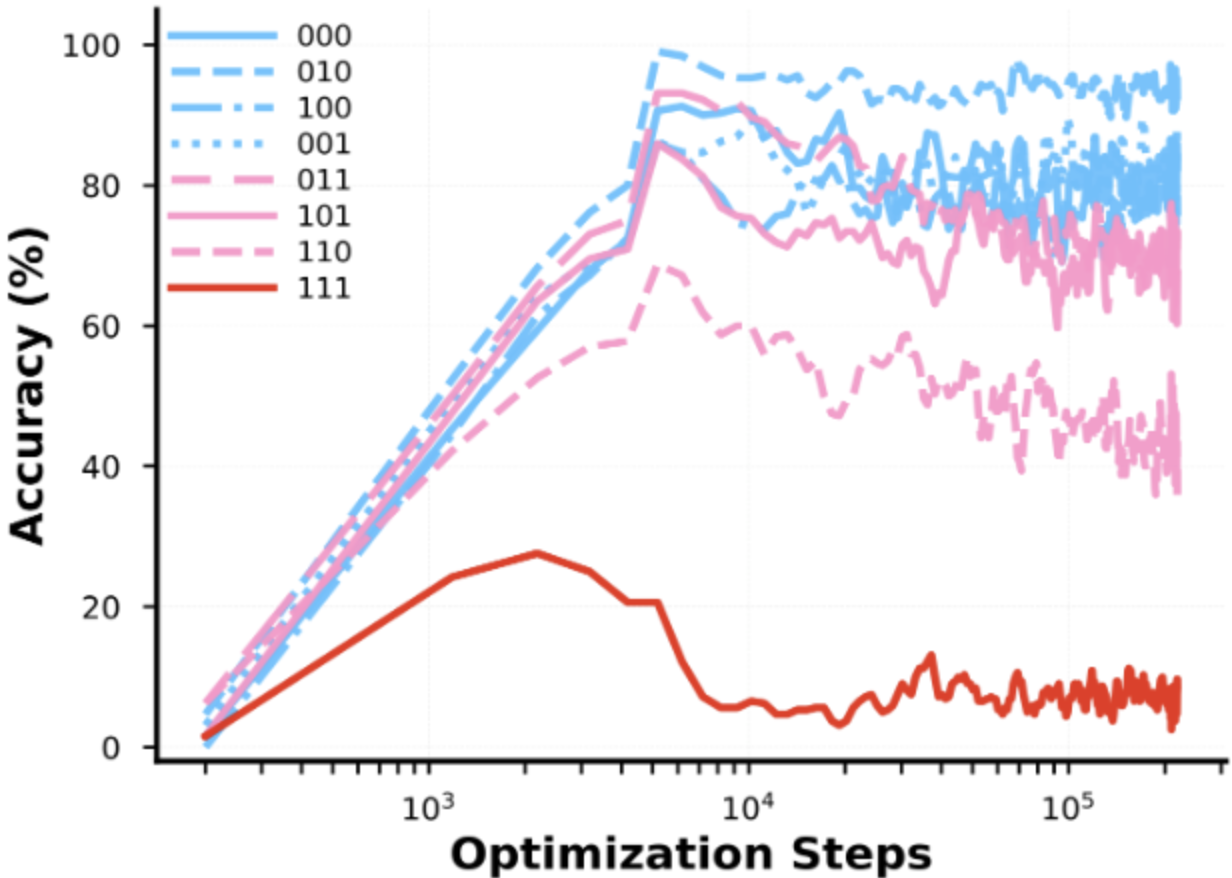}
        \caption*{(d) Long Run (5e-5, 60k $\rightarrow$ 200k steps)}
    \end{minipage}
    \hfill
    \begin{minipage}{0.3\textwidth}
        \centering
        \includegraphics[width=\linewidth]{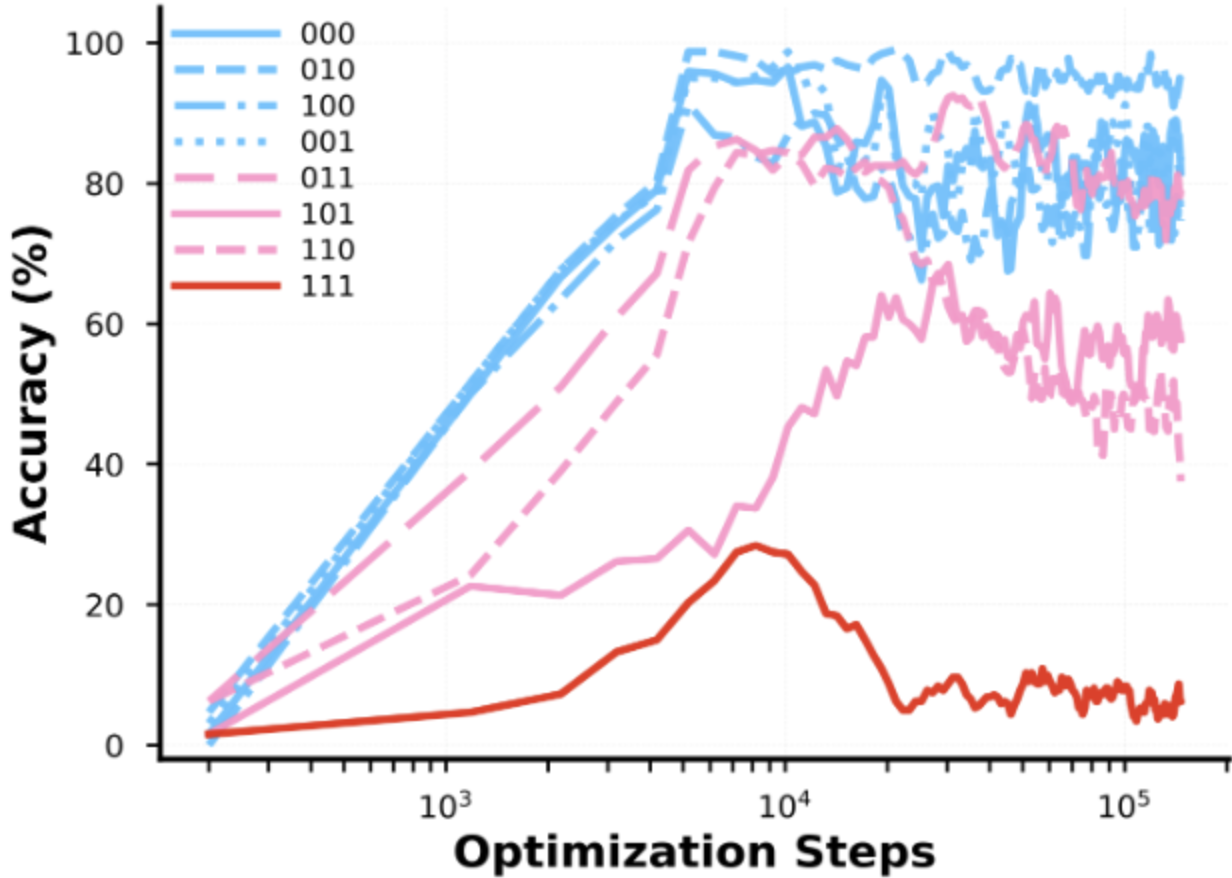}
        \caption*{(e) Small Learning Rate (5e-5 $\rightarrow$ 1e-5) -- Batch Size 128}
    \end{minipage}
    \hfill
    \begin{minipage}{0.3\textwidth}
        \centering
        \includegraphics[width=\linewidth]{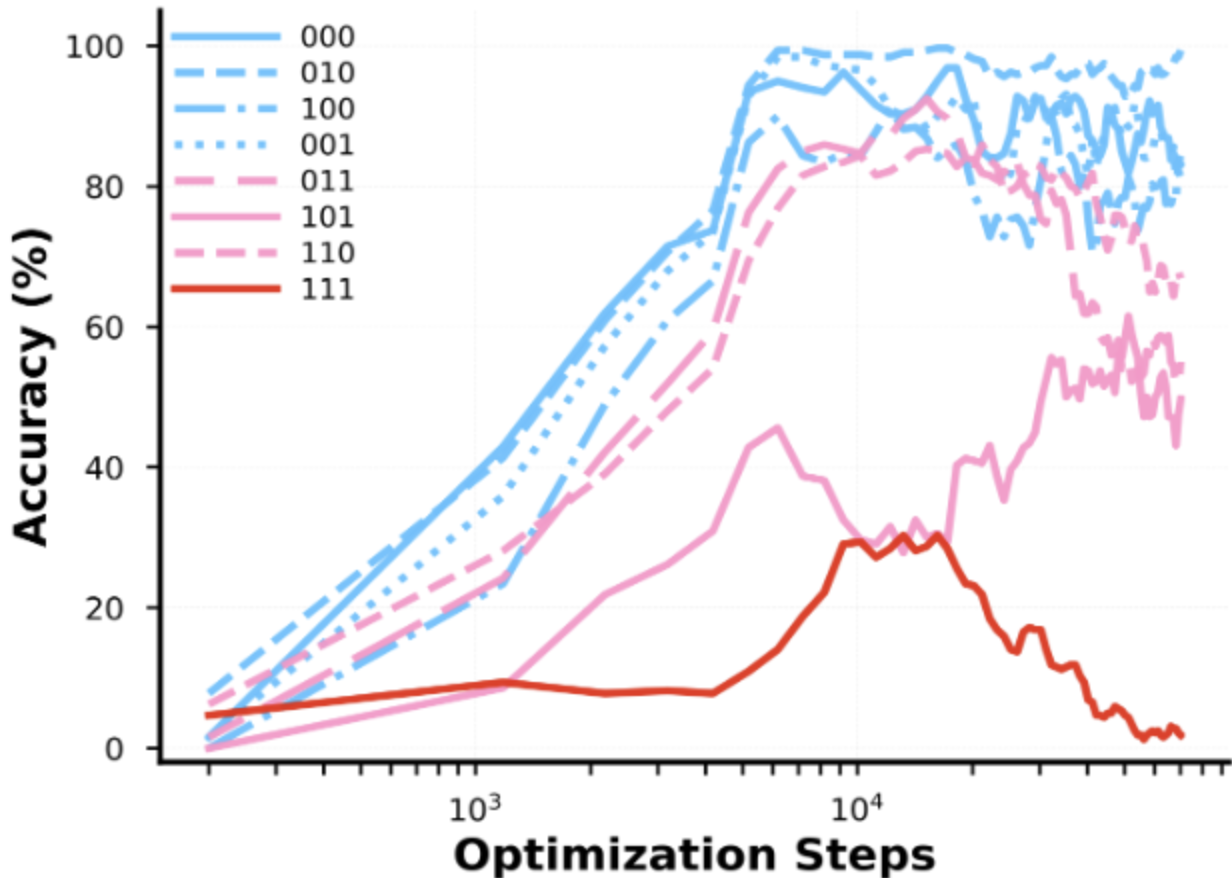}
        \caption*{(f) Small Learning Rate (5e-5 $\rightarrow$ 1e-5) -- Batch Size 256}
    \end{minipage}

    \caption{Compositional generalization across different training conditions for MaskGIT on Shapes2D. We see that MaskGIT struggles to compositionally generalize despite varying multiple different training parameters.}
    \label{fig:training_conditions}
\end{figure}

\paragraph{Ablations across training conditions}
MaskGIT robustness to training configuration. To rule out under-optimization, capacity, or implicit regularization as confounders for the MaskGIT–DiT gap, we ran an extensive sweep on Shapes2D varying learning rate (5e-5 to 3e-4 and 5e-5 to 1e-5), model depth (12 to 18 layers), batch size (128 and 256), and training length (60k to 200k steps). MaskGIT failed to generalize on level-2 compositions across all configurations, indicating that the gap reflects the categorical objective rather than insufficient tuning. We present our results in Figure~\ref{fig:training_conditions}.

\subsection{Autoregressive Decoding}
\label{app:ar_baseline}

MaskGIT is the only discrete model in our main comparison, and it decodes tokens in parallel. To disentangle the effect of the discrete objective from the decoding scheme, we train an autoregressive (causal) transformer over the same discrete visual tokens that MaskGIT uses, and a continuous counterpart in the GIVT-causal setting \citep{tschannen2024givt}, which predicts Gaussian mixture parameters over unquantized latents. Both models share the tokenizer/VAE, architecture, and training data of their parallel counterparts on Shapes2D, so the decoding scheme is the only variable.

As shown in \cref{tab:ar_baseline}, both causal variants fit the seen compositions comparably (86.7 vs.\ 90.2), but diverge on unseen ones (38.5/6.2 vs.\ 64.1/46.9). This reproduces the discrete/continuous gap we observe for the parallel models, even though fixed raster-order causal transformers are often suboptimal for vision in contrast to language.

\begin{table}[h]
\centering
\caption{Probe accuracy (\%) of causal autoregressive models with a discrete (categorical) and a continuous (Gaussian mixture) objective on Shapes2D.}
\small
\begin{tabular}{lccc}
\toprule
Model & Seen Mean & Level-1 Novel Mean & Level-2 Novel \\
 & & (011, 110, 101) & (111) \\
\midrule
AR (discrete)         & 86.7 & 38.5 & 6.2 \\
AR-GMM (continuous)   & \textbf{90.2} & \textbf{64.1} & \textbf{46.9} \\
\bottomrule
\end{tabular}
\label{tab:ar_baseline}
\end{table}

\subsection{Unmasking Order}
\label{app:unmasking_order}

MaskGIT decodes iteratively, unmasking the most confident tokens first. One could hypothesize that this heuristic commits early to high-confidence tokens that lock in seen combinations and thereby causes the compositional failures. To test this, we replace confidence-based unmasking with a random unmasking order while keeping the trained model fixed. \cref{tab:unmasking_order} shows that removing the confidence heuristic does not recover compositional generalization: the novel mean drops by 10.7 points, while seen compositions are unaffected.

\begin{table}[h]
\centering
\caption{Probe accuracy (\%) of MaskGIT on Shapes2D with confidence-based vs.\ random unmasking order (256 samples per composition).}
\small
\setlength{\tabcolsep}{4pt}
\begin{tabular}{lccccc|ccccc}
\toprule
Unmasking & 000 & 001 & 010 & 100 & Seen Mean & 011 & 101 & 110 & 111 & Novel Mean \\
\midrule
Confidence    & 87.8 & 94.3 & 82.9 & 77.3 & 85.6 & 64.9 & 54.0 & 38.4 & 0.7 & \textbf{39.5} \\
Random order  & 87.8 & 95.0 & 86.1 & 74.3 & 85.8 & 45.7 & 41.2 & 26.3 & 2.0 & 28.8 \\
\bottomrule
\end{tabular}
\label{tab:unmasking_order}
\end{table}

\subsection{Conditioning Information}
\label{appendix:conditioning_results}

\begin{figure}
    \centering
    \resizebox{\textwidth}{!}{
        \begin{minipage}{\textwidth}
            \begin{subfigure}[b]{0.44\textwidth}
                \centering
                \includegraphics[width=\textwidth]{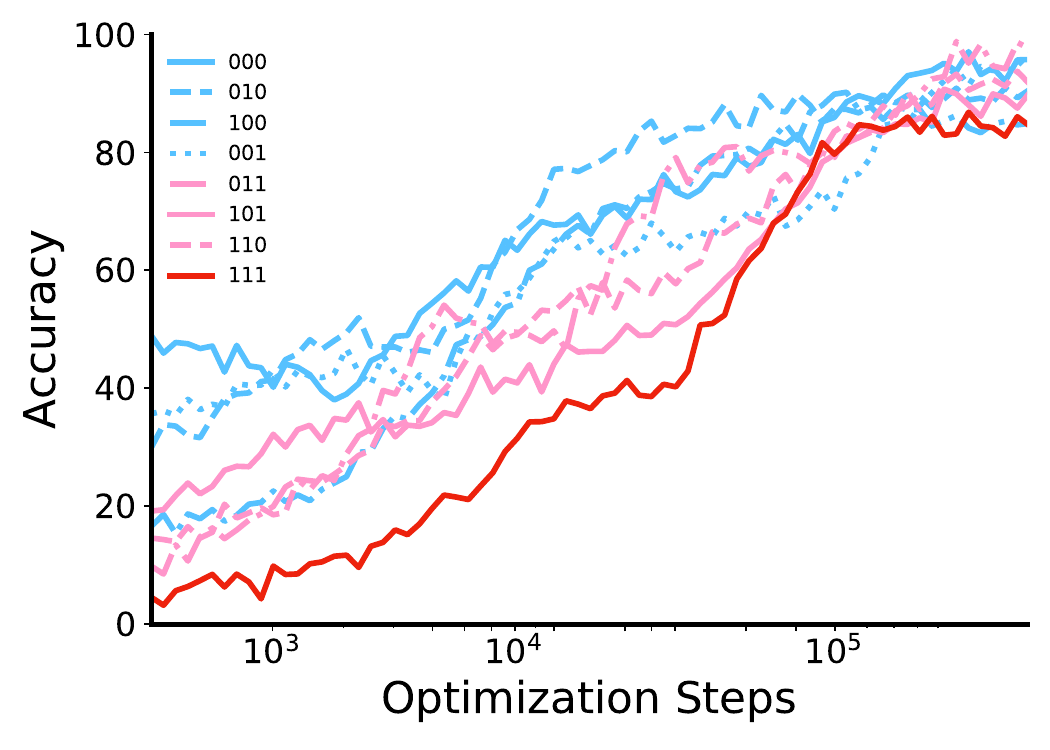}
            \caption{Explicit conditioning}
            \label{fig:dit_clevrer_explicit}
            \end{subfigure}
            \hfill
            \begin{subfigure}[b]{0.44\textwidth}
                \centering
        \includegraphics[width=\textwidth]{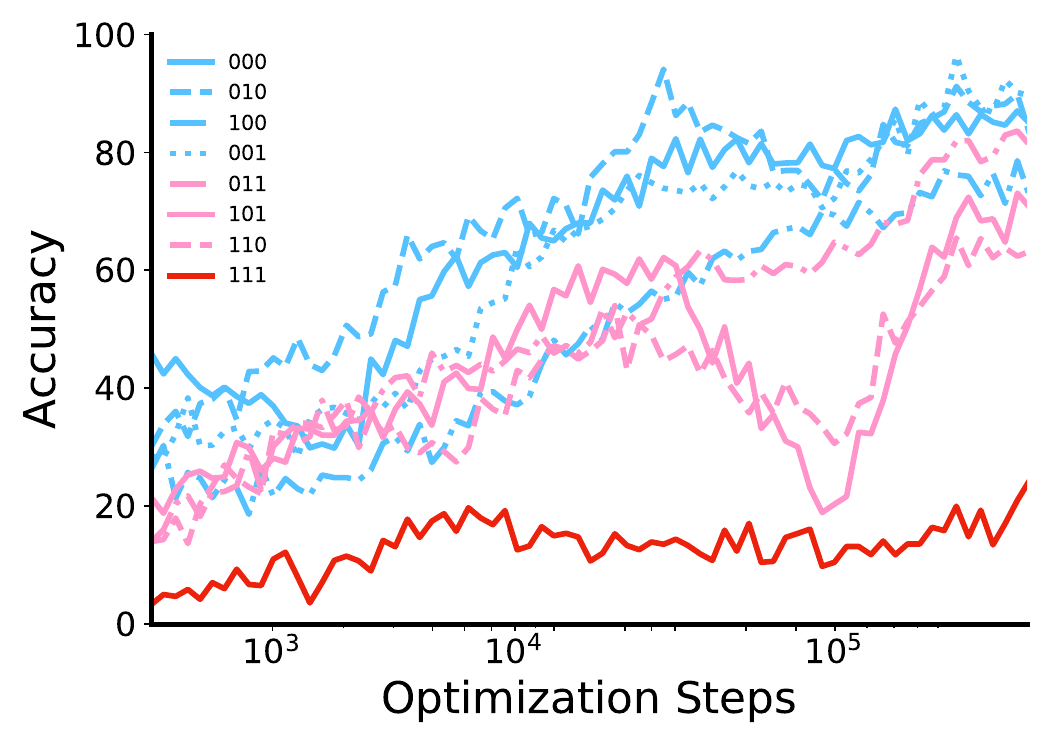}
                \caption{Implicit conditioning}
            \label{fig:dit_clevrer_implicit}
            \end{subfigure}
        \end{minipage}
    }
        \caption{Comparison of linear probe accuracy for DiT trained on CLEVRER-Kubric with (\subref{fig:dit_clevrer_explicit}) Explicit conditioning and (\subref{fig:dit_clevrer_implicit}) Implicit conditioning mechanisms. Training with explicit label information allows for better level-2 compositions.
        The \textcolor{lightblue}{blue} curves show performance on training data, \textcolor{darkpink}{pink} curves depict level-1 compositions, and \textcolor{red}{red} curve denotes level-2 compositions.
        }
    \label{fig:conditioning_results_clevrer}
\end{figure}

Section 4.3 examined the effect of different conditioning mechanisms on a DiT trained on Shapes2D. \cref{fig:conditiong_mechanism_dropout} illustrates a memorization failure mode that arises when concept information is only partially provided during training. In such cases, the model tends to generate realistic images that are misaligned with the intended conditioning concept but resemble the closest training instance, e.g., a blue big triangle is rendered as a red big triangle.
\cref{fig:conditioning_results_clevrer} shows the effect of different conditioning mechanisms for a DiT trained on CLEVRER-Kubric. Given the difference in modalities, we slightly alter our experimental setup to make use of the temporal component in videos that provide a pre-defined context in the form of past frames in which concept information is presented implicitly. For our experiments on CLEVRER-Kubric, we evaluate whether compositional abilities depend on concept information provided explicitly as labels (\cref{fig:dit_clevrer_explicit}) or implicitly through context frames (\cref{fig:dit_clevrer_implicit}). The results show that explicit signals are crucial for compositional generalization, highlighting the models' limitations in abstracting concept information from implicit observations alone. This finding parallels our Shapes2D results, suggesting that when concept information is partially or not explicitly provided, the model must extract it from raw observations in pixel space, a challenge usually present in representation learning that often limits compositional generalization.

\begin{figure}
    \centering
    \includegraphics[width=1.0\linewidth]{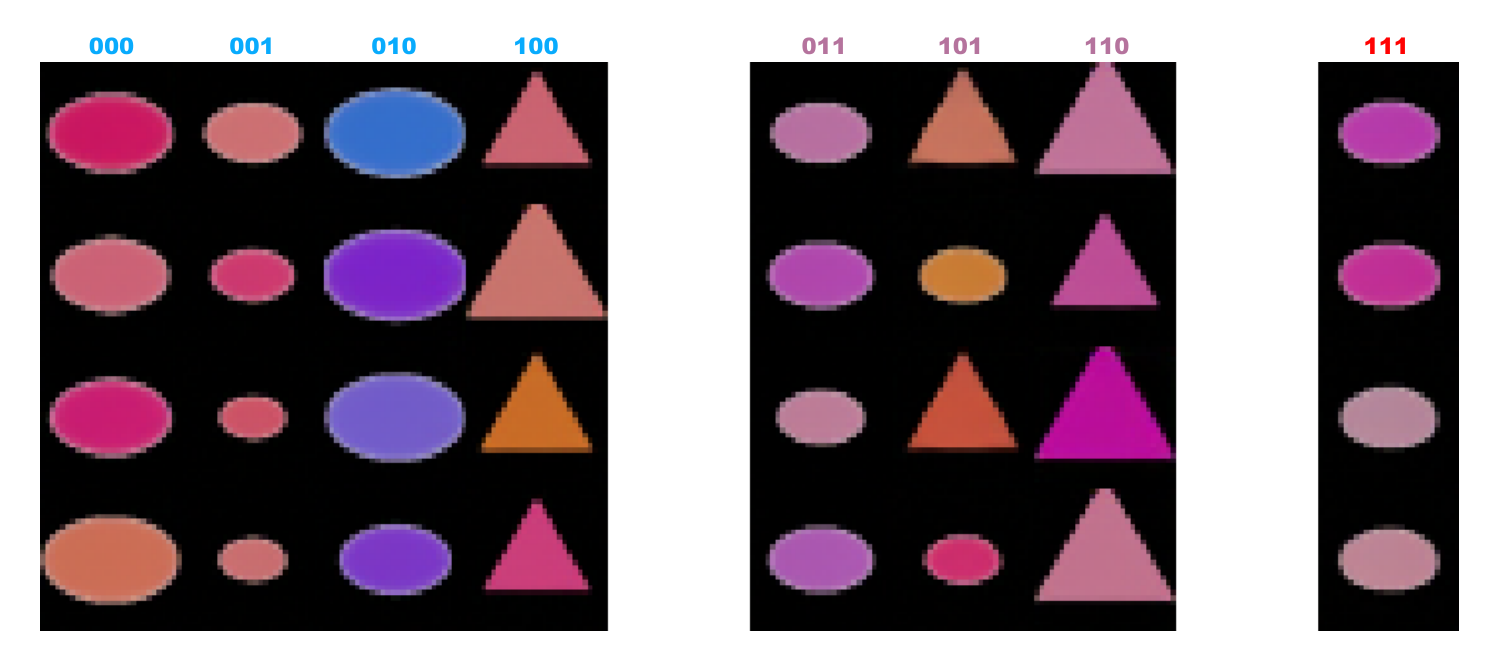}
    \caption{Qualitative Results of the concept dropout with DIT on Shapes2D. The visualization demonstrates that when contextual information about all concepts is incomplete, the model struggles to generalize compositionally. A typical failure mode is the substitution of novel concept combinations with the closest seen ones—for instance, the 110 configuration, which should correspond to a novel blue big triangle, is instead rendered as a red big triangle, likely due to its proximity in the training distribution. Similar behavior is happening to the 101, which should correspond to a novel red small triangle, sometimes result in a red small circle or red big triangle.}
    \label{fig:shap2d_dropout}
\end{figure}

\subsection{Language Conditioning}
\label{app:text_conditioning}

Our main experiments condition on explicit factor tuples to rule out confounding by a text encoder. Language can be viewed as a projection of the underlying image-generating factors: it quantizes continuous attributes into coarse semantic categories (e.g., many areas become ``large'') and often omits factors altogether (e.g., ``a small circle'' instead of ``a small red circle''). Language conditioning therefore introduces the same two information-loss mechanisms that we isolate in \cref{appendix:conditioning_results}.

To test this, we replace factor conditioning with CLIP ViT-B/32 \citep{radford2021clip} text embeddings on Shapes2D, keeping the experimental setup of DiT and MaskGIT otherwise unchanged. All factors are seen during training, but not all of their combinations. We additionally train with partial descriptions, where some factors are omitted from the prompt.

\cref{tab:text_conditioning} shows that, for DiT, language conditioning degrades compositional generalization, and partial descriptions degrade it further, most severely on Level-2 compositions. This matches the degradation predicted by our controlled conditioning study. For MaskGIT, language conditioning does not help the discrete objective, and partial descriptions again hurt. DiT consistently outperforms MaskGIT on novel compositions under both factor and language conditioning.

\begin{table}[h]
\centering
\caption{Probe accuracy (\%) on Shapes2D with explicit factor conditioning vs.\ CLIP text conditioning. L1/L2 denote Level-1/Level-2 novel compositions; All Novel averages over all unseen compositions.}
\small
\begin{tabular}{llcccc}
\toprule
Model & Conditioning & Seen & Novel L1 & Novel L2 & All Novel \\
\midrule
\multirow{3}{*}{DiT}     & Factors                           & \textbf{97.3} & \textbf{99.0} & \textbf{96.9} & \textbf{98.4} \\
                         & Language                          & 74.2 & 71.4 & 48.4 & 65.6 \\
                         & Language (partial descriptions)   & 76.6 & 62.5 & 1.6  & 47.3 \\
\midrule
\multirow{3}{*}{MaskGIT} & Factors                           & 85.6 & 52.4 & 0.7  & 39.5 \\
                         & Language                          & 84.1 & 52.0 & 1.6  & 39.4 \\
                         & Language (partial descriptions)   & 85.2 & 19.9 & 0.3  & 15.0 \\
\bottomrule
\end{tabular}
\label{tab:text_conditioning}
\end{table}

\subsection{Per-experiment Compute Budget}
Most of our image experiments were carried out on $4\times$NVIDIA L40 GPUs, while our video experiments used $2-4\times$NVIDIA H200 GPUs. We would also be releasing code upon publication to ensure full reproducibility of our results.

\section{Enhancing Compositional Learning with Joint Embedding Predictive Architectures}
\label{appendix:jepa}

\subsection{Augmenting the MaskGIT Training Objective with JEPA}

The core idea of JEPA is to predict the \textit{representation} of masked portions of the input (target) blocks from the \textit{representation} of visible portions (context blocks) in an abstract embedding space and to learn both the representational embedding \textit{and} the predictor end-to-end.
By positioning the target not as a noisy, high-dimensional input, but in a lower-dimensional, cleaned-up embedding, it allows the model to capture relevant factors in the target content.
This encourages the encoder to learn representations that capture high-level, predictable semantic information while potentially abstracting away low-level, instance-specific details that may hinder generalization.
We formulate our JEPA-based training objective described below.

An input data sample $x$ is first encoded into a sequence of discrete visual tokens $E_{VQ}(x) \in \{1,...,K\}^{N}$ where $E_{VQ}$ is the encoder trained with a VQ-VAE objective and $N$ is the sequence length. The standard MaskGIT objective $(\mathcal{L}_{MG})$ focuses on reconstructing masked input tokens. We employ block-masking \cite{bardes2024vjepa} for images and causal block masking for videos \cite{bardes2024vjepa} to sample a mask $\mathcal{M} \subset \{1,...,N\}$. The MaskGIT Transformer $T$ predicts the probability distribution $p(z_{i}|z_{\backslash M})$ for each masked token $z_{i}~(i \in M)$ based on unmasked token $z_{\backslash M}$. The loss is the negative log-likelihood of the true tokens at the masked positions:
 \begin{equation}
 \label{eq:maskgit_loss}
     \mathcal{L}_{MG} = -\mathbb{E}_{x,M} \left[\sum_{i \in M}\log p(z_{i}|z_{\backslash M}) \right]
 \end{equation}

Let $H^{(l)}$ represent the sequence of continuous hidden state vectors output by the $l$-th layer of the MaskGIT Transformer $T$. We introduce a JEPA-style objective applied to the hidden states of specific intermediate layers $l \in L_{JEPA}$. This acts as a representation alignment objective within the network.

For each selected layer $l \in L_{JEPA}$, we apply a separate JEPA masking strategy $M_{JEPA}$. This strategy defines a set of context indices $C \subset \{1,...,N\}$ and target indices $T \subset \{1,...,N\}$, typically corresponding to spatial blocks.
We extract the hidden states from layer $l: H^{(l)}=\{h_{1}^{l},...,h_{N}^{l}\}$. The extracted \textit{context} hidden states $H^{(l)}_{C}=\{h_{j}^{(l)}|j \in C \}$ is used as input to a layer-specific predictor network $P^{(l)}_{JEPA}$. In practice, $P^{(l)}_{JEPA}$ is simply parametrized using a multilayer perceptron (MLP). The predictor network aims to predict the hidden states of the target tokens $\hat{H}^{(l)}_{T} =P^{(l)}_{JEPA} (H^{(l)}_{C})$.

The JEPA loss for layer $l$ measures the discrepancy between the predicted target representation (for target index $k \in T$) $\hat{h}_{k}^{(l)}$ and the actual target representations $h_{k}^{(l)}$ (computed by the EMA Transformer $T$ itself) using Mean Squared Error (MSE) as a distance metric $d$
\begin{equation}
 \label{eq:jepa_loss}
     \mathcal{L}_{JEPA}^{(l)} = \mathbb{E}_{x,M_{JEPA}}\left[\frac{1}{|T_{idx}|}\sum_{k \in T_{idx}}d\left(\hat{h}_{k}^{(l)}, sg(h_{k}^{(l)}) \right) \right]
\end{equation}
where $T_{idx}$ denotes the set of target indices and $sg(\circ)$ denotes the stop-gradient operation.
This ensures that the encoder is updated to produce context representations $H_{C}^{(l)}$ that are predictive of the target representations $h_{k}^{(l)}$, rather than trivially learning to reconstruct $h_{k}^{(l)}$ through the target pathway.

The final JEPA loss can be the same as \cref{eq:jepa_loss} if applied only to a single layer, or a summation of losses applied to all layers. In the latter case:
 \begin{equation}
 \label{eq:final_jepa_loss}
     \mathcal{L}_{JEPA} = \sum_{l \in L_{JEPA}}\mathcal{L}_{JEPA}^{(l)}
 \end{equation}

The final training objective is a weighted sum of the standard MaskGIT reconstruction loss (\cref{eq:maskgit_loss} and the JEPA representation alignment loss (\cref{eq:final_jepa_loss}):
 \begin{equation}
 \label{eq:final_train_loss}
     \mathcal{L}_{Total} = \mathcal{L}_{MG} + \lambda\mathcal{L}_{JEPA}
 \end{equation}
where $\lambda$ is a hyperparameter balancing the contribution of the two objectives.

\subsection{Results on CLEVRER-Kubric}

\begin{figure}[t]
    \centering
            \begin{subfigure}[b]{0.32\textwidth}
                \centering
                \includegraphics[width=\textwidth]{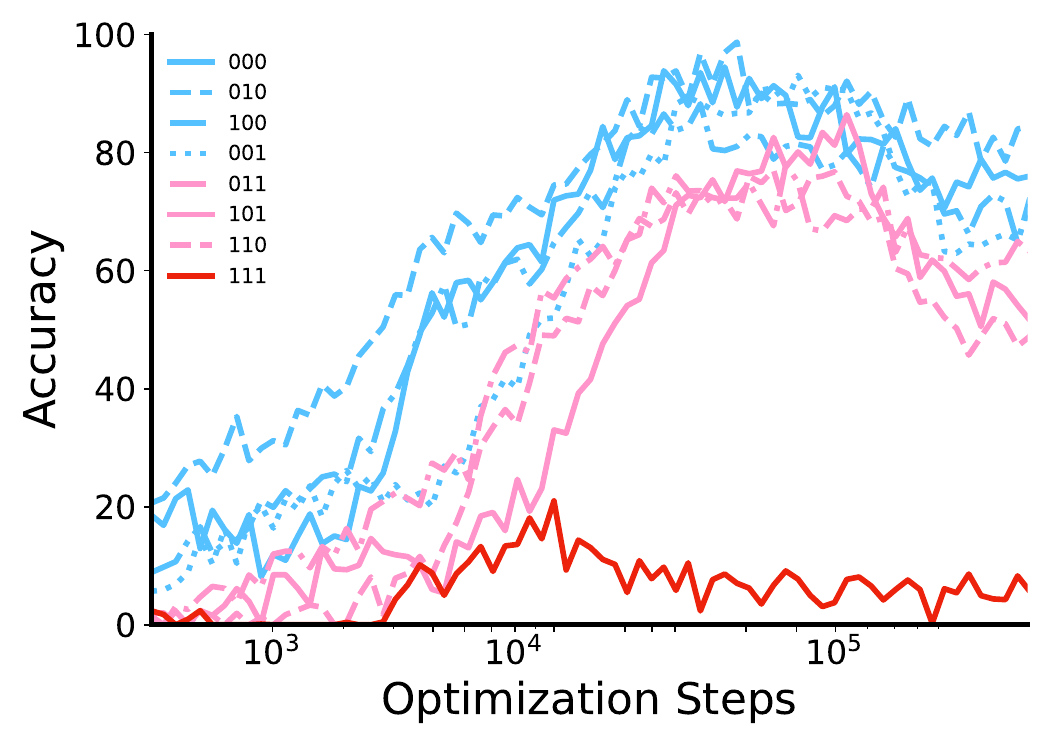}
                \caption{Standard Objective}
                \label{fig:maskgit_vid}
            \end{subfigure}
            \hfill
            \begin{subfigure}[b]{0.32\textwidth}
                \centering
                \includegraphics[width=\textwidth]{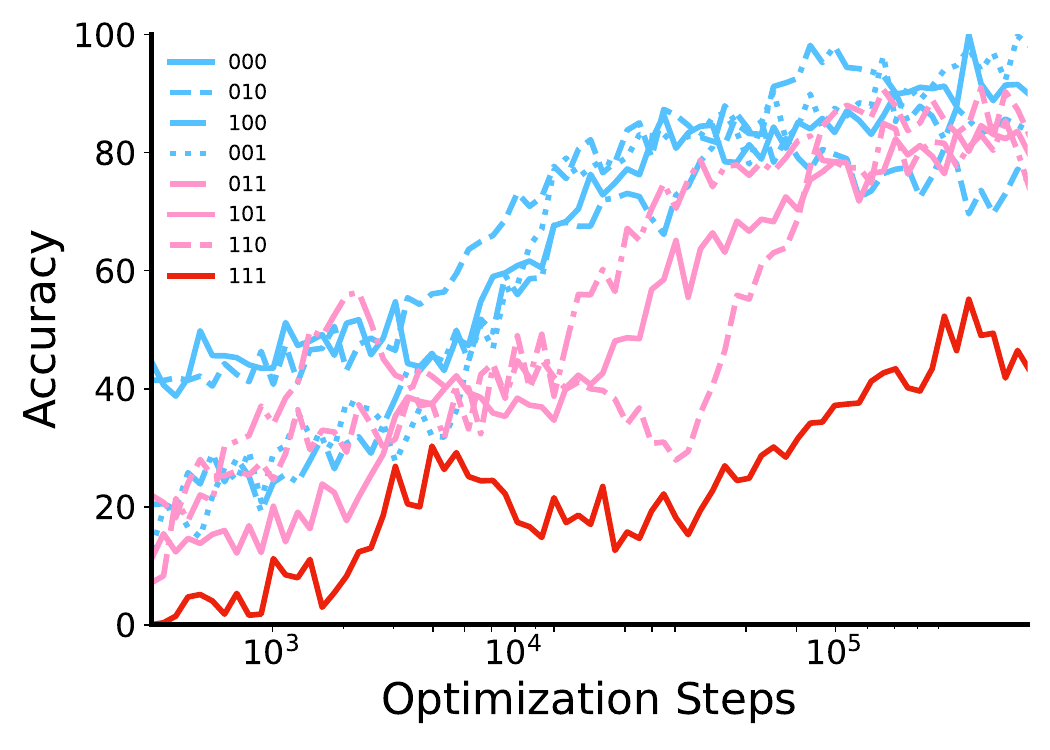}
                \caption{JEPA-based Objective}
                \label{fig:maskgit_vid_jepa}
            \end{subfigure}
            \hfill
            \begin{subfigure}[b]{0.32\textwidth}
                \centering
                \includegraphics[width=\textwidth]{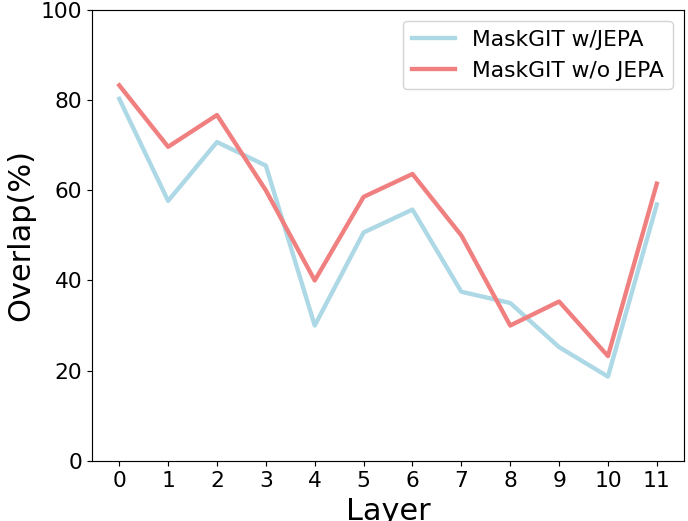}
                \caption{Polysemanticity Analysis}
                \label{fig:polysemanticity_videos}
            \end{subfigure}
    \caption{Comparison of linear probe accuracy for MaskGIT with Standard (\subref{fig:maskgit_vid}) and JEPA-based (\subref{fig:maskgit_vid_jepa}) training objectives for CLEVRER-Kubric. The JEPA-based training objective clearly enhances compositional abilities, and lowers polysemanticity (\subref{fig:polysemanticity_videos}) even though it cannot fully compensate the problems introduced by the discrete space.
    The \textcolor{lightblue}{blue} curves show performance on training data, \textcolor{darkpink}{pink} curves depict level-1 compositions, and \textcolor{red}{red} curve denotes level-2 compositions.
    }
    \label{fig:jepa_video_results_clevrer}
\end{figure}

To reinforce our findings from the main text, we also extend our JEPA-based objective for video generation using the CLEVRER-Kubric dataset. We see that our results follow a similar trend as reported for Shapes2D in the main text. \cref{fig:maskgit_vid} shows that the standard MaskGIT objective is insufficient for reliable compositional generalization. However, augmenting it with our JEPA-based objective yield improvement in compositional abilities (\cref{fig:maskgit_vid_jepa}). We also conduct a polysemanticity analysis (\cref{fig:polysemanticity_videos}), observing the same trend as Shapes2D, with the JEPA-augmented objective exhibiting lower polysemanticity in attention heads- which points to the model learning factor-specific representations.

\subsection{Ablation Studies}

\paragraph{Analysis on JEPA-loss on different layers}

As seen in the previous sections, we apply the JEPA loss to different layers $l$ within the transformer block. We show an empirical ablation below contrasting the performance of applying the JEPA loss to different layers for a MaskGIT model trained on CLEVRER-Kubric \cref{tab:jepa_loss} shows the maximum probe accuracy of the \texttt{111} composition with the JEPA losses at different layers or combinations of layers. Following previous works \cite{yu2025repa}, we also applied predictive coding in the representations of mid-level layers. We see slight differences after varying combinations but we phrase the problem as parameter tuning and empirically select the best combination for optimal trade-off between accuracy and performance.

\begin{table}[htp]
\centering
\caption{Maximum linear probe accuracy after applying the JEPA loss on different layers and combinations of layers for a MaskGIT model trained on CLEVRER-Kubric}
\begin{tabular}{lc}
\toprule
\textbf{Layers} & \textbf{Accuracy}\\
\midrule
\{6\}        & 27.61 \\
\{8\}        & 26.35 \\
\{6,8\}      & 38.16 \\
\{7,9\}      & 36.62 \\
\{6,8,10\}   & 53.52 \\
\{7,9,11\}   & \textbf{56.27} \\
\{7,8,9,11\} & 47.25 \\
\bottomrule
\end{tabular}
\label{tab:jepa_loss}
\end{table}

\paragraph{Effect of $\lambda$}

We also provide an empirical ablation on varying the weighting factor $\lambda$ as seen in \cref{eq:final_jepa_loss}.
Based on the graph, we select 0.6 as the weighting factor for our experiments.

\subsection{Polysemanticity in Attention Heads}
\label{appendix:polysemanticity}

We give a deeper explanation into how we perform the polysemanticity analysis mentioned in the main text. We quantify polysemanticity in attention heads by assessing their impact on the model's ability to distinguish between different visual features like \texttt{color} or \texttt{shape}. An attention head is identified as polysemantic if it influences the processing of multiple distinct features.
The core of this approach involves an intervention- specifically, ablating individual attention heads and measuring the resultant change in the model's representation for contrasting pairs of data samples.

\begin{wrapfigure}{r}{0.5\textwidth}
  \centering
  \includegraphics[width=0.48\textwidth]{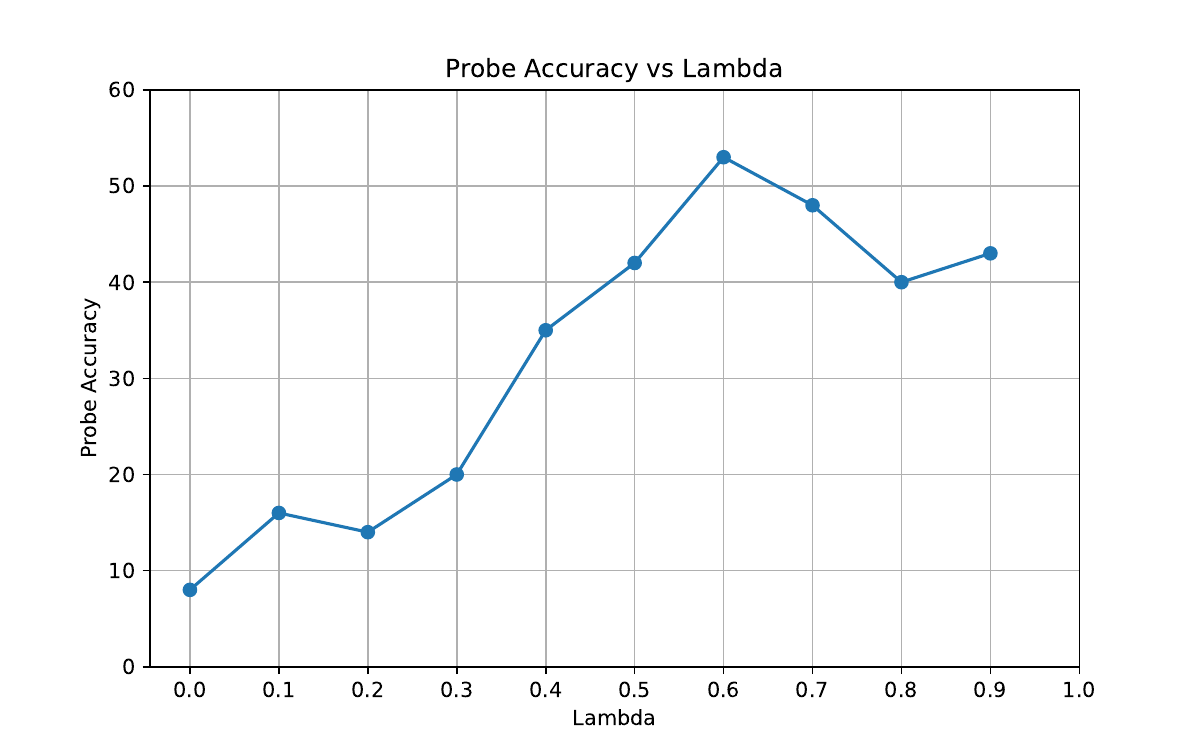}
  \caption{Effect of weighting factor on probe accuracy}
\end{wrapfigure}

First, we establish a baseline for how the original, unablated model represents and distinguishes the features of interest. We do so by passing curated pairs of images or videos, with the samples in each pair differing along a single, targeted visual feature dimension. For instance, to assess color processing, we use pairs of \texttt{red square} and \texttt{blue square}. For each sample $I$, its global representation is extracted from the model, typically by using an additional token that aggregates all features of the sample (similar to the \texttt{[CLS]} token). These embeddings are L2 normalized before further use.

\begin{equation}
    [CLS]_{norm}(I) = \frac{[CLS](I)}{||[CLS](I)||_{2}}
\end{equation}

The model's ability to distinguish a feature $f$ is quantified by the cosine similarity between these embeddings for pair $(I_{a},I_{b})$ contrasting in that specific feature. Given that the tokens are already L2-normalized, their cosine similarity is simply their dot product

\begin{equation}
    sim(I_{a},I_{b})=[CLS]_{norm}(I_{a}) \cdot [CLS]_{norm}(I_{b})
\end{equation}

This gives us a baseline similarity score $sim_{baseline,f}$ for each feature under consideration.

Next, we perform targeted ablation of a specific attention head $h$ within a given layer $l$. This intervention involves creating a copy of the model and modifying the forward pass of its multi-head self-attention (MHSA) module. Specifically, the $QKV$ vectors corresponding to $h$ are set to 0 immediately before the attention scores are computed. This allows us to observe the model's behavior in the absence of the specific attention head $h$.

Using the ablated model, $M_{ablated(l,h)}$, we recalculate the L2-normalized \texttt{[CLS]} token embeddings and subsequently the feature similarity score $sim_{ablated,f}$ for the same pairs. The impact $E_{f}(l,h)$ of ablating head $(l,h)$ on the processing of feature $f$ is defined as the change in this similarity score:

\begin{equation}
    E_{f}(l,h) = sim_{ablated,f} - sim_{baseline,f}
\end{equation}

An attention head $(l,h)$ is considered to significantly impact feature $f$ if the absolute magnitude of its effect $|E_{f}(l,h)|$, exceeds a predefined threshold $\tau$. For our experiments, we consider $\tau=0.005$. The head is then defined polysemantic if it meets this significance criterion for multiple distinct features.

\subsection{Mechanistic Similarity}
\label{appendix:circuits}

We provide a more comprehensive explanation of our method to compare the circuit overlap for different concepts across different models. Similar to \citet{kempf2025clip}, we first identify important neurons within the MLP blocks by perturbing their activations. Specifically, we zero out the output of the fully connected layer for a neuron and measure the impact on the model's classification score for a target class. The impact is defined as

\begin{equation}
    \text{Impact}(a_{l,j}) = |S_{C}(I)-S'_{C}(I|a_{l,j}=0)|
\end{equation}

where $S_{C}(I)$ is the baseline score for class $C$ given data sample $I$ and $S'_{C}(I|a_{l,j}=0)$ is the score when neuron $j$'s activation in block $l$ is zeroed out $(a_{l,j}=0)$. The top $10\%$ neurons causing the largest score change are selected.

Next, we identify important connections between these MLP neurons across consecutive transformer blocks (from block $k$ to $k+1$). We attribute the activation of a target important neuron in the fully connected layer of block $k+1$ to the activations of neurons in block $k$. A connection is deemed important if the source neuron has high attribution and was also identified as an important neuron.

\begin{figure}[]
    \centering
            \begin{subfigure}[b]{0.45\textwidth}
                \centering
                \includegraphics[width=\textwidth]{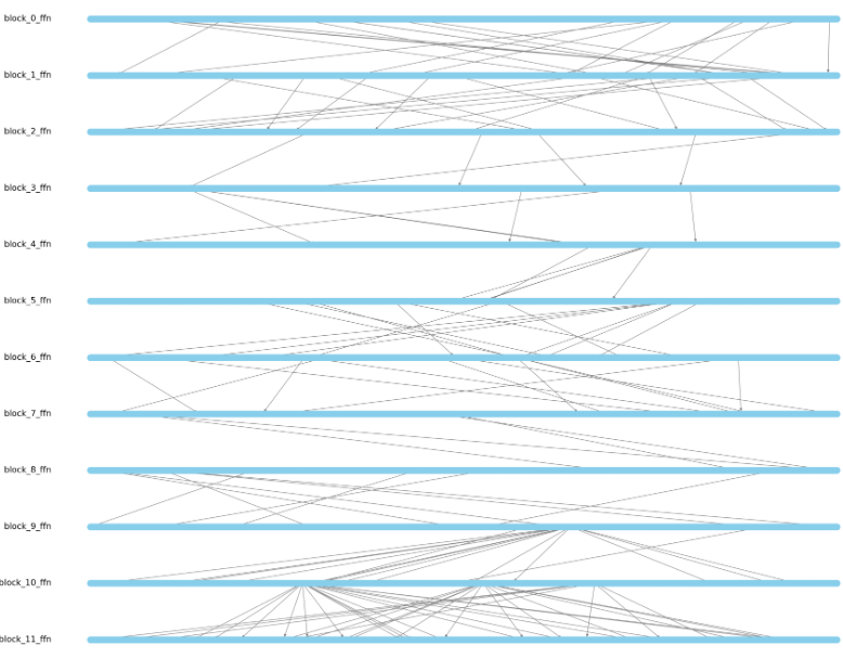}
                \caption{Standard objective}
                \label{fig:red_cube_circuit}
            \end{subfigure}
            \hfill
            \begin{subfigure}[b]{0.45\textwidth}
                \centering
                \includegraphics[width=\textwidth]{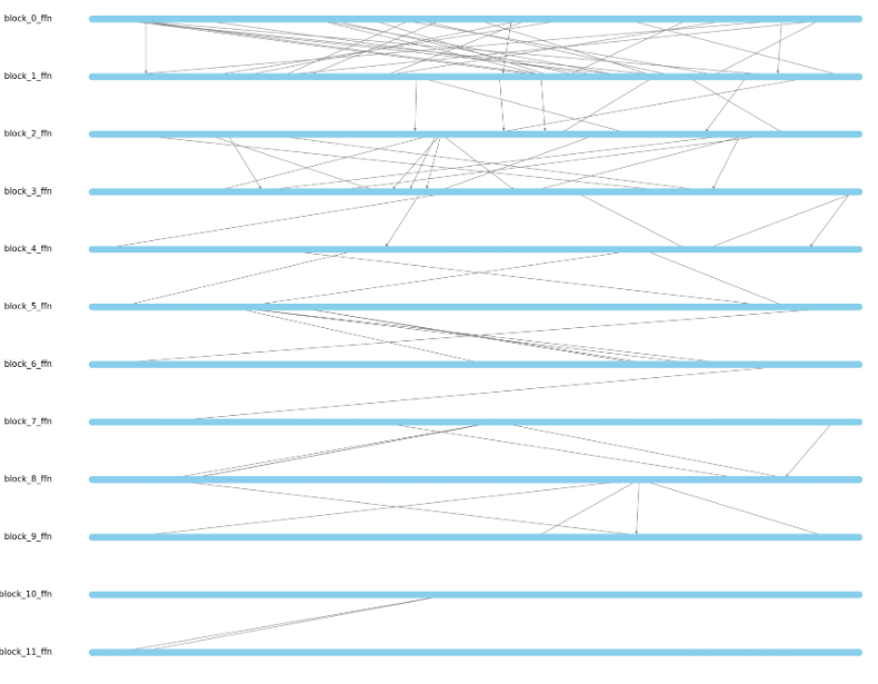}
                \caption{JEPA-based objective}
                \label{fig:red_cube_jepa_circuit}
            \end{subfigure}
            \begin{subfigure}[b]{0.45\textwidth}
                \centering
                \includegraphics[width=\textwidth]{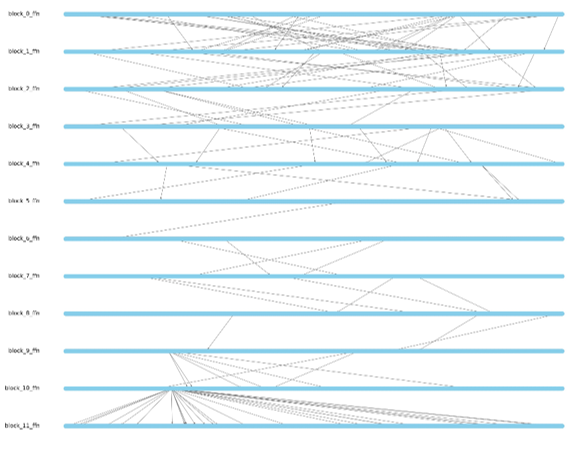}
                \caption{Standard objective}
                \label{fig:blue_sphere_circuit}
            \end{subfigure}
            \hfill
            \begin{subfigure}[b]{0.45\textwidth}
                \centering
                \includegraphics[width=\textwidth]{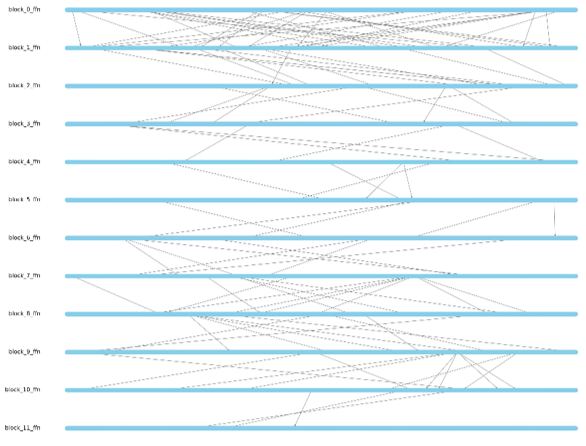}
                \caption{JEPA-based objective}
                \label{fig:blue_sphere_jepa_circuit}
            \end{subfigure}
    \caption{Visualization of MaskGIT circuits for a \texttt{large red cube} and \texttt{small blue sphere} trained with the standard training objective versus the JEPA-based training objective.}
    \label{fig:circuits_maskgit}
\end{figure}

Finally, these important neurons and connections form a class specific circuit-graph (visualized in \cref{fig:circuits_maskgit}). We then compare circuits from different classes using Jaccard similarity for measuring neuron overlap.

\paragraph{Do factor-specific circuits support compositional generalization?}
To provide evidence for the role of factor-specific circuits, we conduct a circuit ablation experiment. For each factor, we identify relevant neurons by comparing circuits of composition pairs differing only in that factor (e.g., large red sphere vs. large blue sphere yields color-relevant neurons), zero them out at inference, and measure the resulting drop in level-2 probe accuracy. Table~\ref{tab:circuit_ablation} shows our results.

\begin{table}[h]
\centering
\caption{Level-2 probe accuracy drops after ablating factor-specific circuits. Bold entries mark factor-aligned ablations. MaskGIT+JEPA shows selective degradation; standard MaskGIT shows uniform drops, consistent with entangled representations.}
\label{tab:circuit_ablation}
\begin{tabular}{lcc}
\toprule
 & Color-ablated & Shape-ablated \\
\midrule
\multicolumn{3}{l}{\textit{MaskGIT+JEPA}} \\
\quad Color-varying & $\mathbf{\downarrow 38}$\textbf{pp} & $\downarrow$ 6pp \\
\quad Shape-varying & $\downarrow$ 5pp & $\mathbf{\downarrow 41}$\textbf{pp} \\
\midrule
\multicolumn{3}{l}{\textit{Standard MaskGIT}} \\
\quad Color-varying & $\downarrow$ 18pp & $\downarrow$ 14pp \\
\quad Shape-varying & $\downarrow$ 15pp & $\downarrow$ 17pp \\
\bottomrule
\end{tabular}
\end{table}

In MaskGIT+JEPA, ablating a factor's neurons selectively degrades the compositions influenced by that factor, whereas standard MaskGIT shows uniform drops across factors, consistent with entangled representations. This provides evidence that the factor-specific circuits induced by the JEPA objective support compositional generalization.

\section{Theoretical Context for Our Empirical Findings}
\label{app:theory}

In this section, we connect our empirical findings to existing theoretical results on compositional generalization. Several independently developed frameworks provide perspectives that are consistent with our observations, although their assumptions do not exactly match our experimental setting. For each finding, we identify the relevant theoretical result and discuss how its assumptions relate to our setup.

\subsection{Finding 2: Theoretical Perspectives on Continuous Objectives and Compositional Generalization}

Our central empirical result is that models trained with continuous objectives (DiT, MAR, GIVT) generalize compositionally to unseen factor combinations, while models trained with discrete categorical objectives (MaskGIT) do not. Three independent theoretical frameworks provide complementary perspectives consistent with this pattern.

\paragraph{Smooth-decoder results on Cartesian-product extrapolation.}

Lachapelle et al. \cite{lachapelle2023additive} (Theorems~1~\&~2) prove that an additive decoder $\mathbf{f}(\mathbf{z}) = \sum_k \mathbf{f}^k(\mathbf{z}_{B_k})$ achieves \emph{Cartesian-product extrapolation}: if the model fits the ground-truth generator on a training support $\mathcal{Z}_{\text{supp}}$, it also agrees on the full Cartesian product $\mathcal{Z}_1 \times \cdots \times \mathcal{Z}_K$. This result requires each component function $\mathbf{f}^k$ to satisfy a \emph{sufficient nonlinearity} condition stated in terms of the linear independence of first- and second-order partial derivatives of $\mathbf{f}$. Brady et al.~\cite{brady2024interaction} generalize this in Theorem~4.4: compositional generalization holds when both the ground-truth generator and the learned model are $C^{n+1}$ diffeomorphisms with at most $n$th-order interactions across concept slots. Wiedemer et al. \cite{wiedemer2023compositional} (Theorem~1) prove a complementary result: a compositional model generalizes from training support to unseen compositions when the support satisfies a \emph{sufficient support} condition (Definition~3) and the model shares the compositional function class with the ground truth.

All three results require the decoder to be a smooth, continuously differentiable function of its latent inputs. Models trained with the continuous objectives considered here (denoising diffusion, Gaussian mixture likelihood) operate on continuous-valued representations and are parameterized by smooth neural networks, making them more closely aligned with these assumptions. In contrast, MaskGIT's categorical cross-entropy loss predicts a distribution over discrete codebook indices, and the derivative-based arguments above do not directly apply to categorical token prediction. This does not imply that categorical models cannot generalize compositionally; rather, these particular theoretical guarantees do not directly characterize categorical token prediction. Specifically, the key mechanism in Brady et al.'s compositional generalization proof---that the partial derivative $\partial f_l / \partial z_1(z_1^*, z_2)$ is a bounded-degree polynomial in $z_2$ and therefore globally determined by its local behavior---is formulated for continuous latent variables and differentiable decoders.

\paragraph{Score-based objectives and compositional inductive biases.}
Beyond the smoothness requirement, recent theoretical work identifies specific inductive biases of score-matching objectives that favor compositional generation. Kamb \& Ganguli \cite{kamb2024analytic} prove that convolutional diffusion models, through their locality and equivariance biases, learn \emph{local score functions} that produce novel images by combinatorially composing local patches from training examples---a mechanism they term ``combinatorial creativity.'' Bradley et al. \cite{bradley2025local} extend this to conditional diffusion, proving an exact equivalence between \emph{conditional projective composition} (a formal definition of correct compositional generation) and \emph{local conditional scores} -- score functions with sparse dependencies on both pixels and conditioning variables. These results provide concrete examples in which score-based objectives admit structural biases favorable to compositional generation, although both analyses rely on locality or architectural assumptions that differ from the transformer models we train. These guarantees do not directly extend to categorical token prediction, where a score with respect to a discrete token variable is not defined in the same sense.

\paragraph{Regression versus classification extrapolation biases.}
From an optimization perspective, results on ReLU networks trained with MSE show a tendency toward linear extrapolation outside the training support under specific assumptions \citep{xu2020neural}. This provides another perspective consistent with the idea that regression objectives can favor smooth extensions beyond observed samples. In contrast, classical results for cross-entropy in separable classification characterize a maximum-margin bias \citep{soudry2018implicit}. Although neither result directly characterizes the transformer settings studied here, they illustrate qualitatively different extrapolation biases induced by regression and classification objectives.

\subsection{Finding 1: Tokenizer Choice Does Not Substantially Alter Compositionality}

Our experiments show that using a VQ-VAE (discrete) versus a VAE (continuous) tokenizer does not alter compositional generalization, as long as the downstream generative objective is continuous. This is consistent with the theoretical framework above: the compositional generalization theorems constrain the \emph{generative model's decoder} (the function from latent space to observations), not the \emph{tokenizer's} encoding scheme. A VQ-VAE tokenizer maps images to discrete codebook indices, but the codebook entries themselves are continuous vectors. When a continuous generative model (e.g., DiT) operates on continuous-valued inputs (like learned token embeddings for CLEVRER-Kubric, Section~\ref{appendix:tokenizer_results}) rather than on the discrete indices, the downstream mapping remains continuous, retaining the type of smooth structure assumed by these results. Barin-Pacela et al. \cite{barin2024identifiability} formalize a related point: latent factors can be identified even when quantized, provided the underlying mixing function remains a smooth diffeomorphism. Their analysis concerns identifiability rather than generation, so we treat it as supporting context.

\subsection{Finding 3: The Role of Conditioning Information}

We find that quantized or incomplete conditioning on generative factors during training degrades compositional generalization, even for continuous models. Wiedemer et al. \cite{wiedemer2023compositional} (Theorem~1) provide one useful theoretical lens: their result requires the training distribution to have \emph{sufficient support} (Definition~3)---the support must lie in an open set and, for each component function, the subset of training points sharing a fixed value of one factor must contain enough variation in the remaining factors for the Jacobian to be identifiable. Viewed through this lens, quantized or incomplete conditioning can make the effective factor information available to the model coarser or ambiguous, potentially moving the learning problem away from the support and identifiability conditions required by these guarantees. The theorem does not directly model conditioning corruption of the form used in our experiments, so we view this connection as suggestive rather than a direct implication.

Brady et al. \cite{brady2024interaction} offer a related lens through a different formalism: their disentanglement result (Theorem~4.3) requires $\mathcal{Z}_{\text{supp}}$ to be \emph{regular closed} (closure of its interior) and \emph{aligned-connected} (for any slot value, the set of training points sharing that value must be path-connected). Under this view, quantized conditioning could make the effective support more fragmented and incomplete conditioning could reduce its effective dimensionality, both of which would move the problem away from aligned-connectedness and regular closedness. As with \citet{wiedemer2023compositional}, these conditions are stated for the latent support rather than for the conditioning signal, so we view this connection as suggestive.

\subsection{Finding 4: JEPA Auxiliary Loss Partially Recovers Compositionality}

Adding a continuous JEPA-based auxiliary loss to MaskGIT partially recovers compositional generalization and yields more disentangled internal representations. The JEPA objective introduces a continuous regression signal on intermediate representations. This is qualitatively aligned with the smooth-regression perspective above and provides one possible interpretation of why the auxiliary objective helps. However, the theoretical results discussed above do not directly apply to these intermediate representations, so we do not interpret them as a proof of the mechanism. The empirical improvement from JEPA, which narrows but does not close the gap to models whose generative loss is itself continuous, is consistent with this interpretation.

\section{Evaluation}
\label{appendix:eval}
\subsection{Validity of Probe-Based Evaluation}
\label{app:probe_validity}

A composition is considered correct if, for every factor, the intended class has the highest probe probability among its $C$ classes (i.e., more than $0.5$ for binary factors). We verify that this criterion does not reward distorted or ambiguous generations on Shapes2D with three analyses.

\paragraph{Consistency with qualitative results.} In randomly sampled generations (\cref{fig:shapes2d_results}), MaskGIT produces visibly worse compositions than DiT, consistent with the probe results.

\paragraph{Probe decisions are rarely borderline.} Across 512 generated images (three factor predictions each), only 11.6\% of MaskGIT's correct predictions fall in the borderline range and 75.5\% exceed 0.9. For DiT, these numbers are 3.6\% and 90.7\%.

\paragraph{Probes are sensitive to visual quality.} On 1600 real test images, probe accuracy degrades monotonically under increasing corruption strength for blur, Gaussian noise, patch shuffling, occlusion, and downsampling (\cref{tab:probe_corruptions}). Factors that depend on spatial structure (shape, size) are affected first, while color remains largely intact, as expected.

\begin{table}[h]
\centering
\caption{Probe accuracy (\%) on real Shapes2D test images under increasing corruptions. Joint denotes all three factors being classified correctly.}
\small
\begin{tabular}{llcccc}
\toprule
Corruption & Strength & Shape & Color & Size & Joint \\
\midrule
\multirow{4}{*}{Blur ($\sigma$)}           & 1    & 98.9 & 100.0 & 89.5 & 88.4 \\
                                           & 2    & 99.2 & 100.0 & 91.7 & 90.9 \\
                                           & 4    & 73.1 & 100.0 & 89.3 & 64.7 \\
                                           & 8    & 0.0  & 100.0 & 50.0 & 0.0 \\
\midrule
\multirow{4}{*}{Noise ($\sigma$)}          & 0.1  & 97.4 & 100.0 & 93.9 & 91.2 \\
                                           & 0.25 & 91.1 & 100.0 & 78.2 & 75.5 \\
                                           & 0.5  & 65.1 & 99.8  & 50.0 & 45.4 \\
                                           & 1    & 4.5  & 94.4  & 50.0 & 4.0 \\
\midrule
\multirow{4}{*}{Patch shuffle ($k\times k$)} & 2  & 42.6 & 99.6  & 57.6 & 18.5 \\
                                           & 4    & 22.6 & 99.9  & 53.1 & 11.7 \\
                                           & 7    & 21.9 & 100.0 & 50.9 & 9.1 \\
                                           & 14   & 12.9 & 100.0 & 50.0 & 3.6 \\
\midrule
\multirow{2}{*}{Occlusion (fraction)}      & 0.25 & 99.1 & 98.5  & 92.1 & 91.5 \\
                                           & 0.5  & 88.4 & 69.7  & 55.6 & 51.2 \\
\midrule
\multirow{3}{*}{Downsample (factor)}       & 4    & 92.2 & 100.0 & 89.0 & 81.2 \\
                                           & 7    & 44.5 & 100.0 & 87.4 & 39.8 \\
                                           & 14   & 0.0  & 100.0 & 50.0 & 0.0 \\
\bottomrule
\end{tabular}
\label{tab:probe_corruptions}
\end{table}

\subsection{Shapes3D}
\begin{table}[h]
\centering
\begin{tabular}{lcccccccc}
\hline
\textbf{Model} & (0,0,0) & (0,0,1) & (0,1,0) & (1,0,0) & (1,0,1) & (1,1,0) & (0,1,1) & (1,1,1) \\
\hline
DiT & 100\% & 100\% & 100\% & 100\% & 100\% & 100\% & 100\% & 100\% \\
MaskGIT  & 94.7\% & 97\% & 94.7\% & 94\% & 79.49\% & 62.6\% & 58.85\% & 30\% \\
\hline
\end{tabular}
\caption{Performance comparison of DiT and MG across different compositions.}
\label{tab:shapes3d_num_results}
\end{table}

We follow the Shapes2D evaluation protocol and measure compositional generalization using probe accuracy. The probe is a lightweight ResNet-style classifier trained to predict object properties from generated images. It consists of three convolutional blocks with residual connections, followed by global average pooling. The output is passed to three separate linear heads that predict the object’s hue, shape, and scale.

As shown in Table~\ref{tab:shapes3d_num_results}, our key finding still holds: DiT (continuous) generalizes effectively to novel compositions, while MaskGIT (discrete) struggles. Notably, MaskGIT exhibits failure cases such as non-monotonic color attribution to objects and regression to nearest-neighbour training examples.

\subsection{CelebA}
\label{celeba_metric}

\begin{wraptable}{hr}{0.4\textwidth}
\centering
\vspace{-4em}
\caption{Compositional Retrieval Accuracy (CRA, \%) 100 generated samples.}
\begin{tabular}{lcc}
\toprule
\textbf{Model} & \textbf{1-NN} & \textbf{5-NN} \\
\midrule
DiT             & 27           & \textbf{36} \\
MaskGIT         & 21           & 16 \\
MaskGIT + JEPA  & \textbf{31}  & 27 \\
\bottomrule
\vspace{-2em}
\end{tabular}
\label{tab:knn_compositional}
\end{wraptable}

Since there are no established metrics that effectively measure compositional generalization in visual generative models, we introduce a retrieval-based evaluation metric that assesses the compositionality. The core idea is to test whether generated samples from an unseen composition (e.g., \texttt{111}, two factors away from the trainset) are closest in feature space to real images of the same composition.

\vspace{-0.5em}

We compute DINOv2~\cite{oquab2024dinov2} distances between each generated image and all real images for the eight compositions. Using Nearest Neighbor retrieval accuracy, we measure how often a generated image is closest to real images from its target composition. We evaluate DiT, MaskGIT, and JEPA-enhanced MaskGIT on samples from the unseen composition \texttt{111}. Models leveraging continuous distributions show a clear advantage in compositional generalization (Table~\ref{tab:knn_compositional}).

We report FID and FDD on seen and unseen splits as complementary distributional metrics (Table~\ref{tab:seen_unseen_fid_fdd}). These Fr\'echet distances capture split-wise alignment between generated and real samples, but they are not compositionality measures. MaskGIT degrades substantially on the unseen split, while JEPA-enhanced MaskGIT reduces this gap, particularly in unseen FDD, indicating improved alignment to the unseen distribution.

\begin{table}[h]
\centering
\caption{FID and FDD scores on seen and unseen splits on CelebA.}
\small
\begin{tabular}{lrrrr}
\toprule
Method & Unseen FID $\downarrow$ & Unseen FDD $\downarrow$ & Seen FID $\downarrow$ & Seen FDD $\downarrow$ \\
\midrule
DiT      & \textbf{118.60} &  993.92 & \textbf{109.96} & 681.78 \\
MG       & 162.38 & 1470.78 & 113.94 & 719.95 \\
MG+JEPA  & 120.76 &  \textbf{980.92} & 117.35 & \textbf{621.43} \\
\bottomrule
\end{tabular}

\label{tab:seen_unseen_fid_fdd}
\end{table}

\section{Results}
\label{appendix:results}

\subsection{Shapes2D}

\begin{figure}
    \centering
    \includegraphics[height=\textheight,keepaspectratio]{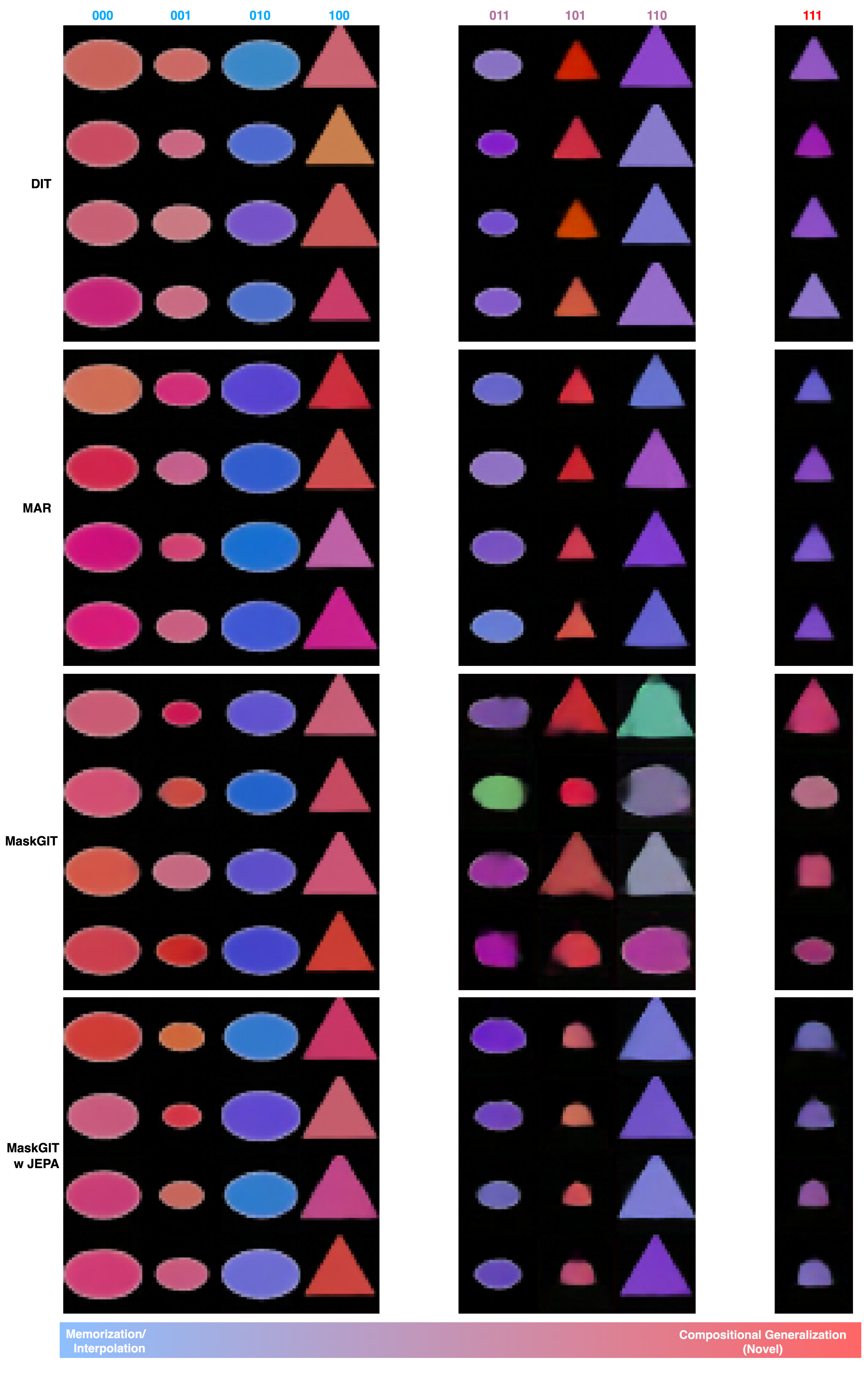}
    \caption{Qualitative Results on Shapes2D}
    \label{fig:shapes2d_results}
\end{figure}

\cref{fig:shapes2d_results} gives a comprehensive overview of qualitative results on Shapes2D.

\subsection{Shapes3D}

\cref{fig:shapes3d_dit,fig:shapes3d_mg} gives a comprehensive overview of qualitative results on Shapes3D. For easier indication of compositional novelty, we use padding colors based on the sum of supergroup assignments across the three factors (color, shape, size), where each factor contributes 0 or 1 depending on its supergroup. Dark \textcolor{darkblue3}{blue} indicates a sum of 0 (all factors from Supergroup~0),
\textcolor{lightblue3}{light blue} indicates a sum of 1,
\textcolor{paleorange3}{pale orange} corresponds to a sum of 2,
and \textcolor{darkred3}{dark red} represents a sum of 3 (all factors from Supergroup~1, i.e., the most novel compositions).

\begin{table}[t]
\centering
\caption{Factorization of the data into \textit{Super Groups}. This partitioning is used to evaluate compositional generalization across disjoint super groups.}
\small
\renewcommand{\arraystretch}{1.15}
\begin{tabularx}{\linewidth}{lXcc}
\toprule
\textbf{Factor} & \textbf{Possible Values} & \textbf{Super Group 0} & \textbf{Super Group 1} \\
\midrule
Color & red, orange, yellow, lime green, green, cyan, blue, indigo, purple, pink
      & red, orange, yellow, lime green, green, cyan, blue
      & indigo, purple, pink \\
Shape & cube, cylinder, sphere
      & cube, cylinder
      & sphere \\
Size  & 0--7 (ordinal)
      & 0, 1, 2, 3, 4, 5
      & 6, 7 \\
\bottomrule
\end{tabularx}

\label{tab:supergroups}
\end{table}

\begin{figure}
    \centering
    \includegraphics[height=\textheight,keepaspectratio]{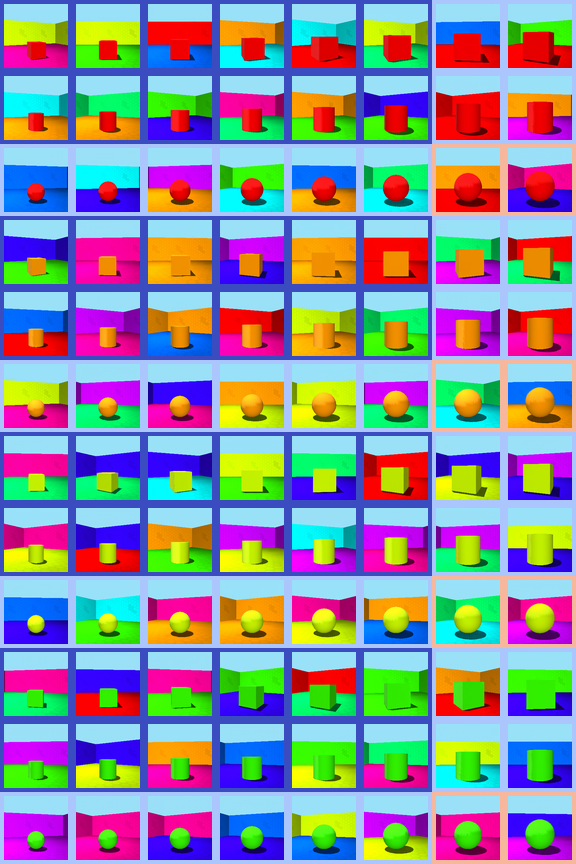}
    \caption{Shapes3D results of DiT.
    \textcolor{darkblue3}{Dark blue}: all factors from Supergroup~0;
    \textcolor{lightblue3}{light blue}: one factor from Supergroup~1;
    \textcolor{paleorange3}{pale orange}: level-1 compositions;
    and \textcolor{darkred3}{dark red}: level-2 novel compositions.}
    \label{fig:shapes3d_dit}

\end{figure}

\begin{figure}\ContinuedFloat
    \centering
    \includegraphics[height=\textheight,keepaspectratio]{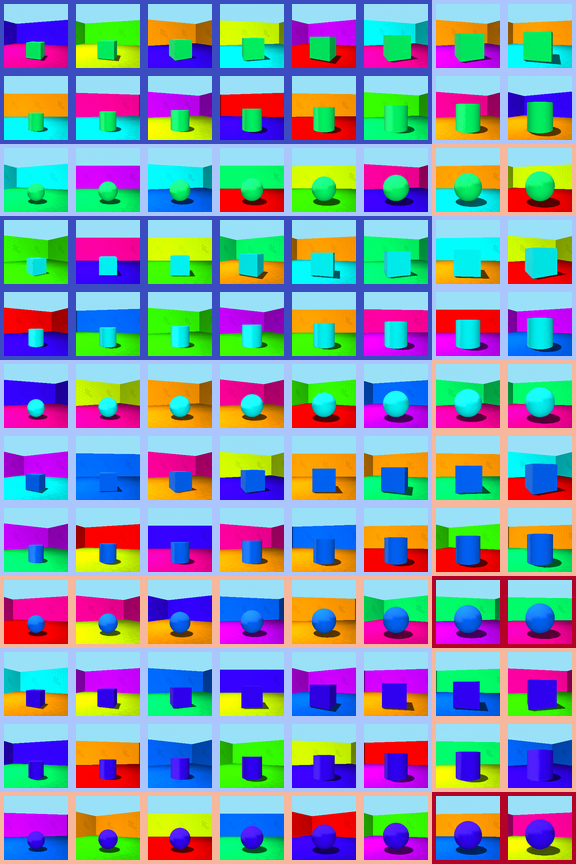}
    \caption{Shapes3D results of DiT.
    \textcolor{darkblue3}{Dark blue}: all factors from Supergroup~0;
    \textcolor{lightblue3}{light blue}: one factor from Supergroup~1;
    \textcolor{paleorange3}{pale orange}: level-1 compositions;
    and \textcolor{darkred3}{dark red}: level-2 novel compositions.}
    \label{fig:shapes3d_dit}
\end{figure}

\begin{figure}\ContinuedFloat
    \includegraphics[width=\textwidth]{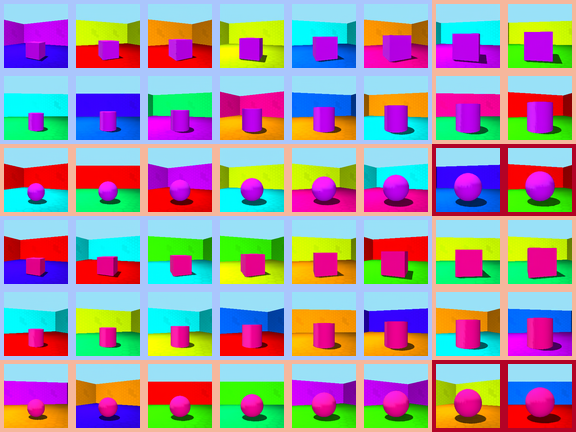}
    \caption{Shapes3D results of DiT.
    \textcolor{darkblue3}{Dark blue}: all factors from Supergroup~0;
    \textcolor{lightblue3}{light blue}: one factor from Supergroup~1;
    \textcolor{paleorange3}{pale orange}: level-1 compositions;
    and \textcolor{darkred3}{dark red}: level-2 novel compositions.}
    \label{fig:shapes3d_dit}
\end{figure}
\begin{figure}
    \centering
    \includegraphics[height=\textheight,keepaspectratio]{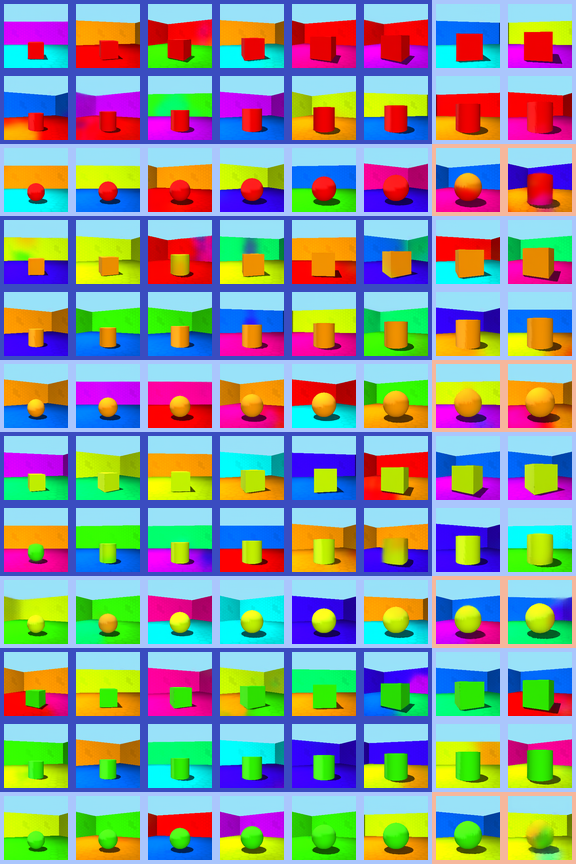}
    \caption{Shapes3D results of MaskGIT.
    \textcolor{darkblue3}{Dark blue}: all factors from Supergroup~0;
    \textcolor{lightblue3}{light blue}: one factor from Supergroup~1;
    \textcolor{paleorange3}{pale orange}: level-1 compositions;
    and \textcolor{darkred3}{dark red}: level-2 novel compositions.}
    \label{fig:shapes3d_mg}

\end{figure}

\begin{figure}\ContinuedFloat
    \centering
    \includegraphics[height=\textheight,keepaspectratio]{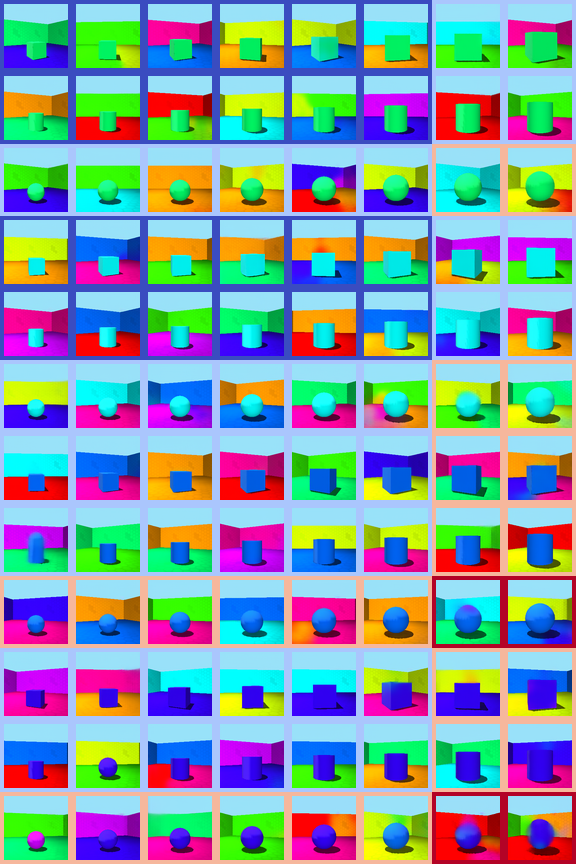}
    \caption{Shapes3D results of MaskGIT.
    \textcolor{darkblue3}{Dark blue}: all factors from Supergroup~0;
    \textcolor{lightblue3}{light blue}: one factor from Supergroup~1;
    \textcolor{paleorange3}{pale orange}: level-1 compositions;
    and \textcolor{darkred3}{dark red}: level-2 novel compositions.}
    \label{fig:shapes3d_mg}
\end{figure}

\begin{figure}\ContinuedFloat
    \includegraphics[width=\textwidth]{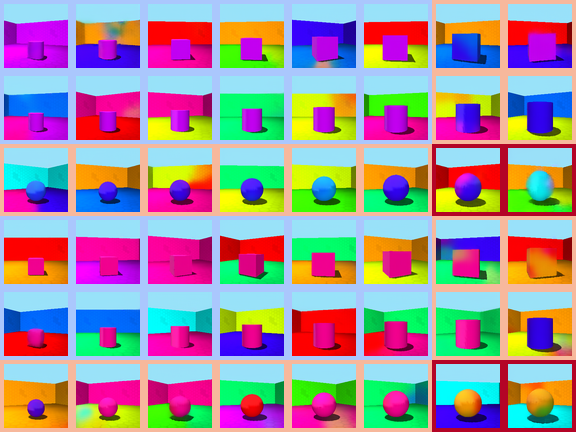}
    \caption{Shapes3D results of MaskGIT.
    \textcolor{darkblue3}{Dark blue}: all factors from Supergroup~0;
    \textcolor{lightblue3}{light blue}: one factor from Supergroup~1;
    \textcolor{paleorange3}{pale orange}: level-1 compositions;
    and \textcolor{darkred3}{dark red}: level-2 novel compositions.}
    \label{fig:shapes3d_mg}
\end{figure}

\subsection{CelebA}

\begin{figure}[p]
 \centering
    \includegraphics[height=\textheight,keepaspectratio]{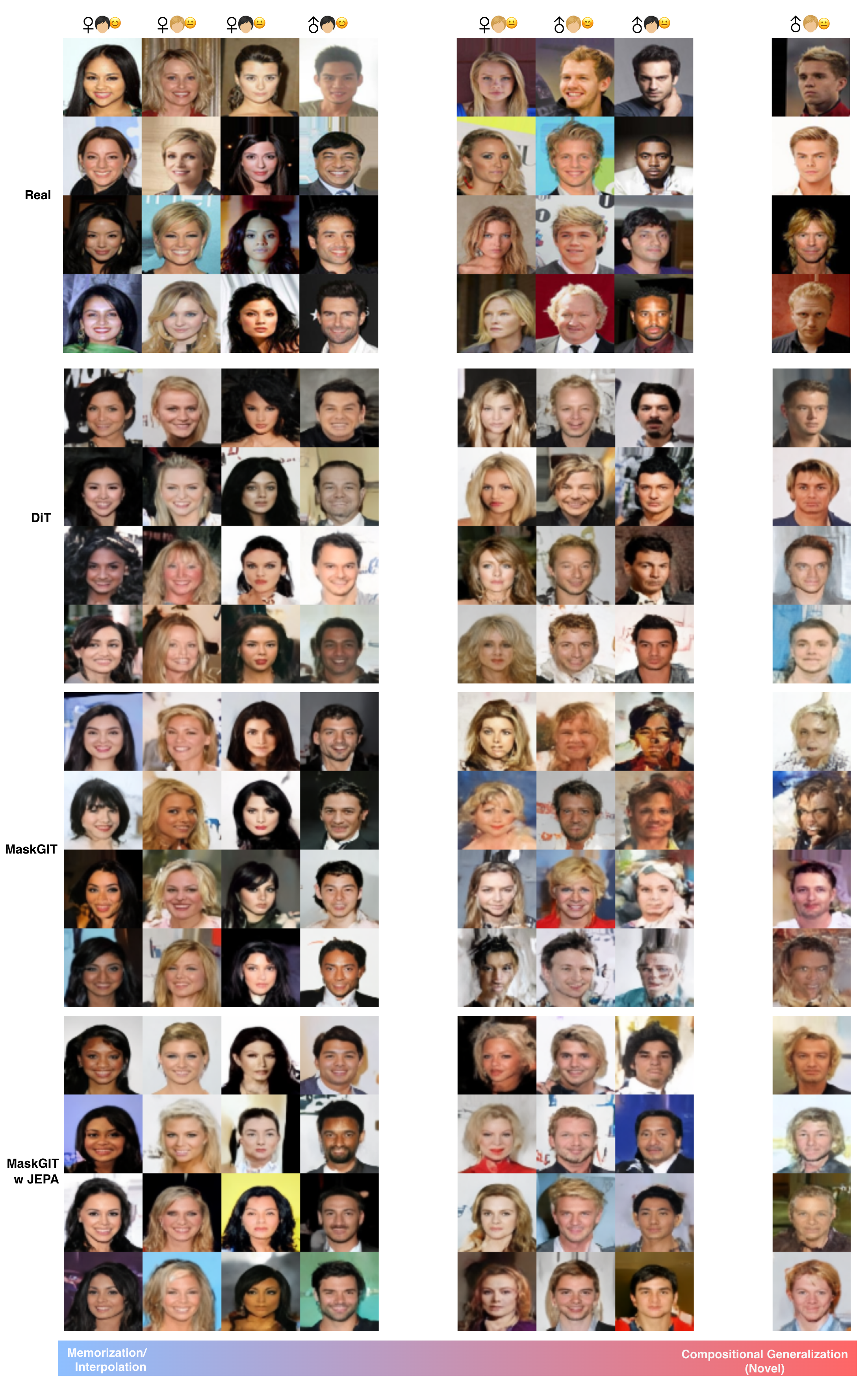}
    \caption{Qualitative Results on CelebA}
    \label{fig:celeb-results}
    \end{figure}

\cref{fig:celeb-results} gives a comprehensive overview of qualitative results on CelebA.

\subsection{CLEVRER-Kubric}

Qualitative results on CLEVRER-Kubric are presented as a separate website in the included HTML file. We show results for \texttt{111 - \{large red cube\}}, and can see that MaskGIT trained with the standard objective generates colors well but struggles to assign them to specific shapes. In contrast, MaskGIT trained with a JEPA-based objective and models like DiT which leverage a continuous latent space show better performance, as reflected by the probe accuracies (\cref{fig:latent_space_results_clevrer}).

\section{Worlds Models: CoVLA}
\label{appendix:covla_eval}

We use CoVLA~\citep{arai2025covla}, a real-world driving video dataset collected in Tokyo with synchronized front-facing camera and CAN/GNSS/IMU signals. We do the compositional splits on two factors: time of day (day~\daysym, night~\nightsym) and turn direction (left~\Ldir, straight~\Sdir, right~\Rdir). The held-out novel compositions are \fact{\daysym}{\Rdir} and \fact{\nightsym}{\Ldir}; all remaining combinations constitute the \emph{real-train} set.

\paragraph{Labeling and segmentation.}
We assign the night label using \texttt{in\_tunnel} flag or explicit night time at the exact timestamp for the frame, label \emph{turns} when speed $\geq 1.8\,\mathrm{km/h}$ and steering angle $\geq 10^\circ$ (left/right), and segment videos so each clip contains exactly one scenario (no factor changes within a clip).

\paragraph{Models and training.}
We instantiate \textsc{Orbis}~\citep{mousakhan2025orbis} in two matched variants: Orbis-MaskGIT (categorical objective) and Orbis-DiT (continuous objective). Both use a spatiotemporal Transformer (24 layers, hidden size 768, input resolution $320\times 640$), trained for 18 epochs with identical step counts and batch sizes; optimizer and schedule follow \citet{mousakhan2025orbis}. Each model observes a 7-frame context (implicitly revealing the factors discussed) and autoregressively generates 30 frames one frame at a time. No clips from  \fact{\daysym}{\Rdir} or \fact{\nightsym}{\Ldir} appear in training.

\paragraph{Evaluation protocol.}
We assess compositionality via nearest-neighbor retrieval against real videos using V-JEPA2 features \citep{assran2025v}. For each generated clip, we: (i) encode frames; (ii) spatially pool to a $4\times 6$ grid (feature dim 1024); (iii) take the last 10 generated frames and uniformly subsample 5; and (iv) flatten to a descriptor of size $4\times 6\times 5\times 1024$. The real gallery contains 100 clips per composition (600 total). Distances are based on cosine similarity; we take $k=1$ nearest neighbors and report
$$
\mathrm{CRA}_k(c^\star)
= \frac{1}{k}\sum_{i=1}^{k}\mathbf{1}\bigl\{\,c(\mathrm{NN}_i)=c^\star\,\bigr\},
$$
averaged over generated clips targeting composition $c^\star$. With a balanced 6-class gallery, the chance is $1/6 \approx 0.167$.

We additionally report Fr\'echet Video Distance (FVD) between generated and real clips, computed separately on the seen and unseen splits (Table~\ref{tab:seen_unseen_fvd_covla}).

\begin{table}[t]
\centering
\caption{Nearest-neighbor compositional retrieval metric (CRA) comparing generated videos to real videos.
Factors are time of day: Day \daysym, Night \nightsym; directions: Left \Ldir, Straight \Sdir, Right \Rdir.
Rows show a generated novel split evaluated against \textit{real train} and \textit{real novel} classes.
The Hit@1 (correct target novel class) is bolded; \textbf{black bold} indicates better results.}

\label{tab:compgen_retrieval_pretty}
\begin{tabular}{ll S S S S S S}
\toprule
\multicolumn{2}{c}{} &
\multicolumn{4}{c}{Real train} &
\multicolumn{2}{c}{Real novel} \\
\cmidrule(lr){3-6} \cmidrule(lr){7-8}
Gen. novel & Model &
{\fact{\daysym}{\Ldir}} & {\fact{\daysym}{\Sdir}} & {\fact{\nightsym}{\Rdir}} & {\fact{\nightsym}{\Sdir}} &
{\fact{\daysym}{\Rdir}} & {\fact{\nightsym}{\Ldir}} \\
\midrule
\multirow{2}{*}{\fact{\daysym}{\Rdir}}
& DiT      & 0.190 & 0.190 & 0.060 & 0.060 & {\bfseries 0.470} & 0.030 \\
& MaskGIT  & 0.380 & 0.260 & 0.110 & 0.050 & \mgitval{0.180} & 0.020 \\
\midrule
\multirow{2}{*}{\fact{\nightsym}{\Ldir}}
& DiT      & 0.130 & 0.020 & 0.340 & 0.040 & 0.040 & {\bfseries 0.430} \\
& MaskGIT  & 0.150 & 0.030 & 0.560 & 0.100 & 0.020 & \mgitval{0.140} \\
\bottomrule
\end{tabular}
\end{table}

\begin{table}[t]

\centering
\caption{FVD on seen and unseen splits on CoVLA.}
\small
\begin{tabular}{lrr}
\toprule
Method & FVD (Seen) $\downarrow$ & FVD (Unseen) $\downarrow$ \\
\midrule
DiT &  \textbf{730.30} & \textbf{2257.51} \\
MG  & 2320.95 & 3197.74 \\
\bottomrule
\end{tabular}

\label{tab:seen_unseen_fvd_covla}
\end{table}

\begin{figure}[t]
  \centering
  \includegraphics[width=1\textwidth]{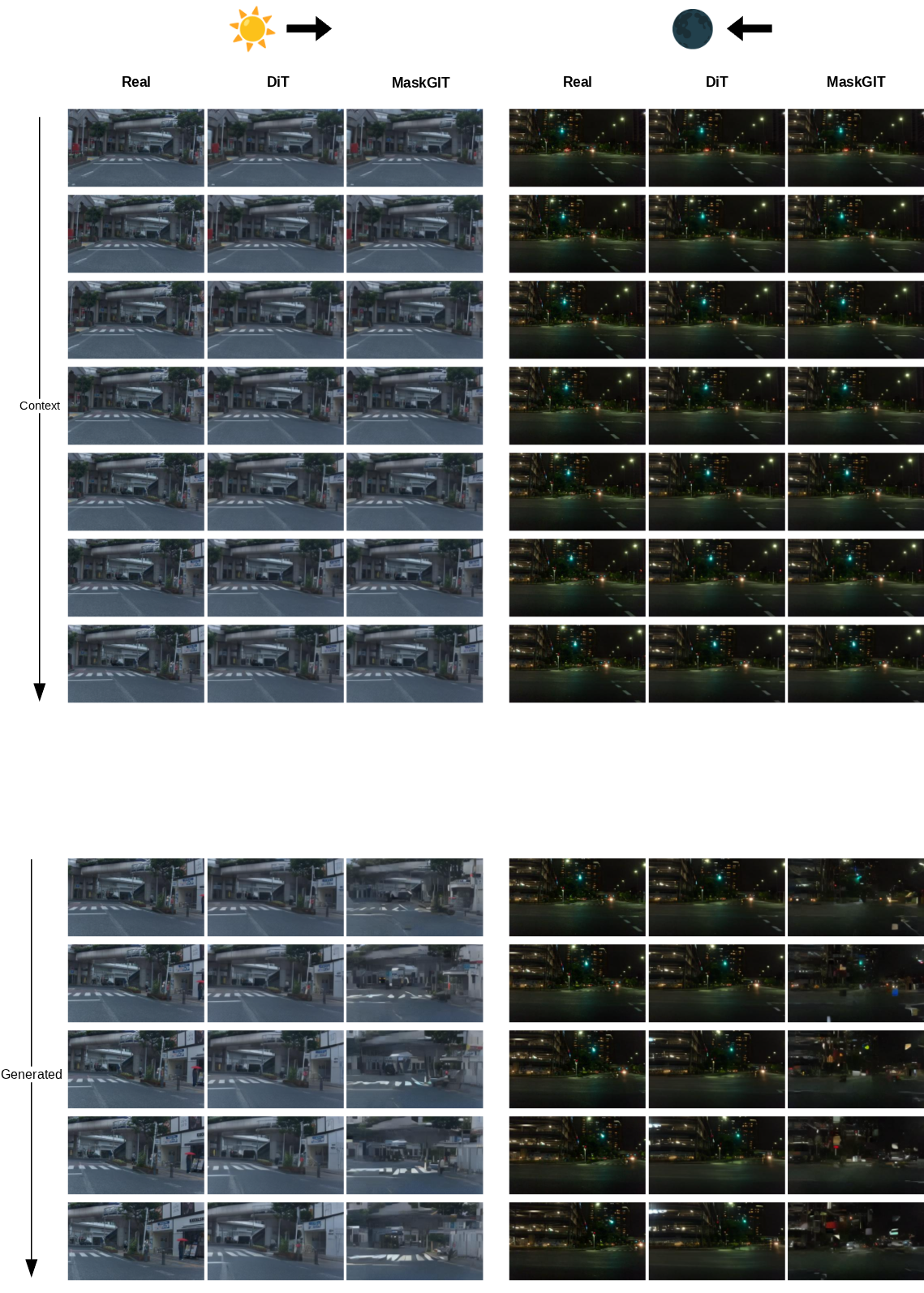}
    \caption{Qualitative results with Orbis on CoVLA. Two novel compositions are shown, \fact{\daysym}{\Rdir} and \fact{\nightsym}{\Ldir}, each conditioned on a 7-frame initial context and predicting 5 frames ahead. Orbis-DiT follows the target compositions, whereas Orbis-MaskGIT often struggles.}
  \label{fig:orbis-wide}
\end{figure}

\section{Language}
\label{appendix:language}

To test whether our findings on continuous representations playing a role in compositional performance extend beyond the visual domain, we constructed a controlled card arithmetic dataset taking inspiration from the Points24 dataset \citep{points24}. Each instance requires forming an arithmetic expression from four playing cards from a standard deck of 52 cards, such that all cards are used exactly once to reach a specified target value. To map cards to numerical values, the Ace was mapped to 1, the number cards (2-10) retained their face value, and face cards were mapped to the next integers (Jack = 11, Queen = 12, King = 13).

The data was split into compositional sets based on rules that the model had to follow while performing the arithmetic task: (i) The model could be asked to formulate an equation from the cards that satisfies a target value of either 12 or 24, (ii) cards belonging to red suits (Hearts, Diamonds) can either retain their original value (same as the cards belonging to the black suit (Clubs, Spades)), or have their value doubled. Training examples to the model contained only single-rule instances, while the test set required applying both rules simultaneously.

We use Llama-3.2 \citep{llama} and treat its reasoning mechanism as an analogue of the objective in our visual experiments. We compare the same base model under two reasoning strategies: (i) normal chain-of-thought (CoT) \citep{cot} which is the standard textual step-by-step reasoning, and (ii) COntinuous-Chain-Of-Thought (COCONUT) \citep{coconut}, where intermediate reasoning states are represented in a continuous latent space as ``thoughts'' rather than purely symbolic text.

Both models were trained on identical data and evaluated on the same compositional test split. We report the success rate, defined as the proportion of trials in which a programmatic verifier confirms that the output applies the correct rules, uses each card exactly once, and evaluates to the target. Averaged over three seeds on 500 compositional test instances, the success rate with continuous reasoning was 12.33 $\pm$ 1.3\%, higher than the standard CoT baseline at 4.87 $\pm$ 0.9\%. Although the absolute numbers are low due to the difficulty of the task, this preliminary result suggests that continuous reasoning states may benefit compositional generalization on this task.

This small-scale experiment serves two purposes: First, it provides preliminary evidence that the advantage of continuous representations observed in visual generative models may extend to a language reasoning task. Second, it highlights an early but promising signal that \textit{continuous reasoning for language} is a fruitful research direction. While the present setup is limited and minimal, it lays the groundwork for future studies that explore richer tasks, larger models, and more principled strategies for continuous reasoning.

\begin{figure}[t]
\centering
\begin{tcolorbox}[colback=white,colframe=black!50,width=\textwidth]

\textcolor{purple}{\textbf{System Prompt}} ($v_0^{in}$) \\[3pt]

\textbf{[Task Description]} \\
You are an expert in the card arithmetic game. You are observing four cards: each card must be used exactly once to form an arithmetic expression. Number cards take their face value (Ace = 1, 2--10 as given). The suits are abbreviated as Hearts = H, Diamonds = D, Clubs = C, Spades = S. Face cards are valued as J = 11, Q = 12, K = 13. In addition, if a card belongs to a red suit (Hearts H or Diamonds D), its value is doubled. Your goal is to output a formula that evaluates to the target number, which is either 12 or 24. The allowed operators are \texttt{+}, \texttt{-}, \texttt{*}, parentheses, and \texttt{=}. \\[6pt]

\textbf{[Input]} \\
Cards: [2H, 3S, QC, 4D] \\
Target: 24 \\[6pt]

\textbf{[Output]} \\
Your response should be a valid JSON file in the following format: \\
\begin{verbatim}
{
  "cards": [x, y, z, w],
  "values_after_rules": [a, b, c, d],
  "target": T,
  "formula": "an equation that equals T"
}
\end{verbatim}

\hrulefill \\[6pt]

\textbf{\textcolor{violet}{Model output} ($v_t^{out}$)} \\
\begin{verbatim}
{
  "cards": ["2H", "3S", "QC", "4D"],
  "values_after_rules": [4, 3, 12, 8],
  "target": 24,
  "formula": "(12+8)+(4*3)"
}
\end{verbatim}

\hrulefill \\[6pt]

\textbf{\textcolor{red}{Verifier Output} ($v_t^{ver}$)} \\
\textcolor{red}{This trial is unsuccessful because the proposed formula does not evaluate to 24.} \\

\end{tcolorbox}
\caption{Example system prompt, unsuccessful model output, and verifier feedback for our language task with compositional rules: target restriction and red-suit value doubling.}
\end{figure}

\begin{figure}[t]
\centering
\begin{tcolorbox}[colback=white,colframe=black!50,width=\textwidth]

\textcolor{purple}{\textbf{System Prompt}} ($v_0^{in}$) \\[3pt]

\textbf{[Task Description]} \\
You are an expert in the card arithmetic game. You are observing four cards: each card must be used exactly once to form an arithmetic expression. Number cards take their face value (Ace = 1, 2--10 as given). The suits are abbreviated as Hearts = H, Diamonds = D, Clubs = C, Spades = S. Face cards are valued as J = 11, Q = 12, K = 13. Your goal is to output a formula that evaluates to the target number, which is either 12 or 24. The allowed operators are \texttt{+}, \texttt{-}, \texttt{*}, parentheses, and \texttt{=}. \\[6pt]

\textbf{[Input]} \\
Cards: [AD, JH, 5D, 9S] \\
Target: 24 \\[6pt]

\textbf{[Output]} \\
Your response should be a valid JSON file in the following format: \\
\begin{verbatim}
{
  "cards": [x, y, z, w],
  "values_after_rules": [a, b, c, d],
  "target": T,
  "formula": "an equation that equals T"
}
\end{verbatim}

\hrulefill \\[6pt]

\textbf{\textcolor{violet}{Model output} ($v_t^{out}$)} \\
\begin{verbatim}
{
  "cards": ["AD", "JH", "5D", "9S"],
  "values_after_rules": [1, 11, 5, 9],
  "target": 24,
  "formula": "(11+9)+(5-1)"
}
\end{verbatim}

\hrulefill \\[6pt]

\textbf{\textcolor{teal}{Verifier Output} ($v_t^{ver}$)} \\
\textcolor{teal}{This trial is successful because the proposed formula evaluates to 24.} \\

\end{tcolorbox}
\caption{Example system prompt, successful model output, and verifier feedback for a single-rule training instance of our language task: target restriction without red-suit value doubling.}
\end{figure}

\end{document}